\documentclass[10pt,twocolumn,letterpaper]{article}
\ifdefined\pdfsuppressptexinfo \pdfsuppressptexinfo=-1 \fi
\ifdefined\pdfinfoomitdate \pdfinfoomitdate=1 \fi
\ifdefined\pdftrailerid \pdftrailerid{} \fi
\usepackage[pagenumbers]{cvpr} %

\usepackage{cuted} %

\usepackage{currfile} %

\usepackage{caption} %

\usepackage{microtype}

\usepackage{booktabs,multirow,graphicx,xspace,makecell}
\usepackage{amssymb}
\usepackage{arydshln} %
\graphicspath{{./}{fig/backbone/}} %
\usepackage{wrapfig} %

\definecolor{cvprblue}{rgb}{0.21,0.49,0.74}
\usepackage[pagebackref,breaklinks,colorlinks,allcolors=cvprblue]{hyperref}

\def\confName{CVPR}
\def\confYear{2027}

\newcommand{\ours}{VoxForge\xspace}
\newcommand{\nnfvit}{nnF-ViT\xspace}
\newcommand{\nnfcnn}{nnF-CNN\xspace}

\title{VoxForge: A Unified Foundation Model for\texorpdfstring{\\}{ } Dense 3D Features Across MRI, CT and PET}

\author{Rushil Anirudh$^{*}$\qquad Jayaraman J. Thiagarajan$^{*}$\\
Amazon AGI\\
{\tt\small \{rushila,\,jayjt\}@amazon.com}}

\usepackage{placeins}
\usepackage{float}   %

\newcommand{\suppcref}[1]{Supp.~\cref{#1}}
\newcommand{\Suppcref}[1]{Supp.~\cref{#1}}
\newcommand{\maincref}[1]{\cref{#1} of the main paper}
\newcommand{\Maincref}[1]{\Cref{#1} of the main paper}
\begin{document}
\maketitle
{\renewcommand{\thefootnote}{*}\footnotetext{Equal contribution.}}

\begin{abstract}
A unified 3D medical foundation model promises reusable, frozen features that serve a wide variety of tasks such as segmentation, classification and registration on MRI, CT and PET as well as specialist models do. Generalist models still need separate backbones for localized and global tasks, specialist models are confined to a single modality, with the concern that spanning modalities under-emphasizes fine anatomy, and both typically rely on end-to-end fine-tuning to be competitive. We present \ours, a single encoder that integrates recent advances in 3D architecture, self-distillation and cross-modal curricula to learn transferable features from permissively licensed MRI and CT volumes. We evaluate it strictly frozen, the test a reusable foundation model must pass, against state-of-the-art generalist
and specialist models on segmentation, classification and registration across MRI and CT, and on PET, which it never
sees. \ours has the best or tied-best average rank in every task group, serves localized and global
tasks across modalities at the level of state-of-the-art specialists, and is extremely label-efficient, matching top-performing models on TotalSegmentator CT with a tenth of the labels. 
\end{abstract}

\begin{figure}[t]
  \centering
  \includegraphics[width=\linewidth]{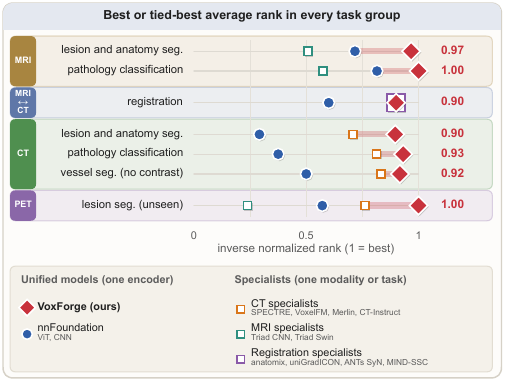}
  \caption{\textbf{\ours has the best or tied-best average rank in every task group.} For each task group, every
encoder is scored by its inverse normalized rank, averaged
over the group's cells (1 = best). All encoders are frozen and share the same heads. Each family shows its best
member, and a missing marker means the family is not scored on that group. A hollow marker around the diamond marks
a tie, and the red segment shows \ours's margin over the best other family, with its score at right. The PET row ranks Curia-MAE, which
belongs to no family and therefore has no marker. Values: \suppcref{tab:supp-teaser,tab:supp-rank-robust}.}
  \label{fig:teaser}
\end{figure}

\section{Introduction}
\label{sec:intro}
A radiologist recognizes the liver on MRI and on CT, on any scanner and under any contrast. A 3D medical foundation
model promises the same of its frozen features: a single encoder, trained once, that segments, classifies and matches
anatomy across tasks, scanners and modalities. In recent years, several insights toward this vision
have emerged, each in its own line of work. Primus~\cite{primus} showed that a plain ViT on fine $8^3$-voxel tokens
matches convolutional networks on 3D segmentation. In 2D, frozen self-distilled features rival a fine-tuned masked
autoencoder~\cite{dinov2,dinov3}, and in 3D, self-distillation has become a strong recipe for foundation models. For example, 
it underlies the CT specialists SPECTRE and VoxelFM~\cite{spectre,voxelfm}, as well as 3DINO~\cite{threedino}, whose multimodal
features generalize across organs and tasks. MedCoSS~\cite{medcoss} found that modalities conflict when pretrained
jointly and resolved the conflict by giving each modality its own stage. Even within a single modality, MRI presents
the same anatomy with different intensities across sequences, scanners and sites~\cite{bitar2006mrpulse}, so useful
features must separate anatomy from intensity.

Yet no model has combined these insights into the single encoder this vision requires. Even the largest generalist,
nnFoundation~\cite{nnfoundation}, concludes that transfer ``\emph{is governed not by a single universal model}'' and releases
two models: a CNN (\nnfcnn) that leads localized tasks when fine-tuned and a ViT (\nnfvit) that leads global and frozen ones. State-of-the-art models such as SPECTRE cover CT alone, and SPECTRE notes that a model spanning X-ray, CT and MRI ``\emph{may under-emphasize fine CT
anatomy compared to CT specific models}''. The primary objective of this work is to test both directly: we ask whether a single frozen encoder, built upon
these established insights, can match or outperform the best generalist and specialist models on both localized and global tasks across modalities.

We answer in the affirmative with \ours, a single frozen 3D encoder that combines these insights in one recipe (\cref{fig:method}). At its core, we build on the Primus ViT~\cite{primus}, whose fine $8^3$-voxel tokens preserve the detail that localized tasks need, which nnFoundation attributes to its CNN. We extend the Primus ViT for self-distillation with a class token, a wider encoder and a position table that grows with the crop. The DINOv3 self-distillation recipe with Gram anchoring~\cite{dinov3,ibot} then gives the tokens the global consistency that frozen readouts need. In contrast, \nnfvit, built on the same Primus family, reconstructs masked voxels~\cite{nnfoundation,mae}. Finally, an MRI-to-CT curriculum over
235K public volumes, staged as in MedCoSS, lets both modalities share one encoder without conflict, and intensity
augmentation separates anatomy from contrast (\cref{sec:method}).

Most 3D foundation models are evaluated by end-to-end fine-tuning, which re-learns the encoder for every task and
obscures what pretraining contributed. The stronger test is whether one frozen encoder can compete across all tasks,
because that is what makes a foundation model reusable. We compare the frozen \ours encoder with a suite of state-of-the-art generalist and specialist models across MRI and CT,
spanning segmentation, classification, registration and a contrast shift, and on PET, a modality it never sees
(\cref{sec:protocol}). All encoders share the
same heads and seeds, every margin carries a paired bootstrap interval, and every claim is checked against
pretraining exposure. \ours has the best or tied-best average rank in every task group (\cref{fig:teaser}). For frozen use, one encoder therefore suffices where nnFoundation released two, and pretraining across modalities keeps
the fine CT anatomy that SPECTRE warns such models under-emphasize.

\paragraph{Contributions.}
\begin{enumerate}
  \item \textbf{A unified frozen 3D encoder.} We introduce \ours, a 388M-parameter ViT that brings established
        insights into one recipe: fine $8^3$-voxel tokens and an extended Primus ViT~\cite{primus} for dense 3D
        prediction, DINOv3 self-distillation~\cite{dinov3} for local and global consistency, and an MRI-to-CT
        curriculum over permissively licensed volumes. In effect, one set of frozen weights serves MRI, CT and PET
        (\cref{sec:method}).
  \item \textbf{A strictly frozen, exposure-aware comparison.} Being competitive without fine-tuning is what lets one
        encoder serve every task. We therefore compare every encoder frozen, with shared heads, its
      own input contract and per-dataset pretraining-exposure checks, on MRI, CT and unseen PET, at label shares from
      1\% to 100\% (\cref{sec:protocol}).
  \item \textbf{State-of-the-art results across modalities.} \ours has the best or tied-best average rank in every
        task group (\cref{fig:teaser}). In particular:
        \begin{itemize}
          \item One frozen ViT serves both localized and global tasks. It matches or beats \nnfcnn on every
                segmentation task and \nnfvit on every classification target, so for frozen use a single model suffices.
          \item On PET, a modality absent from its pretraining, it is state-of-the-art among the frozen encoders we
                test, with the best lesion Dice and F1 across folds, outperforming both encoders pretrained
                on PET (\cref{fig:r1}).
          \item Pretrained across MRI and CT with an MRI-to-CT curriculum, it matches or exceeds the CT specialists on CT, with the best average rank in CT segmentation and classification and the best circle-of-Willis Dice from non-contrast
                CT, precisely the fine CT anatomy that SPECTRE warns models spanning modalities under-emphasize (\cref{fig:r4,fig:r1,fig:r2}).
          \item On MRI, it is state-of-the-art with every readout, including readouts that train nothing: a k-NN vote
                ranks it first on all three classification targets, and with no head its tokens localize lesions
                at about four times the AUPRC of the next-best encoder (\cref{fig:analysis}).
          \item Its tokens hold under contrast and modality shifts. Matched across MRI sequences, or between non-contrast CT
      and CT angiography of the same patient, they find their own location better than every other encoder. Across
      modalities, added to the standard multimodal descriptor MIND-SSC~\cite{heinrich2013mindssc}, they register MRI to CT as well as the registration-specialist model anatomix~\cite{dey2025anatomix} (\cref{fig:r3}).
          \item It is label-efficient: a quarter of the labels brings it within half a Dice point of \nnfvit trained on all of
      them, and with only 10\% of the TotalSegmentator CT labels it retains 97\% of its full-label Dice
                (\cref{sec:label-eff}).
        \end{itemize}
\end{enumerate}

\section{Related work}
\label{sec:related}
\noindent\textbf{Transformers and self-distillation for 3D segmentation.}
For 3D medical segmentation, CNNs have remained the strongest architectures. In a controlled from-scratch benchmark, MedNeXt-L and
ResEnc-L have been shown to outperform the transformer SwinUNETR~\cite{nnunetrevisited,mednext,swinunetr}. Primus~\cite{primus} traced
this gap to convolutional hybrids rather than to attention itself, and showed that a plain ViT on fine $8^3$-voxel
tokens, PrimusV2-M, matches ResEnc-L to within a tenth of a Dice point. In 2D, self-distillation
has made such ViTs strong frozen backbones. A frozen DINOv2 ViT with a linear head comes within a point of a
fine-tuned MAE~\cite{dinov2,mae}, and DINOv3 improves it even further~\cite{dinov3}. To our knowledge, these two lines of work have not been combined in 3D, and it is not typical to evaluate a 3D ViT solely as a frozen encoder. \ours addresses this gap by extending Primus, training it with a self-distillation recipe and evaluating it without fine-tuning as has become standard practice in general computer vision \cite{dinov2, dinov3}.

\noindent\textbf{Evaluating frozen features.}
Frozen 3D encoders are widely used as feature extractors for segmentation, classification and report
generation~\cite{ctclip,m3d,radfm,merlin}. Their relative ranking, however, depends strongly on the evaluation
protocol. On CT, the order of encoders changes with the evaluation setting~\cite{bigbright}, TAP-CT~\cite{tapct}
reports a linear head that outperforms Curia's~\cite{curia}, and Curia-MAE~\cite{curiamae} finds that nnU-Net trained
from scratch~\cite{nnunet} still surpasses frozen encoders. Head-free readouts avoid this dependence on a trained
head. Hummingbird~\cite{hummingbird} retrieves dense labels from a bank of patch features, and
PatchCore~\cite{patchcore} and AnomalyDINO~\cite{damm2025anomalydino} detect anomalies by their distance to normal
patches. Since pretraining data can also inflate any readout, overlap between pretraining and test data must be
checked for each benchmark~\cite{barz2020,sainz2023}. We therefore evaluate \ours with both trained heads and
head-free readouts (\cref{sec:backbone}), and flag every encoder whose pretraining data include a test set
(\cref{sec:protocol}).

\noindent\textbf{Combining MRI and CT in pretraining.}
Most 3D medical foundation models are pretrained on a single modality, most often CT~\cite{spectre,voxelfm,tapct,
merlin,ctfm,ctinstruct} due to abundance of available data, while the large public MRI corpora, OpenMind~\cite{openmind}, FOMO60K~\cite{fomo25} and Triad-131K~\cite{triad}, consist mostly of brain scans. The few models trained on both MRI and CT, such as
3DINO~\cite{threedino}, a self-distilled ViT, and Curia-MAE~\cite{curiamae}, a masked-reconstruction CNN, pool the
two modalities in a single training stage. Pooling, however, has a cost. MedCoSS~\cite{medcoss} shows that modalities
conflict in self-supervised pretraining and resolves the conflict by training each modality in its own stage while
rehearsing earlier ones, related to curriculum learning~\cite{bengio2009curriculum} and to continued
in-domain pretraining in language models~\cite{gururangan2020dontstop}. \ours adopts this staging and rehearsal but
reverses the order: MRI before CT. Because MRI sequences present the same anatomy under many intensities,
this order teaches the encoder to separate anatomy from intensity before it encounters CT.

\section{VoxForge}
\label{sec:method}
This section describes how \ours turns the insights of \cref{sec:intro} into one unified 3D encoder across modalities.

\subsection{Pretraining data}
\label{sec:method-data}
\begin{figure}[t]
  \centering
  \includegraphics[width=\linewidth]{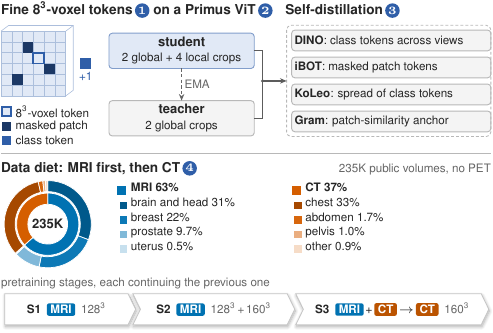}
  \caption{\textbf{The \ours recipe: four insights (circled) in one pipeline.} Rings: pretraining volumes by modality (inner) and anatomy (outer), with CT collections grouped by their target anatomy. Counts per collection and stage: \suppcref{tab:supp-corpus,tab:supp-corpus-mri,tab:supp-corpus-ct,tab:pretrain-stages}.}
  \label{fig:method}\label{fig:corpus}
\end{figure}

We curate a pre-training corpus consisting of 235K public, license-vetted adult volumes from about 71K subjects: 63\% MRI, 37\% CT and no PET
(\suppcref{tab:supp-corpus}). The MRI comes from 21 TCIA~\cite{clark2013tcia} collections and 895 OpenNeuro~\cite{markiewicz2021openneuro}
datasets and covers mostly the brain, breast and prostate, and the CT comes from 49 TCIA collections, mostly
NLST~\cite{nlst2011} chest CT (\cref{fig:method}). A collection is included only if its license permits commercial
use and derived works, so the corpus can be released with the model. We audit the corpus against every evaluation
set: it holds none of our test sets (\suppcref{sec:supp-scope}).

\begin{figure*}[t]
  \centering
  \includegraphics[width=\textwidth]{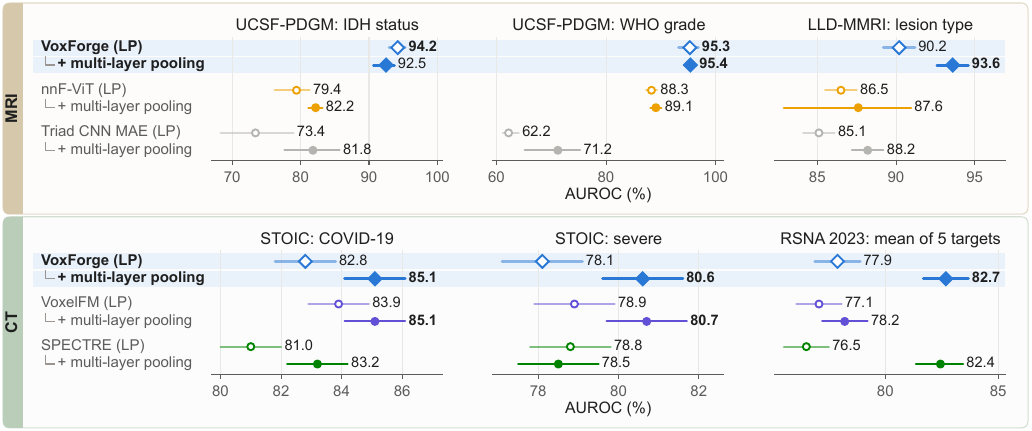}
 \caption{\textbf{Frozen \ours tokens retain patient-level labels on MRI and CT.} AUROC (\%) of a linear probe (LP)
and of the multi-layer attention probe (\cref{sec:protocol}), both trained on frozen tokens, with 95\% bootstrap
CIs. Bold marks the best result per panel. The RSNA 2023 score averages its targets. VoxelFM's self-supervised pretraining includes the RSNA 2023 images. The figure shows the top three encoders. All values: \suppcref{tab:mri-cls,tab:ct-cls}.}

  \label{fig:r4}
\end{figure*}

\subsection{Model, objective and staged pretraining}
\label{sec:method-pretrain}
\noindent\textbf{A Primus ViT modified for self-distillation.}
We modify Primus-L~\cite{primus}, a plain ViT on $8^3$-voxel tokens designed for 3D segmentation, so that it
self-distills. We add a learnable class token, which the DINO loss needs and Primus does not have, and widen the
encoder from 1056 to 1152 channels, to 388M parameters at $128^3$ (392.6M at $160^3$, with the larger position table). We resize its learned position table as the crop grows from
$128^3$ to $160^3$, and we drop the Primus patch decoder (\suppcref{sec:supp-impl}). Positions use the learned table and
axial 3D rotary position embeddings (RoPE), and one $160^3$ crop gives 8,000 patch tokens and the class token. The patch embedding is the only convolution, so every task head reads transformer
tokens.

\noindent\textbf{Self-distillation.}
We use the DINOv3 recipe~\cite{dinov3} because it trains every patch token as well as the class token. The student
sees global and local crops, its exponential-moving-average teacher only the global ones. DINO~\cite{dino} matches the
student's class tokens to the teacher's across views, and KoLeo~\cite{koleo} spreads them across a batch.
iBOT~\cite{ibot} masks some patch tokens in half of the global crops, and the student predicts the teacher's tokens there.
Gram anchoring holds patch-token pairwise similarities to an earlier teacher snapshot, which keeps dense structure
stable~\cite{dinov3}. A new snapshot at each stage lets each stage anchor the next.

\noindent\textbf{One intensity range for both modalities.}
MRI intensities depend on the sequence~\cite{bitar2006mrpulse}, while CT uses Hounsfield units. Both are rescaled to
$[0,1]$ between low and high per-volume percentiles, CT after clamping at $-1024$\,HU, so every stage sees one input
range.

\noindent\textbf{Curriculum in three stages.}
Pretraining runs in three stages (S1 to S3). Each stage starts from the previous student, teacher and heads, with a
fresh optimizer and schedule. S1 trains on MRI from random weights at $128^3$. S2 trains on MRI at higher resolution, \textit{i.e.}, adds $160^3$ crops and alternates the two sizes, so that volumes too small for $160^3$ stay in training. In S1 and
S2, GIN~\cite{gin} remaps the MRI intensities, and a histogram shift remaps half of the crops at random. Finally, the S3 stage adds CT.
As MedCoSS~\cite{medcoss} rehearses earlier modalities when it adds a new one, S3 first trains on MRI and CT together,
with CT close to half of the draws, so that the encoder learns CT contrast while its MRI features keep a training
signal, and then on CT alone. The schedule sees about 60M samples in 216
training hours on 64 H200 GPUs (\suppcref{tab:pretrain-stages,sec:supp-impl}). We evaluate a checkpoint from each stage
(\cref{sec:le-stages}). The S3 stage, which has seen the most diverse data, delivers the final \ours checkpoint.

\begin{figure*}[t]
  \centering
  \includegraphics[width=\textwidth]{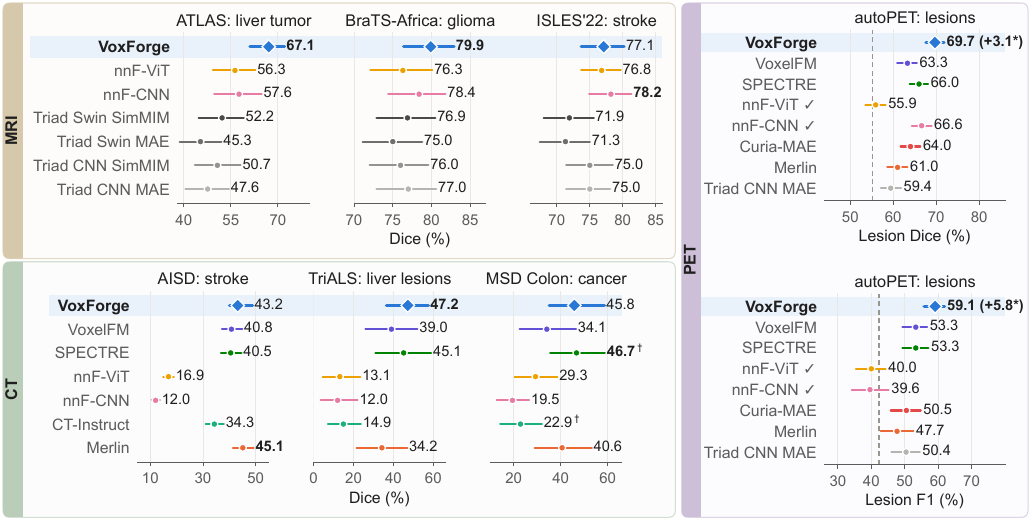}
\caption{\textbf{\ours is first or close behind on every lesion task and first on held-out PET.} Dice (\%)
of frozen encoders with trained decoder heads, with 95\% bootstrap CIs, and bold marks the best result. TriALS is
scored on the 37 cases whose lesions are predominantly visible on non-contrast CT (\suppcref{tab:ct-seg}). PET reports Dice over lesion-positive studies (top) and lesion F1 (bottom). Parentheses give
\ours's margin over the next encoder, and $\ast$ marks a significant margin. $^\dagger$ denotes an encoder whose
pretraining includes the test-set images, and for CT-Instruct also their labels. On PET, $\checkmark$ marks
pretraining on PET, and the dashed line shows \ours with random weights. Values:
\suppcref{tab:mri-seg,tab:ct-seg,tab:pet,tab:supp-pet-froc,sec:supp-pet}.}
  \label{fig:r1}\label{fig:ct-path}
\end{figure*}

\section{Evaluation protocol}
\label{sec:protocol}
Every encoder is frozen as released, and only a task head is trained on it. Every decoder head is scored at its final checkpoint, so that the comparison measures the frozen features.

\noindent\textbf{Encoders.}
We compare against recent, widely used state-of-the-art encoders, both generalist (i.e. intended to be an FM for MRI/CT/PET) and specialist (modality specific, only MRI or only CT). The generalist is
nnFoundation~\cite{nnfoundation}, which releases a CNN (102M parameters) and a ViT on $8^3$-voxel tokens (\nnfvit,
674M), both trained by masked reconstruction on 2.1M CT, MRI and PET volumes. The CT specialists are two
self-distilled ViTs, SPECTRE~\cite{spectre} (339M, about 230K CT volumes) and VoxelFM~\cite{voxelfm} (110M, about 137K
CT studies), and two encoders trained with other objectives, Merlin~\cite{merlin}, an inflated 3D ResNet-152
supervised by radiology reports, and CT-Instruct~\cite{ctinstruct}, a 330M hybrid ResNet--transformer. The MRI
specialists are four Triad encoders~\cite{triad}, a CNN and a Swin-B, each trained with MAE~\cite{mae} and with
SimMIM~\cite{simmim}.

\noindent\textbf{Inputs and heads.}
Each encoder reads its input under its own intensity normalization (\suppcref{sec:supp-impl}). The encoders were
pretrained under different protocols that cover different spacing ranges (\suppcref{tab:supp-geometry}), so no single
grid suits all of them. When nnFoundation's native spacing falls within an encoder's pretraining range, we use it. Otherwise, each axis is clipped into the encoder's
pretraining range, which gives the grid nearest the native spacing that the encoder has seen. Fairness comes from matching each encoder's downstream input to its pretraining. For segmentation, every ViT gets the same
UNETR decoder~\cite{unetr}, and every hierarchical encoder (the CNNs, CT-Instruct and Triad's Swin) gets a
UNet3D~\cite{cicek2016unet3d} on its feature pyramid. Classification uses three readouts of increasing capacity:
a training-free k-NN vote on the patch mean ($k{=}20$), a linear probe on the mean token of the last tap, and a
multi-layer attention probe~\cite{ilse2018abmil} that pools the tokens of all $T$ taps with gated attention.

\noindent\textbf{Benchmarks.}
We evaluate on nine MRI and seven CT segmentation benchmarks, patient-level classification on two MRI and two CT
datasets, a contrast shift on ISLES'24 (\cref{sec:res-ncct}) and the held-out modality on autoPET (\cref{sec:pet}).
\Suppcref{tab:datasets} lists every dataset with its targets.

\noindent\textbf{Statistics.}
Unless stated otherwise, tables report means over three folds. For segmentation, we first average each case's Dice across the three head-training seeds and then average over cases. Classification uses one fixed split, and its point estimates are the mean AUROC over the three probe seeds. Segmentation, classification and contrast shift results carry 95\% intervals from a case-and-seed bootstrap. In each draw, cases or patients are resampled with replacement within folds, and one of the three seeds is sampled independently for each model. Pairwise draws use the same resampled cases for both models and record the difference. Finally, the 2.5th and 97.5th percentiles is used to define the 95\% interval. A margin is significant when its paired 95\% interval excludes zero. Note, $\dagger$ marks an encoder whose pretraining sources include images (not labels) of the test set.

\begin{figure*}[t]
  \centering
  \includegraphics[width=\textwidth]{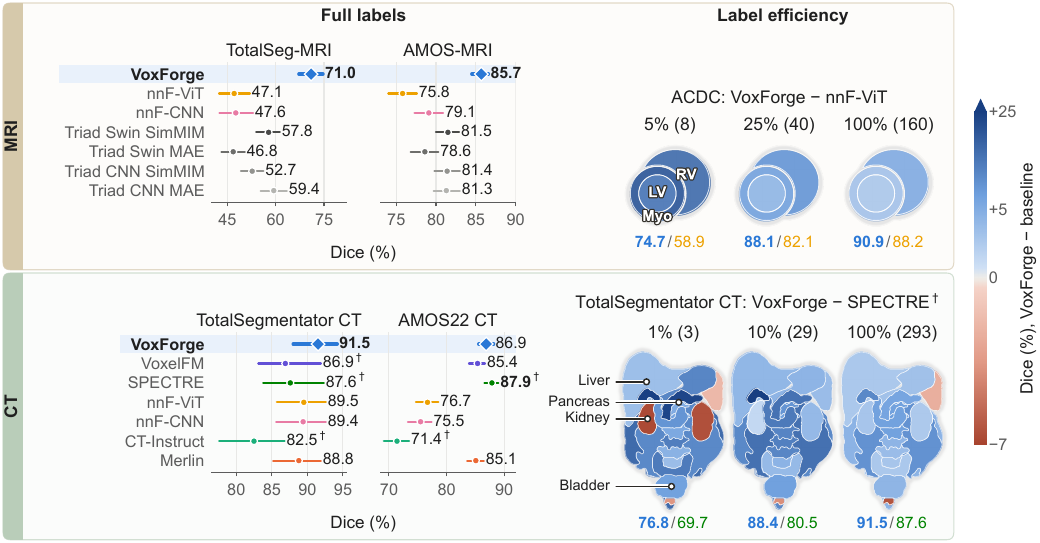}
\caption{\textbf{\ours leads three of four multi-organ benchmarks with all labels, and its lead over \nnfvit on ACDC
grows as labels shrink.} The left panels show Dice (\%) with all labels, and 95\% bootstrap CIs. The maps show the
per-structure Dice difference between \ours and the baseline on a schematic heart and torso, at three label shares
with the number of training cases in parentheses, and the mean Dice of \ours and the baseline below each map.
$^\dagger$ marks an encoder whose pretraining includes the test-set images.
Values: \suppcref{tab:mri-seg,tab:ct-seg,tab:supp-le,tab:supp-ts-le,tab:supp-anat-maps}.}

  \label{fig:r2}\label{fig:ct-ts}
\end{figure*}

\section{Results}
\label{sec:results}
\subsection{\ours tops MRI classification and matches the CT specialists}
\label{sec:res-cls}

The frozen \ours tokens carry patient-level labels on both modalities, and \cref{fig:r4} compares them with other encoders. On MRI, the linear probe puts \ours first on all three targets, 4 to 15 AUROC points ahead
of the next encoder, \nnfvit (\suppcref{tab:mri-cls}). On CT, the same probe puts \ours within about one point of the CT specialist
VoxelFM~\cite{voxelfm} on STOIC and ahead of it on RSNA 2023, although VoxelFM used RSNA 2023 images in pretraining.

\begin{figure*}[t]
  \centering
  \includegraphics[width=\textwidth]{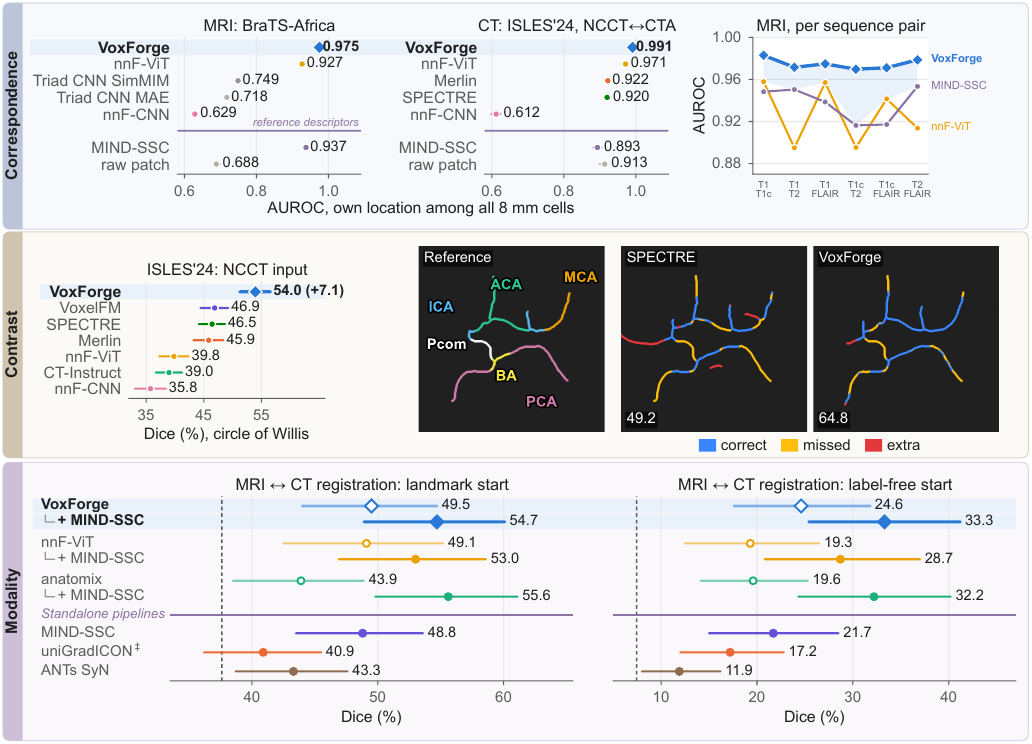}
  \caption{\textbf{\ours tokens hold under shifts in contrast and modality.} \textbf{Correspondence}: AUROC with which each
token ranks its own location among all cells of a co-registered second contrast, across the MRI sequences of
BraTS-Africa and between non-contrast CT (NCCT) and CT angiography (CTA) on ISLES'24. \textbf{Contrast}: circle-of-Willis
Dice (\%) segmented from NCCT. \textbf{Modality}: Dice (\%) of MRI-to-CT registration, with the dashed line marking
pre-alignment only. Bars show 95\% bootstrap CIs, and parentheses give the margin over the next encoder. $^\ddagger$ marks a method
for which no setting meets the 1\% folding cap. All rows:
\suppcref{tab:supp-correspondence,tab:supp-corr-pairs,tab:isles,tab:supp-reg}.}

  \label{fig:r3}\label{fig:ct-cow}\label{fig:registration}
\end{figure*}

On CT, the mean token hides part of what the tokens hold. Consequently, the attention probe raises \ours by 2 to 5 AUROC points. It
then ties VoxelFM on STOIC COVID-19 at 85.1, is level on severity and leads on RSNA 2023 at 82.7 (\suppcref{tab:ct-cls}).
The gain from the attention probe is specific to the self-distilled ViTs. It improves SPECTRE and VoxelFM by similar
amounts, whereas every other encoder loses performance with it (\suppcref{tab:supp-ct-probe-gain}). Because an
organ injury occupies only a small part of an abdominal CT, averaging all tokens into one dilutes it. We
attribute the gain to self-distilled tokens preserving this local evidence, which the attention probe can recover
and the mean token cannot.

\subsection{Lesion segmentation at specialist level, and best on unseen PET}
\label{sec:res-lesion}\label{sec:pet}

\Cref{fig:r1} compares \ours with state-of-the-art encoders on six representative lesion tasks, three on MRI and
three on CT, and \suppcref{tab:mri-seg,tab:ct-seg} report every segmentation benchmark. No specialist leads across
the six, and \ours is first or close behind on every one. It has the best Dice on three, most clearly on ATLAS liver
tumors, where it is about 10 points ahead of the next encoder, \nnfcnn. On the other three, it trails by one to two
points, and the leading encoder changes from task to task.

FDG PET is the sharpest test of a unified representation, because no stage of \ours trains on it. We segment
FDG-avid lesions from the PET channel alone, in five-fold cross-validation over all 1,014 autoPET studies, and each
encoder reads PET on its own CT input scale (\suppcref{sec:supp-pet}). For this task we add the baseline Curia-MAE~\cite{curiamae}, a
ResEnc-L pretrained on CT and MRI. \ours segments the lesions best and has the highest lesion F1 in every fold (\suppcref{tab:pet,tab:supp-pet-froc,sec:supp-pet}). It also has the
highest threshold-free FROC score and the fewest false-positives. Both encoders that nnFoundation
pretrains on PET fall behind it: \nnfcnn is second on lesion Dice and last on lesion F1, and \nnfvit is last on lesion Dice.
Finally, our control, \ours with random weights trails \ours
consistently (\suppcref{tab:pet,sec:supp-pet}).
\subsection{Anatomy segmentation and label efficiency}
\label{sec:res-anatomy}\label{sec:label-eff}

With all labels, \ours leads three of the four multi-organ benchmarks, by about 12 Dice points on TotalSeg-MRI
and by smaller margins on AMOS-MRI and TotalSegmentator CT (\cref{fig:r2}). It trails only on AMOS22 CT, by about a
point, against SPECTRE, which saw AMOS22 images in pretraining (\suppcref{tab:mri-seg,tab:ct-seg}). Across all segmentation tasks, \ours matches or beats \nnfcnn, the model nnFoundation recommends for localized tasks.

\begin{figure*}[t]
  \centering
  \includegraphics[width=\textwidth]{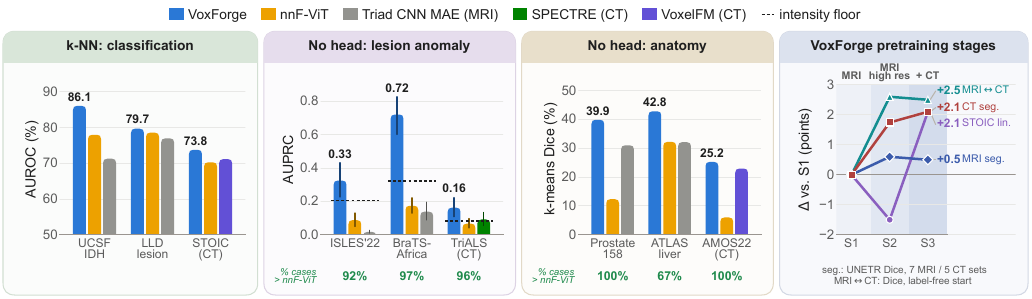}
  \caption{\textbf{Frozen \ours tokens carry the task with a k-NN vote and with no head at all.} \textbf{k-NN}:
AUROC (\%) of a training-free k-NN vote on the patch mean, with COVID-19 as the STOIC target. \textbf{Lesion anomaly}:
voxel AUPRC of the cosine distance to a bank of lesion-free tissue, with 95\% bootstrap CIs, where the TriALS bank
holds every lesion-free training liver. \textbf{Anatomy}: Dice (\%) of k-means with 16 clusters, each structure
matched to one cluster. The green row gives the share of cases on which \ours beats \nnfvit. \textbf{Stages}: change
from S1 to S2 and S3, one run each, where lin.\ denotes the linear probe. Values:
\suppcref{sec:supp-analysis,tab:supp-le-stages}.}

  \label{fig:analysis}
\end{figure*}

Labels are scarce in practice, so we retrain every head on nested subsets of the training
cases, the same for every encoder we sweep (\suppcref{tab:supp-le}). Against \nnfvit, the margin grows as subsets shrink on
ACDC, MSD Hippocampus and ISLES'24. On cardiac MRI, it reaches 13 to 19 points per structure with eight cases. A
quarter of the labels brings \ours within half a Dice point of \nnfvit trained on all of them on ACDC and MSD
Hippocampus, and a tenth on ISLES'24 and AISD. On TotalSegmentator CT, \ours leads by about two points with all 293 cases.
Training with just three cases (1\%), it beats SPECTRE on 16 of the 20 structures, and with 29 cases (10\%) it matches SPECTRE trained on all 293
(\suppcref{tab:supp-ts-le,tab:supp-anat-maps}).

\subsection{Generalization: the tokens hold under shifts in intensity, contrast and modality}
\label{sec:res-robust}\label{sec:res-ncct}\label{sec:registration}

\noindent\textbf{Correspondence.} A change of MRI sequence or of contrast phase moves intensities in ways no synthetic
remap reproduces, so we ask whether a token still finds its own location in the other image (\cref{fig:r3}). We shift the crop of a
co-registered second contrast by 16 to 32\,mm and let every 8\,mm cell of one image rank all cells of the other. Across
the four MRI sequences of BraTS-Africa and from non-contrast CT to CT angiography, \ours ranks the true location highest
of all encoders, and its best match lies within 8\,mm of it for half of the MRI cells and two thirds of the CT cells,
1.7 to 1.8 times as often as MIND-SSC~\cite{heinrich2013mindssc}, the classical cross-contrast descriptor
(\suppcref{tab:supp-correspondence}). Under seven intensity remappings, \ours tokens change less
than those of every other encoder on MRI, CT and the unseen PET, except
under Gaussian noise on CT (\suppcref{tab:supp-robust}).

\noindent\textbf{Contrast.} ISLES'24 gives circle-of-Willis masks on CT angiography, a silver standard not traced by
experts. We train and test the head on the
co-registered non-contrast CT, where the vessels are barely visible and the encoder has to infer them from the
surrounding anatomy. \ours segments the circle of Willis best, about 7 Dice points above the next encoder. This is the fine CT anatomy that SPECTRE warns models spanning modalities under-emphasize. \ours also loses
the least to the missing contrast (\suppcref{tab:isles}).

\noindent\textbf{Modality.} In MRI-to-CT registration, \ours tokens match anatomix~\cite{dey2025anatomix}, a network
built for the task. We register TotalSeg-MRI to TriALS CT of other patients, both ways, from landmarks or label-free.
Following DINO-Reg~\cite{song2024dinoreg}, a PCA of the frozen tokens drives ConvexAdam~\cite{siebert2025convexadam},
alone or concatenated with MIND-SSC. Every method is tuned on the same 28 non-test pairs under a 1\% folding cap
(\suppcref{sec:supp-registration}). Alone, the tokens beat anatomix by about 6 Dice points from landmarks and \nnfvit
by about 5 from the label-free start. Added to MIND-SSC, they raise its Dice by about 6 points from landmarks and 12 from the label-free start,
level with anatomix combined the same way and ahead of every published pipeline we test, among them
uniGradICON~\cite{tian2024unigradicon} and ANTs SyN~\cite{avants2008syn} (\suppcref{tab:supp-reg}).
\section{Analysis}
\label{sec:analysis}

To see what the frozen tokens hold without a decoder, we read them with a k-NN vote and with no head at all (\cref{fig:analysis}).

\noindent\textbf{On MRI, clinical labels form neighborhoods.}
\label{sec:ana-retrieval}
A k-NN vote over the nearest training patients puts \ours first on all three MRI classification targets, by the widest margin on the two glioma targets (\suppcref{tab:mri-cls}). The class token carries the glioma labels: its vote reaches about 92 AUROC on both, against 86 for the patch mean (\suppcref{tab:supp-clstok}).

\noindent\textbf{On MRI, the tokens localize lesions with no head.}
\label{sec:backbone}
Following PatchCore and AnomalyDINO~\cite{patchcore,damm2025anomalydino}, we score each token by its cosine distance to a bank of lesion-free tissue. On stroke and glioma MRI, \ours scores about four times the AUPRC of \nnfvit, the next encoder, and it is the only MRI encoder that beats the z-scored voxel intensity on both (\suppcref{tab:supp-lesion}). On CT liver lesions, the result depends on the bank. With every lesion-free training liver, \ours leads SPECTRE, and with three reference livers it trails. Large lesions dominate both means.

\noindent\textbf{The tokens group anatomy with no labels.}
k-means on each case's frozen tokens, with each structure matched to one of 16 clusters, gives \ours the highest Dice on
Prostate158, ATLAS liver and AMOS22 CT (\suppcref{tab:supp-kmeans}).

\noindent\textbf{The full curriculum gives the most well-rounded checkpoint.}
\label{sec:le-stages}
Each stage raises the mean score over the 20 stage results, and since each changes the data, not every task rises in turn. The S2 checkpoint, after higher-resolution MRI, raises CT segmentation by 1.7 Dice points, label-free registration by 2.6 and CT token robustness by 4.1 points, and its linear probes fall below S1. The S3 checkpoint, after CT at the same resolution, closes these gaps, lifting the linear probes by 2.4 and 3.6 points, and holds S2's other gains (\suppcref{sec:supp-stage-ablation}). Each stage is one run that continues the previous one, so data and training length change together. The stages trace the path of one model through the curriculum, and do not isolate the effect of resolution or of CT.

\section{Limitations and conclusion}
\label{sec:conclusion}
A controlled ablation of each ingredient (objective, data mix and token size) costs thousands of GPU-hours per arm and remains future work. Public, permissively licensed 3D data is another bottleneck: our 235K volumes are about 10\% of nnFoundation's corpus, and the scaling of frozen 3D features with data, model size and training is unmeasured. The corpus holds no PET, ultrasound or radiography, and every pretraining crop spans at most $160^3$ voxels, so pretraining across all modalities, at native resolution and over whole-body fields of view remains open. Finally, published 3D foundation models report results with different heads, splits and exposure checks, so their numbers do not compare across papers. We match each encoder's input to its own pretraining and document every choice (\suppcref{tab:supp-geometry,sec:supp-scope}).

\ours shows that one frozen 3D encoder can serve MRI, CT and an unseen modality at the level of state-of-the-art
specialists. Evaluated strictly frozen, it has the best or tied-best average rank in every task group, leads on FDG
PET and keeps its lead on MRI even without a trained head.

\section*{LLM Usage Statement}
The authors used LLMs to assist with experimentation, data curation, auditing the evaluation sets
for leakage into pretraining data, figure plotting, identifying related work, and writing and framing the text. After multiple rounds of revision, the authors rewrote the final text themselves and checked every number, claim and reference in the paper.

{
    \small
    \ifdefined\bibsep \setlength{\bibsep}{0pt}\fi %
    \bibliographystyle{ieeenat_fullname}
    \bibliography{main}
}
\clearpage
\makeatletter\def\Hy@backout#1{}\makeatother %
\onecolumn   %
{\centering\Large\textbf{Supplementary Material: \thetitle}\par\vspace{1.0em}}
\setcounter{section}{0}\renewcommand{\thesection}{A.\arabic{section}}   %
\renewcommand{\theHsection}{A.\arabic{section}}   %
\setcounter{table}{0}\renewcommand{\thetable}{S\arabic{table}}\renewcommand{\theHtable}{S\arabic{table}}   %
\setcounter{figure}{0}\renewcommand{\thefigure}{S\arabic{figure}}\renewcommand{\theHfigure}{S\arabic{figure}}   %
{\centering\small Citations refer to the reference list of the main paper.\par\vspace{0.6em}}
\FloatBarrier
\section{Implementation details}
\label{sec:supp-impl}
This section gives the settings behind \maincref{sec:method-pretrain,sec:protocol}: the model and its pretraining
objective; the heads, splits, statistics, significance rules and lesion-anomaly probe of the frozen evaluation; and
the intensity normalization and input spacing of every encoder. \Cref{sec:supp-stages} gives the pretraining data, stages and settings, and
\cref{sec:supp-registration} the registration protocol.

\paragraph{Architecture.}
We modify Primus-L~\cite{primus}, a plain ViT on $8^3$-voxel tokens designed for 3D segmentation, in four ways so that
it self-distills. First, we add a learnable class token, which the DINO loss needs and Primus lacks. Second, we
widen the encoder from 1056 to 1152 channels and keep its 24 blocks and 16 heads, which raises it from 326.5M to 388.1M
parameters at a $128^3$ input. Third, its learned position table includes the class token and grows from $16^3{+}1$ to
$20^3{+}1$ entries when the crop grows to $160^3$, where the encoder has 392.6M parameters. Fourth, we drop the Primus
patch decoder, since pretraining uses only the encoder. Positions use the learned table and axial 3D RoPE.

\paragraph{Objective.}
\ours self-distills with the DINOv3 recipe~\cite{dinov3}. DINO~\cite{dino} and iBOT~\cite{ibot} each use 16,384
prototypes, and KoLeo~\cite{koleo} has weight 0.1. Each volume gives two global and four local crops, and half of the
global crops have 10--50\% of their tokens masked. The exponential-moving-average (EMA) teacher runs at temperature 0.07, reached from 0.04 over the
first 30 passes of S1, and the optimizer is AdamW (weight decay 0.04, layer-wise decay 0.9). The three pretraining stages see
59.7M samples, about 254 passes over the 235K-volume corpus, in 216 hours on 64 H200 GPUs.

\paragraph{Heads.}
All encoders share each dataset's head schedule. UNETR~\cite{unetr} has about 9.5M parameters. On MRI and CT, every
head trains with AdamW (weight decay 0.05), a 15-epoch warm-up and polynomial decay, for 300 epochs of 250
iterations at a peak learning rate of $10^{-3}$.
TotalSegmentator CT and TotalSeg-MRI train on 8 GPUs at batch 16 for 250 epochs at $3{\times}10^{-3}$, the
square-root scaling of the batch-2 rate. Every decoder head, on every dataset, is scored at its final checkpoint, with no
early stopping and no checkpoint selection. The $T$ feature taps are blocks 0, 6, 12,
17 and 23 of 24. The multi-layer attention probe projects each tap to 256 channels, concatenates them and pools them
with gated attention, followed by one linear output per target. The linear and attention classification probes are
trained with three seeds. The training-free $k$-NN classifier uses $k{=}20$ and weights each neighbor by
$\exp(s/0.07)$, where $s$ is the cosine similarity.

\paragraph{Statistics.}
Splits and seeds are fixed per result family. Segmentation tables, ISLES'24 included, report means over three folds,
except TotalSegmentator CT~\cite{totalseg}, which uses its official 34-case test split, and a segmentation point estimate
averages each case's Dice across the three head-training seeds, then over cases. Classification uses one fixed
patient-level split per dataset and averages AUROC over the three probe seeds. PET pools the five autoPET folds, the
stage ablation and the k-means probe use one fold, and the label-efficiency sweeps report means and SDs over three
folds of their own splits. Every 95\% interval runs from the 2.5th to the 97.5th percentile of a bootstrap. For
segmentation and classification, each draw resamples cases or patients with replacement within folds and samples one
of the three seeds independently for each model. A paired comparison uses the same resampled cases for both models and
records their difference, \ours minus baseline. The PET table of paired differences (\cref{tab:supp-pet-froc}) writes
it as encoder minus \ours, so a negative $\Delta$ there favors \ours.

\paragraph{Significance.}
Three rules decide what the text calls significant, level or tied. A difference is significant when its paired 95\%
interval excludes 0, and $\ast$ marks it in the tables, whichever way the difference is written. In the segmentation
and ISLES'24 tables (\cref{tab:mri-seg,tab:ct-seg,tab:isles}), the text calls an encoder level with the best when it is
not significantly behind the best after Holm correction over the table's comparisons, and \cref{tab:supp-anat-maps} bolds a per-structure $\Delta$ that stays significant after Holm correction within
its panel. The win/tie/loss (W/T/L) counts of \cref{tab:supp-rank-robust} call a cell tied when the 95\% intervals of
the two encoders overlap, and on registration they use the paired intervals of \cref{tab:supp-reg}. The win shares of
\cref{tab:supp-lesion,tab:supp-kmeans} count the cases on which \ours scores higher than \nnfvit, with no test. The
bootstrap uses
2,000 draws, except the paired tests of the segmentation, label-efficiency, ISLES'24 and registration tables, which use
10,000, the paired PET differences, which use 4,000 for Dice and F1 and 2,000 for detection, and the dataset bootstrap
of \cref{tab:supp-rank-robust}, which uses 5,000.

\paragraph{Lesion-anomaly probe.}
The probe reads each model on a common 8\,mm grid with one cell-centered readout, and the anomaly score is 1 minus the
mean cosine similarity to the 5 nearest bank tokens. On MRI, the bank holds the lesion-free tissue of one reference
patient, and each value of \cref{tab:supp-lesion} averages three reference patients, the first three training patients
with a lesion. TriALS~\cite{data_trials} uses 5 folds in this probe, and TriALS structure labels are
TotalSegmentator~\cite{totalseg} predictions.

\paragraph{Intensity normalization.}
Intensity rules depend on the modality, in pretraining and in evaluation. In pretraining, MRI is clipped at its 1st and
99th percentiles (0.1st and 99.9th from S2 on) and rescaled to $[0,1]$. CT is resampled to 2\,mm, clamped below at
$-1024$\,HU and rescaled to $[0,1]$ between its 0.01st and 99.99th percentiles. In evaluation, each encoder reads its
input under its own normalization. \ours rescales MRI to $[0,1]$ between the 1st and 99.9th percentiles of the non-zero
voxels, and CT between the 1st and 99.9th percentiles of the whole image. The baselines use z-scoring for
nnFoundation~\cite{nnfoundation}, percentiles 0.5 to 99.5 mapped to $[0,1]$ for the four Triad encoders~\cite{triad}, $[-1000, 1000]$\,HU
mapped to $[0,1]$ for Merlin~\cite{merlin} and SPECTRE~\cite{spectre}, clipping to $[-1000,1900]$\,HU and a global
z-score for VoxelFM~\cite{voxelfm}, and per-image percentile clipping and z-scoring for CT-Instruct~\cite{ctinstruct}.
Curia-MAE~\cite{curiamae}, scored on PET only, takes the per-image z-score of its pretraining (\cref{sec:supp-pet}).

\paragraph{Input spacing.}
The encoders were pretrained under different preprocessing, so each has seen a different range of voxel spacings, and
no single grid suits all of them (\cref{tab:supp-geometry}). For each segmentation dataset we start from nnFoundation's
native spacing and clip each axis ($z$, $y$, $x$) into the encoder's pretraining range, which gives the grid nearest
the native spacing that the encoder has seen. An axis coarser than the range's maximum is upsampled, as for thick
slices, and an axis finer than its minimum is downsampled. VoxelFM and the CT side of \ours saw only isotropic views
and read an isotropic grid. Merlin was pretrained on one fixed grid and reads it on every segmentation dataset, and nnFoundation and
CT-Instruct, pretrained at native spacing, read the native grid. The ranges differ widely: \ours spans
0.15 to 2.5\,mm along $z$ on MRI, reflecting the variety of MRI acquisitions, but only 0.8 to 2.3\,mm on CT, close to
SPECTRE's range along $z$. Fairness therefore comes from matching each encoder's downstream input to its own
pretraining, and every choice is fixed per dataset before any head is trained.

\begin{table}[!htbp]
\centering\small\setlength{\tabcolsep}{6pt}\renewcommand{\arraystretch}{1.12}
\caption{\textbf{Segmentation input spacing range per encoder.} $z$ and $y$, $x$: the approximate voxel spacing (mm) of the
views each encoder was pretrained on, as implied by its preprocessing (resampling, crop-scale and resize ranges).
Downstream, each axis of the native spacing is clipped into this range, and grid gives the resulting input grid.}
\label{tab:supp-geometry}
\begin{tabular}{@{}l l c c l@{}}
\toprule
encoder & pretraining input & $z$ & $y$, $x$ & grid \\
\midrule
\multicolumn{5}{@{}l}{\textbf{CT}} \\
\cmidrule(r){1-1}
nnF-ViT, nnF-CNN        & native spacing, $192^3$ crops            & 0.5--5   & 0.5--1.0 & anisotropic \\
SPECTRE                 & resampled to $1.0{\times}0.5{\times}0.5$, rescaled crops & 1.0--2.0 & 0.5--1.0 & anisotropic \\
VoxelFM                 & rescaled crops & 0.7--5 & 0.7--5 & isotropic \\
Merlin                  & resampled to $3.0{\times}1.5{\times}1.5$  & 3.0      & 1.5      & anisotropic \\
CT-Instruct             & native spacing, $128^3$ crops            & 0.5--5   & 0.5--1.0 & anisotropic \\
\ours                   & 2\,mm, foreground fill, rescaled crops   & 0.8--2.3 & 0.8--2.3 & isotropic \\
\midrule
\multicolumn{5}{@{}l}{\textbf{MRI}} \\
\cmidrule(r){1-1}
nnF-ViT, nnF-CNN        & native spacing, $192^3$ crops            & 0.5--6   & 0.3--1.5 & anisotropic \\
Triad (all four)        & foreground resized to $96^3$             & 0.6--4   & 0.6--4   & anisotropic \\
\ours                   & native, foreground fill, rescaled crops & 0.15--2.5 & 0.03--2.5 & anisotropic \\
\bottomrule
\end{tabular}
\end{table}

\FloatBarrier
\section{Evaluation datasets}
\label{sec:supp-datasets}
This section defines the benchmarks and the exposure rules that the later sections use. \Cref{tab:datasets} lists
every evaluation dataset with its modality, its target and the experiments that read it. The paragraphs below set the
exposure audit behind the $\dagger$ and $\checkmark$ marks, the excluded BraTS21 set and the single-run stages, and name
the results where other encoders lead.

\paragraph{Datasets.} The benchmark holds nine MRI and seven CT segmentation sets, two MRI and two CT classification
sets, the contrast shift of ISLES'24 from CT angiography to non-contrast CT, and held-out fluorodeoxyglucose (FDG) PET on
autoPET (\maincref{sec:protocol}). None of these datasets is in the pretraining data of \ours (\cref{sec:supp-stages}).
Classification settings per dataset are in \cref{sec:supp-cls}. On autoPET, every
encoder is scored in five-fold cross-validation over all 1,014 studies, and the four CNNs use a UNet3D head matched to
UNETR, whose size and deepest stride differ from their standard decoders (\cref{sec:supp-pet}). ATLAS 2023 and STOIC
2021 are called ATLAS and STOIC elsewhere.

\begin{table}[!htbp]
\centering\footnotesize\setlength{\tabcolsep}{3pt}
\caption{\textbf{Evaluation datasets} and the experiments that use them. Seg.: segmentation (Dice). Cls.:
classification (AUROC). Det.: detection, by the competition performance metric (CPM; \cref{tab:isles,tab:supp-pet-froc}).
LE: label efficiency. Anom.:
lesion-anomaly probe. k-means: anatomy clustering with no head (\cref{tab:supp-kmeans}).
Rob.: intensity robustness. Reg.: MRI-to-CT registration. Corr.:
cross-contrast correspondence (\cref{tab:supp-correspondence}). NCCT: non-contrast CT. CTA: CT angiography. LVO:
large-vessel occlusion. IDH: isocitrate dehydrogenase.}
\label{tab:datasets}
\resizebox{\ifdim\width>\linewidth\linewidth\else\width\fi}{!}{%
\begin{tabular}{@{}l l l l@{}}
\toprule
dataset & modality & target & used in \\
\midrule
Prostate158~\cite{data_prostate158} & MRI & prostate zones & Seg., k-means \\
ACDC~\cite{data_acdc} & cine MRI & cardiac structures & Seg., LE, k-means \\
ATLAS 2023~\cite{data_atlas23} & MRI & liver, liver tumor & Seg., k-means \\
AMOS-MRI~\cite{amos} & MRI & abdominal organs & Seg., k-means \\
MM-WHS~\cite{data_mmwhs} & MRI & whole heart & Seg. \\
MSD Hippocampus~\cite{data_msd} & MRI & hippocampus & Seg., LE \\
TotalSeg-MRI~\cite{dantonoli2025totalsegmri} & MRI & body structures & Seg., Rob., Reg. \\
BraTS-Africa~\cite{data_bratsafrica} & MRI & glioma & Seg., Anom., Corr. \\
ISLES'22~\cite{data_isles22} & MRI & ischemic stroke & Seg., Anom. \\
UCSF-PDGM~\cite{data_ucsfpdgm} & MRI & IDH status, WHO grade & Cls. \\
LLD-MMRI~\cite{data_lldmmri} & MRI, 8 phases & 7 liver lesion types & Cls. \\
\midrule
AMOS22~\cite{amos} & CT & abdominal organs & Seg., Rob., k-means \\
MSD Pancreas, Colon~\cite{data_msd} & CT & pancreas and tumor; colon tumor & Seg. \\
TotalSegmentator~\cite{totalseg} & CT & 20 body structures & Seg., LE \\
AISD~\cite{data_aisd} & non-contrast CT & ischemic stroke & Seg., LE \\
TriALS~\cite{data_trials} & multi-phase CT & liver tumor & Seg., Anom., Reg. \\
ImageCAS~\cite{data_imagecas} & CT angiography & coronary arteries & Seg. \\
ISLES'24~\cite{data_isles24} & NCCT and CTA & circle of Willis, LVO & Seg., Det., LE, Corr. \\
STOIC 2021~\cite{data_stoic} & CT & COVID-19, severity & Cls. \\
RSNA 2023~\cite{data_rsna23} & CT & 6 abdominal trauma targets & Cls. \\
\midrule
autoPET~\cite{autopet} & FDG PET & FDG-avid lesions & Seg., Rob., Det. \\
\bottomrule
\end{tabular}}
\end{table}

\paragraph{Exposure audit.}\label{sec:supp-scope}
We audit the pretraining data of every encoder, ours included, against our test sets as far as each publication
documents its sources, and SPECTRE~\cite{spectre} also image by image. Where a test set is in an encoder's pretraining
data, the tables mark the encoder with $\dagger$ and still report its numbers. The exposure is to the test images, and
for CT-Instruct also to the segmentation labels (\cref{tab:ct-seg}). Of nnFoundation's~\cite{nnfoundation} 2.1M
volumes, about 1.3M are private and cannot be checked. On PET, $\checkmark$ marks the encoders pretrained on PET, the two
nnFoundation models (\cref{tab:pet}).

\paragraph{Exclusions.} We do not evaluate on BraTS21, because about a third of its test cases are
UPENN-GBM~\cite{bakas2022upenngbm} patients, who are in the pretraining data of \ours and in a source nnFoundation lists.

\paragraph{Single-run stages.} Each pretraining stage is one run that continues the previous one
(\maincref{sec:le-stages}), and \cref{sec:supp-stage-ablation} scores the end of each stage.

\paragraph{Where other encoders lead.} Other encoders lead in places, and the tables report each case. Among them,
SPECTRE leads AISD at 5 and 10\% of the labels (\cref{tab:supp-le}) and large-vessel occlusion detection from
non-contrast CT, within overlapping intervals (\cref{tab:isles}), and SPECTRE and \nnfcnn keep more token similarity
under Gaussian noise on CT (\cref{tab:supp-robust}).

\FloatBarrier
\section{Pretraining data and stages}
\label{sec:supp-stages}
This section gives the pretraining corpus and the rule that admits a dataset to it, then the three training stages
and every pretraining setting. \Maincref{fig:method} shows the corpus by modality and anatomy.

\paragraph{Corpus.}
The corpus holds 234,919 public, license-vetted adult volumes, 63\% of them MRI (\cref{tab:supp-corpus}). They come
from The Cancer Imaging Archive (TCIA)~\cite{clark2013tcia}, whose collections include the National Lung Screening Trial
(NLST)~\cite{nlst2011}, and from OpenNeuro~\cite{markiewicz2021openneuro}. Collections that also hold PET contribute only their MRI or CT volumes, so
the corpus holds no PET.

\begin{table}[!htbp]
  \centering
  \footnotesize
  \setlength{\tabcolsep}{3pt}
  \caption{\textbf{Pretraining corpus of \ours.} Subjects are counted within each collection, so a person in two
  collections counts twice.}
  \label{tab:supp-corpus}
  \resizebox{\ifdim\width>\linewidth\linewidth\else\width\fi}{!}{%
\begin{tabular}{@{}lrrl@{}}
    \toprule
    Modality & Volumes & Subjects & Sources \\
    \midrule
    MRI & 148,831 & 41,192 & 21 TCIA collections, 895 OpenNeuro datasets \\
    CT  & 86,088  & 29,777     & 49 TCIA collections, NLST among them \\
    \midrule
    Total & 234,919 & & \\
    \bottomrule
  \end{tabular}}
\end{table}

\paragraph{Collections.}
\Cref{tab:supp-corpus-mri} lists the MRI side: the 21 TCIA collections one by one, mostly breast and prostate, and the
895 OpenNeuro datasets of the brain, head and neck, grouped by license.

\begin{table}[!htb]
  \centering
  \footnotesize
  \setlength{\tabcolsep}{3pt}
  \caption{\textbf{Pretraining corpus, MRI.} TCIA volumes cover the breast (52,550), the prostate (22,813), the brain
  (4,265) and the uterus (1,080).}
  \label{tab:supp-corpus-mri}
  \resizebox{\ifdim\width>\linewidth\linewidth\else\width\fi}{!}{%
\begin{tabular}{@{}llrr@{\hspace{2em}}llrr@{}}
    \toprule
    Dataset & License & Volumes & Subjects & Dataset & License & Volumes & Subjects \\
    \midrule
    \multicolumn{8}{@{}l}{\textit{TCIA, 21 collections: 80,708 volumes, 5,840 subjects}} \\
    PROSTATEx & CC BY 3.0 & 18,743 & 346 & Breast-MRI-NACT-Pilot & CC BY 3.0 & 917 & 64 \\
    ISPY2 & CC BY 4.0 & 13,731 & 398 & ReMIND & CC BY 4.0 & 670 & 114 \\
    ACRIN-6698 & CC BY 4.0 & 12,803 & 385 & QIN Breast DCE-MRI & CC BY 3.0 & 632 & 10 \\
    ACRIN-Contralateral-Breast-MR & CC BY 4.0 & 8,118 & 980 & Vestibular-Schwannoma-SEG & CC BY 4.0 & 476 & 242 \\
    Advanced-MRI-Breast-Lesions & CC BY 4.0 & 6,039 & 632 & PROSTATE-DIAGNOSIS & CC BY 3.0 & 368 & 92 \\
    EA1141 & CC BY 4.0 & 4,560 & 500 & Prostate Fused-MRI-Pathology & CC BY 3.0 & 289 & 28 \\
    ISPY1 & CC BY 3.0 & 4,071 & 222 & BREAST-DIAGNOSIS & CC BY 3.0 & 237 & 84 \\
    UPENN-GBM & CC BY 4.0 & 3,119 & 628 & TCGA-PRAD & CC BY 3.0 & 162 & 10 \\
    Prostate-MRI-US-Biopsy & CC BY 4.0 & 3,057 & 842 & PROSTATE-MRI & CC BY 3.0 & 130 & 26 \\
    TCGA-BRCA & CC BY 3.0 & 1,442 & 137 & Prostate-3T & CC BY 3.0 & 64 & 64 \\
    CPTAC-UCEC & CC BY 4.0 & 1,080 & 36 &  & & &  \\
    \midrule
    \multicolumn{8}{@{}l}{\textit{OpenNeuro, 895 datasets: 68,123 volumes, 35,352 subjects}} \\
    \multicolumn{2}{@{}l}{848 datasets, CC0 1.0} & 66,397 & 34,061 & \multicolumn{2}{@{}l}{3 datasets, Public domain} & 37 & 37 \\
    \multicolumn{2}{@{}l}{31 datasets, PDDL 1.0} & 1,247 & 1,002 & \multicolumn{2}{@{}l}{2 datasets, CC BY (version not stated)} & 36 & 6 \\
    \multicolumn{2}{@{}l}{11 datasets, CC BY 4.0} & 406 & 246 &  &  \\
    \midrule
    \multicolumn{8}{@{}c@{}}{Total: 916 datasets, 148,831 volumes, 41,192 subjects} \\
    \bottomrule
  \end{tabular}}
\end{table}

\Cref{tab:supp-corpus-ct} lists the 49 TCIA collections of the CT side, all under CC~BY 3.0 or 4.0.
\Cref{sec:supp-attribution} gives the attribution and DOI of every collection.

\begin{table}[!htb]
  \centering
  \footnotesize
  \setlength{\tabcolsep}{3pt}
  \caption{\textbf{Pretraining corpus, CT.} NLST holds 72,804 of the 86,088 volumes (84.6\%).}
  \label{tab:supp-corpus-ct}
  \resizebox{\ifdim\width>\linewidth\linewidth\else\width\fi}{!}{%
\begin{tabular}{@{}llrr@{\hspace{2em}}llrr@{}}
    \toprule
    Collection & License & Volumes & Subjects & Collection & License & Volumes & Subjects \\
    \midrule
    NLST & CC BY 4.0 & 72,804 & 26,229 & RIDER Pilot & CC BY 4.0 & 127 & 8 \\
    COVID-19-NY-SBU & CC BY 4.0 & 2,774 & 457 & ADRENAL-ACC-KI67-SEG & CC BY 4.0 & 119 & 53 \\
    TCGA-KIRC & CC BY 3.0 & 1,153 & 229 & Pelvic-Reference-Data & CC BY 3.0 & 116 & 58 \\
    VAREPOP-APOLLO & CC BY 4.0 & 932 & 38 & QIN-BREAST & CC BY 3.0 & 110 & 43 \\
    CPTAC-UCEC & CC BY 4.0 & 700 & 75 & CT-VS-PET-VENTILATION-IMAGING & CC BY 4.0 & 99 & 20 \\
    HCC-TACE-SEG & CC BY 4.0 & 561 & 105 & TCGA-ESCA & CC BY 3.0 & 98 & 16 \\
    MEDIASTINAL-LYMPH-NODE-SEG & CC BY 4.0 & 513 & 513 & CPTAC-SAR & CC BY 4.0 & 84 & 18 \\
    CPTAC-PDA & CC BY 4.0 & 505 & 94 & CPTAC-STAD & CC BY 4.0 & 80 & 19 \\
    TCGA-OV & CC BY 3.0 & 481 & 132 & Pancreas-CT & CC BY 3.0 & 80 & 80 \\
    TCGA-UCEC & CC BY 3.0 & 477 & 65 & CTPRED-SUNITINIB-PANNET & CC BY 4.0 & 76 & 38 \\
    CPTAC-CCRCC & CC BY 4.0 & 476 & 64 & NaF PROSTATE & CC BY 3.0 & 72 & 9 \\
    ANTI-PD-1\_LUNG & CC BY 3.0 & 431 & 46 & TCGA-KIRP & CC BY 3.0 & 72 & 23 \\
    SPINAL-MULTIPLE-MYELOMA-SEG & CC BY 4.0 & 360 & 67 & SPIE-AAPM-LUNG-CT-CHALLENGE & CC BY 3.0 & 70 & 70 \\
    CPTAC-LUAD & CC BY 4.0 & 258 & 36 & CC-TUMOR-HETEROGENEITY & CC BY 4.0 & 65 & 22 \\
    CMB-OV & CC BY 4.0 & 249 & 16 & CPTAC-CM & CC BY 4.0 & 61 & 12 \\
    TCGA-STAD & CC BY 3.0 & 248 & 46 & LUNGCT-DIAGNOSIS & CC BY 3.0 & 58 & 58 \\
    STAGEII-COLORECTAL-CT & CC BY 4.0 & 225 & 225 & SPINE-METS-CT-SEG & CC BY 4.0 & 56 & 55 \\
    TCGA-LUAD & CC BY 3.0 & 209 & 53 & TCGA-COAD & CC BY 3.0 & 51 & 24 \\
    PANCREATIC-CT-CBCT-SEG & CC BY 4.0 & 198 & 40 & QIN LUNG CT & CC BY 3.0 & 46 & 46 \\
    CT-LYMPH-NODES & CC BY 3.0 & 176 & 176 & CPTAC-AML & CC BY 4.0 & 31 & 5 \\
    RIDER-LUNG-CT & CC BY 3.0 & 171 & 32 & BREAST-DIAGNOSIS & CC BY 3.0 & 18 & 3 \\
    COLORECTAL-LIVER-METASTASES & CC BY 4.0 & 160 & 160 & TCGA-PRAD & CC BY 3.0 & 14 & 4 \\
    COVID-19-AR & CC BY 4.0 & 148 & 23 & EA1141 & CC BY 4.0 & 4 & 1 \\
    TCGA-LUSC & CC BY 3.0 & 137 & 37 & TCGA-READ & CC BY 3.0 & 4 & 3 \\
    PROSTATE-ANATOMICAL-EDGE-CASES & CC BY 4.0 & 131 & 131 &  & & &  \\
    \midrule
    \multicolumn{8}{@{}c@{}}{CC BY 4.0: 27 collections, 81,796 volumes, 28,524 subjects\quad CC BY 3.0: 22 collections, 4,292 volumes, 1,253 subjects} \\
    \multicolumn{8}{@{}c@{}}{Total: 49 collections, 86,088 volumes, 29,777 subjects} \\
    \bottomrule
  \end{tabular}}
\end{table}

\paragraph{Inclusion rule.}
A dataset is included when its license permits commercial use and the distribution of derived works, and no
source states a restrictive term. Restrictive terms include NonCommercial, NoDerivatives, ShareAlike and copyleft
licenses, data-use agreements, registration requirements, research-only use and limited access. We read the license in
the archive (TCIA, OpenNeuro, and for CT also the NCI Imaging Data Commons, IDC, and DataCite) and in each dataset's own
files, and the most restrictive statement governs. Reading the files excludes 10 OpenNeuro datasets and 3 CT
collections whose metadata state a permissive license. We also exclude datasets with participants under 18 and every
evaluation dataset. The rule acts on whole datasets, and UPENN-GBM, which it admits, shares patients with BraTS21, so we
do not evaluate on BraTS21 (\cref{sec:supp-scope}). All remaining datasets are CC0, the Open Data Commons Public Domain
Dedication and License (PDDL) or public domain, or CC~BY 3.0 or 4.0, and two OpenNeuro datasets state CC~BY without a
version.

\paragraph{Stages.}
Pretraining runs in three stages, and each changes the crop size or the data mix (\cref{tab:pretrain-stages}). Each
stage starts from the previous stage's student, teacher and heads, with a new optimizer state and schedules. S1 trains
on MRI at $128^3$. S2 trains on MRI at higher resolution: its first phase runs 40 passes at $128^3$ and then 32 at
$160^3$, the latter over the 113,970 MRI volumes of its $160^3$ list, and its second phase alternates the two crop
sizes, so that the volumes too small for $160^3$ stay in training. S3 trains on MRI and CT at $160^3$, then on CT
alone. We end this CT-only phase after 18 of its 24 planned passes, once the training loss stops changing, and its last checkpoint is the evaluated model.

\begin{table}[!htbp]
  \centering
  \footnotesize
  \setlength{\tabcolsep}{3pt}
  \caption{\textbf{Pretraining stages of \ours.} The evaluated model is the end of S3. A paired MRI count gives the
  volumes on the $128^3$\,/\,$160^3$ lists. Passes: $40 + 32$ is 40 passes at $128^3$ and then 32 at $160^3$, and
  $4 \times 12$ is four phases of 12 passes that alternate the two crop sizes.}
  \label{tab:pretrain-stages}
  \resizebox{\ifdim\width>\linewidth\linewidth\else\width\fi}{!}{%
  \begin{tabular}{@{}clcc@{}}
    \toprule
    Stage & Data (volumes) & Crop & Passes \\
    \midrule
    S1 & MRI 148,831 & $128^3$ & 250 \\
    \addlinespace[2pt]
    S2 & (i) MRI 138,109 / 113,970 & $128^3$, then $160^3$ & 40 + 32 \\
       & (ii) MRI 138,109 / 113,970 & $128^3$, $160^3$ alternating & $4 \times 12$ \\
    \addlinespace[2pt]
    S3 & (i) MRI 113,970 + CT 86,088 & $160^3$ & 24 \\
       & (ii) CT 86,088 & $160^3$ & 18 of 24 \\
    \bottomrule
  \end{tabular}}
\end{table}

The $128^3$ and $160^3$ lists of S2 and S3 hold the
volumes whose shortest foreground axis reaches the crop size, plus thin volumes, under 96 voxels on that axis, which are
padded. A pass draws as many crops as its list holds entries, and at $128^3$ in S2 each draw takes two crops of one
volume. Each volume is one entry, except in the mixed phase of S3, whose list counts each CT volume 1.5 times: 113,970
MRI and 129,132 CT entries, 243,102 in all. Which volumes a pass draws follows the sampling weights of
\cref{tab:supp-pretrain-hparams}. \Cref{sec:supp-stage-ablation} scores the end of each stage, and S3 gives the most well-rounded checkpoint.

\begin{table}[!tb]
\centering\scriptsize\setlength{\tabcolsep}{3pt}\renewcommand{\arraystretch}{0.9}
\newcolumntype{Q}[1]{>{\raggedright\arraybackslash}p{#1}}
\newcolumntype{W}[1]{>{\centering\arraybackslash}p{#1}}
\newcolumntype{Y}{!{\color{black!25}\vrule}}
\newcommand{\all}[1]{\multicolumn{5}{Q{14.25cm}}{#1}}
\newcommand{\gap}{\addlinespace[0.5pt]}
\newcommand{\shared}{\addlinespace[1.5pt]\cmidrule[0.3pt](l){2-6}\addlinespace[1pt]}
\caption{\textbf{Pretraining settings of \ours,} one column per phase of \cref{tab:pretrain-stages}. In each group,
per-phase settings come first and settings shared by every phase follow the thin rule. $a$\,/\,$b$: value at
$128^3$\,/\,$160^3$. $B$: crops per optimizer step.}
\label{tab:supp-pretrain-hparams}
\begin{tabular}{@{}Q{2.55cm}Y W{2.7cm}Y W{2.7cm} W{2.7cm}Y W{2.7cm} W{2.7cm}@{}}
\toprule
 & S1 & \multicolumn{2}{cY}{S2} & \multicolumn{2}{c}{S3} \\
\cmidrule(lr){3-4}\cmidrule(l){5-6}
phase & MRI & (i) MRI & (ii) MRI & (i) MRI and CT & (ii) CT \\
\midrule
\multicolumn{6}{@{}l}{\textbf{Optimization}} \\
optimizer steps & 36,250 & 12,536 & 8,592 & 11,376 & 3,024 \\ \gap
base learning rate & \multicolumn{3}{W{8.5cm}Y}{$2\times10^{-3}$} & \multicolumn{2}{W{5.6cm}}{$10^{-3}$} \\ \gap
linear warmup & 10 passes & 1 pass per crop size & $\tfrac14$ pass per phase & 1 pass & 1 pass \\ \gap
learning-rate schedule & constant & constant\,/\,cosine to $10^{-6}$ & one cosine across the four phases, ending at half
the peak & cosine to $10^{-6}$ & cosine over 24 planned passes, ending near $10^{-4}$ \\ \gap
last-layer freeze & 1 pass & \multicolumn{4}{W{11.4cm}}{none} \\
\shared
GPUs & \all{64 H200} \\ \gap
crops per step ($B$) & \all{1,024\,/\,512 (at $128^3$, S1 takes one crop of each of 1,024 volumes and S2 two crops of
each of 512)} \\ \gap
peak learning rate & \all{base rate $\times\sqrt{B/1024}$:\quad $2.0\times10^{-3}$\,/\,$1.41\times10^{-3}$ in S1 and
S2,\quad $7.1\times10^{-4}$ in S3} \\ \gap
optimizer & \all{AdamW ($\beta$ 0.9, 0.999), weight decay 0.04 (constant, none on biases, norms, LayerScale, the
position table and the class and mask tokens) and layer-wise learning-rate decay 0.9} \\ \gap
regularization & \all{stochastic depth 0.2, gradient-norm clipping at 3.0 and bf16 mixed precision} \\
\midrule
\multicolumn{6}{@{}l}{\textbf{Self-distillation}} \\
teacher EMA (per step) & 0.99 & 0.99\,/\,cosine from 0.995 to 1 & cosine toward 1, ending at 0.9975 & cosine from 0.995
to 1 & cosine from 0.995 to 1, ending at 0.9993 \\ \gap
teacher temperature & 0.04 to 0.07 over 30 passes & \multicolumn{4}{W{11.4cm}}{0.07} \\ \gap
Gram anchor & snapshot at pass 50, active from 52 &
\multicolumn{4}{W{11.4cm}}{new teacher snapshot at pass 1 of each phase, active from pass 2} \\
\shared
student temp. & \all{0.1, with softmax centering of the teacher (momentum 0.9) on both heads} \\ \gap
heads & \all{separate DINO and iBOT heads: MLP 1152--2048--2048--256, $\ell_2$ normalization, weight-normalized layer
to 16,384 prototypes} \\ \gap
loss weights & \all{DINO 1, iBOT 1 and KoLeo 0.1, with Gram set to 2\% of the total loss when its anchor activates and about 0.02\% over most of S1} \\ \gap
Gram tokens & \all{512 random visible patch tokens per global crop, after the final LayerNorm, with a mean squared error
between the cosine-similarity matrices of student and anchor} \\ \gap
iBOT masking & \all{half of the global crops, with the masked share drawn from $U[0.1, 0.5]$ and blocks of up to 3\,/\,4 tokens a
side} \\
\midrule
\multicolumn{6}{@{}l}{\textbf{Views}} \\
volume crop & $128^3$ & $128^3$, then $160^3$ & $128^3$ and $160^3$ & \multicolumn{2}{W{5.6cm}}{$160^3$} \\ \gap
GIN on MRI & \multicolumn{3}{W{8.5cm}Y}{$p$ 0.3, $\alpha < 0.15$} & \multicolumn{2}{W{5.6cm}}{none} \\ \gap
CT augmentation & \multicolumn{3}{W{8.5cm}Y}{--} & \multicolumn{2}{Q{5.6cm}}{affine, flips, histogram shift and
smoothing as on MRI, Gaussian noise ($\sigma$ 0.004, $p$ 0.5), in-plane low-resolution zoom (0.85--1, $p$ 0.25), local
crops at half strength and no inversion} \\
\shared
volume-crop rule & \all{axes thinner than the crop are padded, then the volume is resized down so its shortest
foreground axis fills the crop, at the best of 8 random offsets with at least 50\% tissue} \\ \gap
global crops & \all{2 per volume crop: a box of 0.4--0.9 of each axis, resized to the volume-crop size} \\ \gap
local crops & \all{4 per volume crop: a box of 0.2--0.5 of each axis, resized to $24^3$\,/\,$32^3$} \\ \gap
MRI augmentation & \all{on every crop: affine ($p$ 0.5, rotation up to $10^\circ$, shear 0.1), flips on two
axes ($p$ 0.2), histogram shift ($p$ 0.5, 10 control points) and Gaussian smoothing ($\sigma$ 0.25--1.5, $p$ 0.5, or 0.1 on the
second global crop). On the second global crop, inversion ($p$ 0.2)} \\
\midrule
\multicolumn{6}{@{}l}{\textbf{Data and checkpoints}} \\
sampling & \multicolumn{3}{W{8.5cm}Y}{per volume, with replacement, and OpenNeuro volumes 1.5 times as often as TCIA} & CT and
OpenNeuro 1.5$\times$ other MRI (CT 47.4\% of draws) & uniform \\
\shared
volume & \all{each series is one single-channel 3D volume, so every MRI sequence and every CT phase is its own volume.
4D diffusion series are split into 3D volumes, and multi-channel volumes are left out} \\ \gap
checkpoint & \all{the last checkpoint of each phase starts the next. The CT-only phase ends after 18 of its 24 planned
passes, once the training loss stops changing, and its last checkpoint is the evaluated model} \\
\bottomrule
\end{tabular}
\end{table}

\paragraph{Settings.}
\Cref{tab:supp-pretrain-hparams} gives every pretraining setting, one column per phase. The phases share the heads,
losses, crop rules and optimizer, and differ in schedule, teacher EMA and temperature, Gram anchor, crop size, augmentation
and sampling. Every phase with MRI draws OpenNeuro volumes 1.5 times as often as TCIA MRI volumes, and the mixed phase
of S3 draws CT at the same rate, so that CT makes up 47\% of its draws. The CT-only phase draws uniformly. The
augmentation of every crop, MRI or CT, includes a histogram shift ($p$ 0.5) in all three stages, and MRI also receives
global intensity non-linearity (GIN) augmentation~\cite{gin} ($p$ 0.3) in S1 and S2.

\FloatBarrier
\section{Pretraining-stage ablation}
\label{sec:supp-stage-ablation}
\Cref{tab:supp-le-stages} scores the checkpoint at the end of each pretraining stage on every task family, and S3, which
sees the most diverse data, is the most well-rounded: it leads all four summary rows at the bottom of the table. The stages need not improve every task in turn, since each changes the data, and
the two gray columns give what each stage adds.

\begin{table}[!htbp]
\centering\footnotesize\setlength{\tabcolsep}{3.5pt}
\caption{\textbf{Pretraining-stage ablation.} Frozen checkpoint at the end of each stage, with UNETR and probe heads.
Brackets: 95\% bootstrap interval. Gap: points by which a stage trails the best of the three stages in
a row (0 when it is the best), averaged over the 20 rows or at its largest. Mean score: mean of the 20 rows, all on a 0--100 scale.}
\label{tab:supp-le-stages}
\begin{tabular}{@{}l c@{\hspace{2pt}}c@{\hspace{2pt}}c c@{\hspace{2pt}}c@{\hspace{2pt}}c@{}}
\toprule
 & \makecell{S1\\[-2pt]{\tiny MRI, $128^3$}} & \makecell{\color{black!45}S2$-$S1\\[-2pt]{\tiny\color{black!45} high-res MRI adds}} & \makecell{S2\\[-2pt]{\tiny high-res MRI}} & \makecell{\color{black!45}S3$-$S2\\[-2pt]{\tiny\color{black!45} CT adds}} & \makecell{S3\\[-2pt]{\tiny MRI $+$ CT}} \\
\midrule
\multicolumn{6}{@{}l}{\textit{Segmentation, Dice \% (UNETR)}} \\
MRI, 7 datasets & 82.4 & {\color{black!45}$+$0.6} & 83.0 & {\color{black!45}$-$0.1} & 82.9 \\
CT, 5 datasets & 67.8 & {\color{black!45}$+$1.7} & 69.5 & {\color{black!45}$+$0.4} & 69.9 \\
ISLES'24 circle of Willis, NCCT & 49.5 & {\color{black!45}$+$0.8} & 50.3 & {\color{black!45}$+$0.1} & 50.4 \\
ISLES'24 LVO region, CTA and NCCT & 14.0 & {\color{black!45}$-$6.4} & 7.6 & {\color{black!45}$+$3.9} & 11.5 \\
\addlinespace[2pt]
\multicolumn{6}{@{}l}{\textit{FDG PET lesions (UNETR, mean $\pm$ SD over head runs)}} \\
lesion F1 & 45.9\,{\tiny$\pm$3.1} & {\color{black!45}$+$2.9} & 48.8\,{\tiny$\pm$5.2} & {\color{black!45}$+$1.1} & 49.9\,{\tiny$\pm$1.4} \\
lesion-positive Dice & 64.2\,{\tiny$\pm$0.9} & {\color{black!45}$+$2.1} & 66.3\,{\tiny$\pm$0.4} & {\color{black!45}$-$0.8} & 65.5\,{\tiny$\pm$0.5} \\
\addlinespace[2pt]
\multicolumn{6}{@{}l}{\textit{Classification, AUROC \% (mean over targets)}} \\
MRI, k-NN & 83.9 & {\color{black!45}$-$0.7} & 83.2 & {\color{black!45}$+$0.6} & 83.8 \\
MRI, linear probe & 92.3 & {\color{black!45}$-$0.8} & 91.5 & {\color{black!45}$+$2.4} & 93.9 \\
MRI, attention probe & 92.4 & {\color{black!45}$+$0.8} & 93.2 & {\color{black!45}$-$0.3} & 92.9 \\
STOIC, linear probe & 75.2 & {\color{black!45}$-$1.5} & 73.7 & {\color{black!45}$+$3.6} & 77.3 \\
STOIC, attention probe & 80.8 & {\color{black!45}$+$1.1} & 81.9 & {\color{black!45}$-$0.5} & 81.4 \\
\addlinespace[2pt]
\multicolumn{6}{@{}l}{\textit{MRI-to-CT registration, Dice \%, $+$ MIND-SSC}} \\
label-free start & \makecell{30.8\\[-2pt]{\tiny[23.5, 38.5]}} & {\color{black!45}$+$2.6} & \makecell{33.4\\[-2pt]{\tiny[25.7, 41.4]}} & {\color{black!45}$-$0.1} & \makecell{33.3\\[-2pt]{\tiny[25.4, 41.2]}} \\
landmark start & \makecell{53.3\\[-2pt]{\tiny[47.4, 58.8]}} & {\color{black!45}$+$0.9} & \makecell{54.2\\[-2pt]{\tiny[48.2, 59.8]}} & {\color{black!45}$+$0.5} & \makecell{54.7\\[-2pt]{\tiny[48.9, 60.1]}} \\
\addlinespace[2pt]
\multicolumn{6}{@{}l}{\textit{Intensity robustness (100 = unchanged)}} \\
MRI & 87.9 & {\color{black!45}$+$1.7} & 89.6 & {\color{black!45}$-$0.4} & 89.2 \\
CT & 82.1 & {\color{black!45}$+$4.1} & 86.2 & {\color{black!45}$+$0.6} & 86.8 \\
PET & 87.1 & {\color{black!45}$+$3.0} & 90.1 & {\color{black!45}$+$0.2} & 90.3 \\
\addlinespace[2pt]
\multicolumn{6}{@{}l}{\textit{Cross-contrast correspondence, \%}} \\
BraTS-Africa, AUROC & 97.5 & {\color{black!45}$+$0.1} & 97.6 & {\color{black!45}$-$0.1} & 97.5 \\
BraTS-Africa, top-1 & 49.1 & {\color{black!45}$+$2.4} & 51.5 & {\color{black!45}$-$1.1} & 50.4 \\
ISLES'24, AUROC & 99.1 & {\color{black!45}$+$0.1} & 99.2 & {\color{black!45}$-$0.1} & 99.1 \\
ISLES'24, top-1 & 64.1 & {\color{black!45}$+$2.8} & 66.9 & {\color{black!45}$-$0.5} & 66.4 \\
\midrule
mean score over the 20 rows & 70.0 & {\color{black!45}$+$0.9} & 70.9 & {\color{black!45}$+$0.5} & 71.4 \\
mean rank (1 = best) & 2.65 & & 1.70 & & 1.65 \\
mean gap to the best stage & 1.72 & & 0.80 & & 0.33 \\
largest gap to the best stage & 4.7 & & 6.4 & & 2.5 \\
\bottomrule
\end{tabular}
\end{table}

The S2 checkpoint, after higher-resolution MRI, raises CT segmentation by 1.7 Dice points, label-free registration by
2.6 and CT token robustness by 4.1 points, and all three checkpoints match non-contrast CT to CT angiography at an
AUROC of about 99. S2 also opens
gaps: the linear probes and the k-NN vote fall below S1, and the two 30-case large-vessel occlusion (LVO) tasks lose
6.4 points.

The S3 checkpoint, after CT at the same resolution, closes the probe and k-NN gaps: the linear probes gain 2.4 and 3.6
points and end above S1, and the k-NN vote returns to the level of S1. It recovers 3.9 of the 6.4 LVO points and holds
S2's other gains within about a point, trailing S2 on MRI robustness and BraTS-Africa top-1.

The lead holds without the noisy LVO row: S3 still has the highest mean score (74.5, against 74.2 and 72.9) and the
smallest gaps (0.22 on average and 1.1 at most, against 0.51 and 3.6 for S2). Each stage is one pretraining run that
continues the previous one, so data and training length change together: the table traces the path of one model, and
the effects of resolution, of CT and of training length stay confounded in it.

\paragraph{Protocols.}
The three checkpoints are scored together on one fold, with one protocol for all of them, so the S3
column differs from the final model's cells in the result tables of \cref{sec:supp-cls,sec:supp-seg,sec:supp-robust}
(PET lesion F1: 49.9 here, 59.1 pooled over five folds in \cref{tab:pet}). Segmentation averages the MRI datasets of
\cref{tab:mri-seg} except BraTS-Africa and ISLES'22, which the ablation omits, and the five CT datasets of
\cref{tab:ct-seg} that it covers (AMOS22, TotalSegmentator, AISD, TriALS, ImageCAS). The
circle-of-Willis row, from non-contrast CT (NCCT), follows \cref{tab:isles}, and the LVO row is the Dice of the occlusion
region on its two 30-case tasks, on CT angiography (CTA) and on NCCT, where \cref{tab:isles} scores LVO detection by the
competition performance metric (CPM). Fluorodeoxyglucose (FDG) PET averages 3 head runs for S1 and 4 for S2 and S3.
Classification averages the three MRI targets and the two STOIC targets over 3 probe seeds. Registration (TotalSeg-MRI
to TriALS CT, with MIND-SSC) follows \cref{tab:supp-reg}, robustness \cref{tab:supp-robust} and correspondence
\cref{tab:supp-correspondence}. Every S3$-$S2 robustness and correspondence difference has a paired 95\% bootstrap
interval that excludes 0. The summary rows run over all 20 rows, and ties share a rank.

\FloatBarrier
\section{Ranking summary}
\label{sec:supp-rank}
\Maincref{fig:teaser} condenses the frozen comparison into one number per task group and encoder family. This section
defines that number (\cref{tab:supp-teaser}) and tests it against its two design choices and baseline exposure
(\cref{tab:supp-rank-robust}): \ours stays first or tied first in every group when datasets are balanced, ranks first
in most bootstrap draws, and leads by a modest effect size on CT segmentation and classification.

\paragraph{The plotted value.}
In every cell of a table that feeds a group, the $N$ encoders compared are ranked by the cell's metric, and ties share
the mean rank. The inverse normalized rank is $(N-r)/(N-1)$, which is 1 for the best encoder and 0 for the worst. A
group's value for an encoder is the mean over the cells of that table, so a dataset with several rows, targets or
readouts counts once per cell, and a family is drawn at its best member. The group tables rank seven encoders, with three
exceptions: PET ranks eight, since Curia-MAE, scored on PET only, enters the PET ranks without a column of
\cref{tab:supp-teaser}; CT classification ranks six, since VoxelFM is shown but not ranked (\cref{tab:ct-cls}); and
registration ranks the six rows of \cref{tab:supp-reg} with MIND-SSC or their own pipeline. The CT vessel group averages
circle-of-Willis Dice and LVO detection, both from NCCT.
Every table that feeds the figure lists the mean rank and the inverse normalized rank of each encoder, split into
unified models and specialists, and \cref{tab:supp-teaser} collects the plotted values: \ours is the best family in
every group, tied with anatomix on registration.

\begin{table}[!htbp]
\centering\footnotesize\setlength{\tabcolsep}{4pt}
\caption{\textbf{Values plotted in \maincref{fig:teaser}} (inverse normalized rank, 1 = best), each family at its best
member, named below its value. Table: source of the per-encoder values. Bold: best family. --: not scored.}
\label{tab:supp-teaser}
\begin{tabular}{@{}l l c c ccc@{}}
\toprule
 & & \multicolumn{2}{c}{Unified models} & \multicolumn{3}{c}{Specialists} \\
\cmidrule(lr){3-4}\cmidrule(l){5-7}
task group & table & \ours & nnFoundation & CT & MRI & registration \\
\midrule
MRI lesion and anatomy segmentation & \cref{tab:mri-seg} & \textbf{0.97} & \makecell{0.72\\[-2pt]{\tiny nnF-CNN}} & -- & \makecell{0.51\\[-2pt]{\tiny Triad Swin $\cdot$ SimMIM}} & -- \\
MRI pathology classification & \cref{tab:mri-cls} & \textbf{1.00} & \makecell{0.81\\[-2pt]{\tiny nnF-ViT}} & -- & \makecell{0.57\\[-2pt]{\tiny Triad CNN $\cdot$ MAE}} & -- \\
MRI$\leftrightarrow$CT registration & \cref{tab:supp-reg} & \textbf{0.90} & \makecell{0.60\\[-2pt]{\tiny nnF-ViT}} & -- & -- & \makecell{\textbf{0.90}\\[-2pt]{\tiny anatomix}} \\
CT lesion and anatomy segmentation & \cref{tab:ct-seg} & \textbf{0.90} & \makecell{0.29\\[-2pt]{\tiny nnF-ViT}} & \makecell{0.71\\[-2pt]{\tiny SPECTRE}} & -- & -- \\
CT pathology classification & \cref{tab:ct-cls} & \textbf{0.93} & \makecell{0.38\\[-2pt]{\tiny nnF-ViT}} & \makecell{0.81\\[-2pt]{\tiny SPECTRE}} & -- & -- \\
CT vessel segmentation (no contrast) & \cref{tab:isles} & \textbf{0.92} & \makecell{0.50\\[-2pt]{\tiny nnF-ViT}} & \makecell{0.83\\[-2pt]{\tiny SPECTRE}} & -- & -- \\
PET lesion segmentation (unseen) & \cref{tab:pet} & \textbf{1.00} & \makecell{0.57\\[-2pt]{\tiny nnF-CNN}} & \makecell{0.76\\[-2pt]{\tiny SPECTRE}} & \makecell{0.24\\[-2pt]{\tiny Triad CNN $\cdot$ MAE}} & -- \\
\bottomrule
\end{tabular}
\end{table}

\paragraph{Sensitivity.}
The figure averages over table cells, so datasets with many targets or readouts weigh more, and it uses point
estimates. \Cref{tab:supp-rank-robust} recomputes each group with one value per dataset (cells averaged within a
dataset first), resamples datasets with replacement to give an interval and the probability that \ours ranks first, removes the cells whose images our audit finds likely in a baseline's pretraining (SPECTRE and CT-Instruct on
AMOS22 and the MSD sets), and reports a margin that keeps effect size: \ours minus the best other encoder per cell, in units of
the across-encoder standard deviation of the cell, averaged within and then across datasets. The closest rival is the
encoder with the smallest balanced margin to \ours, given in parentheses. Family members compete individually in this table, while the figure
draws a family at its best member, which favors the baselines. Dataset balancing moves no group by more than 0.02, and
\ours ranks first in 75 to 100\% of the draws in every group with two or more datasets or tasks (75\% on the two-task
no-contrast vessel group), with or without the exposed cells. On registration it ties anatomix, and against \nnfvit it
loses no cell. The CT margins are small (0.04 and 0.11 standard deviations on segmentation and classification), so there
the lead is consistent but modest.

\begin{table}[!htbp]
\centering\footnotesize\setlength{\tabcolsep}{3pt}
\caption{\textbf{Sensitivity of the ranking summary} (\maincref{fig:teaser}). $P$(first): share of bootstrap draws in
which \ours is ahead of every other encoder on the datasets they share; no exposure: the same without the cells whose
images our audit finds likely in a baseline's pretraining. W/T/L: cells won, tied and lost against \nnfvit (ties:
\cref{sec:supp-impl}). --: a single dataset and task, no bootstrap.}
\label{tab:supp-rank-robust}
\begin{tabular}{@{}l c c c c l c c@{}}
\toprule
task group & \makecell{plotted\\[-2pt]{\tiny(\maincref{fig:teaser})}} & \makecell{dataset-\\[-2pt]balanced} & \makecell{$P$(first)} & \makecell{$P$(first)\\[-2pt]{\tiny no exposure}} & closest rival & \makecell{standardized\\[-2pt]margin} & \makecell{W/T/L\\[-2pt]{\tiny vs \nnfvit}} \\
\midrule
MRI lesion and anatomy seg. & 0.97 & 0.96\,{\tiny[0.91, 1.00]} & 0.99 & 0.99 & nnF-CNN (+0.26) & +0.73 & 4/6/0 \\
MRI pathology classification & 1.00 & 1.00\,{\tiny[1.00, 1.00]} & 1.00 & 1.00 & nnF-ViT (+0.19) & +0.61 & 9/0/0 \\
MRI$\leftrightarrow$CT registration & 0.90 & 0.90 & -- & -- & anatomix (+0.00) & $-$0.01 & 2/0/0 \\
CT lesion and anatomy seg. & 0.90 & 0.88\,{\tiny[0.79, 0.95]} & 0.84 & 0.94 & Merlin (+0.17) & +0.04 & 6/2/0 \\
CT pathology classification & 0.93 & 0.94\,{\tiny[0.93, 0.95]} & 1.00 & 1.00 & SPECTRE (+0.11) & +0.11 & 16/0/0 \\
CT vessel seg. (no contrast) & 0.92 & 0.92\,{\tiny[0.83, 1.00]} & 0.75 & 0.75 & SPECTRE (+0.08) & +0.34 & 1/1/0 \\
PET lesion seg. (unseen) & 1.00 & 1.00 & -- & -- & SPECTRE (+0.24) & +0.84 & 3/0/0 \\
\bottomrule
\end{tabular}
\end{table}

\FloatBarrier
\section{Classification}
\label{sec:supp-cls}
Classification tests whether the frozen tokens carry patient-level labels. This section gives the values behind \maincref{fig:r4} and the MRI cells of the k-NN panel of \maincref{fig:analysis}, whose STOIC cells are in \cref{sec:supp-analysis}: the MRI results (\cref{tab:mri-cls}), the CT results (\cref{tab:ct-cls,tab:supp-ct-probe-gain}) and the splits (\cref{tab:supp-cls-splits}).

On MRI, \ours is best in all nine columns, from training-free k-NN to the attention probe, with an average rank of 1.0 (\cref{tab:mri-cls}). Its linear probe is within 1.7 points of its attention probe on the two UCSF-PDGM targets, isocitrate dehydrogenase (IDH) status and WHO grade, so the mean token already holds the glioma labels.

\begin{table}[H]
\centering\small\setlength{\tabcolsep}{3.5pt}
\caption{\textbf{MRI classification} (AUROC \%, frozen encoders): UCSF-PDGM IDH status and WHO grade and LLD-MMRI liver lesion type, by training-free k-NN on the patch mean and two trained probes (\maincref{sec:protocol}). Bold: best per column.}
\label{tab:mri-cls}
\begin{tabular}{l ccc c ccc c ccc c cc}
\toprule
 & \multicolumn{3}{c}{k-NN (no training)} && \multicolumn{3}{c}{linear probe} && \multicolumn{3}{c}{multi-layer attention probe} && \multicolumn{2}{c}{over 9 cells} \\
\cmidrule{2-4}\cmidrule{6-8}\cmidrule{10-12}\cmidrule{14-15}
encoder & IDH & grade & LLD && IDH & grade & LLD && IDH & grade & LLD && \makecell{mean\\rank} & \makecell{inv.\ norm.\\rank} \\
\midrule
\multicolumn{15}{@{}l}{\textit{Unified models}} \\
\ours & \makecell{\textbf{86.1}\\[-2pt]{\phantom{\tiny[x, x]}}} & \makecell{\textbf{85.6}\\[-2pt]{\phantom{\tiny[x, x]}}} & \makecell{\textbf{79.7}\\[-2pt]{\phantom{\tiny[x, x]}}} && \makecell{\textbf{94.2}\\[-2pt]{\tiny[94.0, 94.3]}} & \makecell{\textbf{95.3}\\[-2pt]{\tiny[95.1, 95.5]}} & \makecell{\textbf{90.2}\\[-2pt]{\tiny[89.9, 90.5]}} && \makecell{\textbf{92.5}\\[-2pt]{\tiny[92.1, 92.9]}} & \makecell{\textbf{95.4}\\[-2pt]{\tiny[95.3, 95.5]}} & \makecell{\textbf{93.6}\\[-2pt]{\tiny[93.1, 94.0]}} && \textbf{1.0} & \textbf{1.00}  \\
nnF-ViT~\cite{nnfoundation} & \makecell{78.0\\[-2pt]{\phantom{\tiny[x, x]}}} & \makecell{74.7\\[-2pt]{\phantom{\tiny[x, x]}}} & \makecell{78.6\\[-2pt]{\phantom{\tiny[x, x]}}} && \makecell{79.4\\[-2pt]{\tiny[77.6, 80.9]}} & \makecell{88.3\\[-2pt]{\tiny[88.1, 88.4]}} & \makecell{86.5\\[-2pt]{\tiny[86.2, 86.9]}} && \makecell{82.2\\[-2pt]{\tiny[82.0, 82.3]}} & \makecell{89.1\\[-2pt]{\tiny[88.9, 89.3]}} & \makecell{87.6\\[-2pt]{\tiny[82.4, 91.3]}} && 2.1 & 0.81  \\
nnF-CNN~\cite{nnfoundation} & \makecell{58.2\\[-2pt]{\phantom{\tiny[x, x]}}} & \makecell{66.7\\[-2pt]{\phantom{\tiny[x, x]}}} & \makecell{66.4\\[-2pt]{\phantom{\tiny[x, x]}}} && \makecell{57.4\\[-2pt]{\tiny[53.8, 62.9]}} & \makecell{59.7\\[-2pt]{\tiny[57.5, 66.3]}} & \makecell{61.5\\[-2pt]{\tiny[55.3, 62.4]}} && \makecell{73.6\\[-2pt]{\tiny[70.5, 74.9]}} & \makecell{68.7\\[-2pt]{\tiny[67.4, 71.2]}} & \makecell{60.7\\[-2pt]{\tiny[59.3, 63.5]}} && 6.0 & 0.17  \\
\addlinespace[3pt]
\multicolumn{15}{@{}l}{\textit{Specialists}} \\
Triad CNN $\cdot$ MAE~\cite{triad} & \makecell{71.3\\[-2pt]{\phantom{\tiny[x, x]}}} & \makecell{61.2\\[-2pt]{\phantom{\tiny[x, x]}}} & \makecell{77.0\\[-2pt]{\phantom{\tiny[x, x]}}} && \makecell{73.4\\[-2pt]{\tiny[69.4, 79.6]}} & \makecell{62.2\\[-2pt]{\tiny[62.1, 62.5]}} & \makecell{85.1\\[-2pt]{\tiny[84.7, 85.8]}} && \makecell{81.8\\[-2pt]{\tiny[78.7, 85.2]}} & \makecell{71.2\\[-2pt]{\tiny[64.6, 76.8]}} & \makecell{88.2\\[-2pt]{\tiny[87.1, 88.8]}} && 3.6 & 0.57  \\
Triad CNN $\cdot$ SimMIM~\cite{triad} & \makecell{64.3\\[-2pt]{\phantom{\tiny[x, x]}}} & \makecell{57.2\\[-2pt]{\phantom{\tiny[x, x]}}} & \makecell{66.1\\[-2pt]{\phantom{\tiny[x, x]}}} && \makecell{68.9\\[-2pt]{\tiny[62.5, 69.7]}} & \makecell{62.4\\[-2pt]{\tiny[60.3, 64.9]}} & \makecell{69.7\\[-2pt]{\tiny[66.9, 77.5]}} && \makecell{66.3\\[-2pt]{\tiny[62.8, 70.5]}} & \makecell{58.9\\[-2pt]{\tiny[55.4, 64.7]}} & \makecell{70.4\\[-2pt]{\tiny[66.6, 72.9]}} && 6.0 & 0.17  \\
Triad Swin $\cdot$ MAE~\cite{triad} & \makecell{62.8\\[-2pt]{\phantom{\tiny[x, x]}}} & \makecell{64.4\\[-2pt]{\phantom{\tiny[x, x]}}} & \makecell{69.2\\[-2pt]{\phantom{\tiny[x, x]}}} && \makecell{65.7\\[-2pt]{\tiny[64.8, 69.3]}} & \makecell{64.9\\[-2pt]{\tiny[61.5, 67.3]}} & \makecell{79.4\\[-2pt]{\tiny[77.1, 70.9]}} && \makecell{67.0\\[-2pt]{\tiny[65.4, 68.2]}} & \makecell{56.6\\[-2pt]{\tiny[54.9, 65.9]}} & \makecell{77.6\\[-2pt]{\tiny[74.9, 79.7]}} && 5.4 & 0.26  \\
Triad Swin $\cdot$ SimMIM~\cite{triad} & \makecell{69.4\\[-2pt]{\phantom{\tiny[x, x]}}} & \makecell{67.6\\[-2pt]{\phantom{\tiny[x, x]}}} & \makecell{73.4\\[-2pt]{\phantom{\tiny[x, x]}}} && \makecell{72.0\\[-2pt]{\tiny[70.9, 73.8]}} & \makecell{69.1\\[-2pt]{\tiny[68.4, 72.6]}} & \makecell{83.2\\[-2pt]{\tiny[82.3, 83.6]}} && \makecell{76.2\\[-2pt]{\tiny[74.8, 76.9]}} & \makecell{66.6\\[-2pt]{\tiny[63.6, 68.8]}} & \makecell{86.7\\[-2pt]{\tiny[85.1, 86.9]}} && 3.9 & 0.52  \\
\bottomrule
\end{tabular}
\end{table}

On CT, \ours has the best average rank with either probe among the six ranked encoders: 1.3 with the linear probe, 1.4 with the attention probe and 1.3 over all 16 cells. VoxelFM, whose RSNA 2023 images are in its pretraining, is shown but not ranked. With the attention probe, \ours leads on extravasation, liver and spleen and ties VoxelFM on COVID-19.

\begin{table}[H]
\centering\footnotesize\setlength{\tabcolsep}{4pt}
\caption{\textbf{CT classification} (AUROC \%, frozen encoders) on STOIC and RSNA 2023 abdominal trauma, by two trained
probes (\maincref{sec:protocol}). Bold: best per column. $^\dagger$: images (not labels) of the dataset are in the encoder's
self-supervised pretraining. VoxelFM saw RSNA 2023 images, so it is shown but not ranked (N/A): the ranks run over the
other six encoders, here and in \maincref{fig:teaser}.}
\label{tab:ct-cls}
\begin{tabular}{@{}l cc cccccc@{}}
\multicolumn{9}{@{}l}{(a) Linear probe} \\
\toprule
 & \multicolumn{2}{c}{STOIC} & \multicolumn{6}{c}{RSNA 2023} \\
\cmidrule(lr){2-3}\cmidrule(lr){4-9}
encoder & COVID & severe & any & bowel & extrav. & kidney & liver & spleen \\
\midrule
\multicolumn{9}{@{}l}{\textit{Unified models}} \\
\ours & \makecell{82.8\\[-2pt]{\tiny[81.9, 83.8]}} & \makecell{78.1\\[-2pt]{\tiny[76.6, 79.2]}} & \makecell{77.3\\[-2pt]{\tiny[75.6, 78.8]}} & \makecell{\textbf{82.3}\\[-2pt]{\tiny[79.9, 83.5]}} & \makecell{74.5\\[-2pt]{\tiny[72.9, 76.4]}} & \makecell{\textbf{79.9}\\[-2pt]{\tiny[77.1, 81.6]}} & \makecell{78.6\\[-2pt]{\tiny[77.5, 81.0]}} & \makecell{\textbf{79.1}\\[-2pt]{\tiny[77.1, 80.2]}}  \\
nnF-ViT~\cite{nnfoundation} & \makecell{76.4\\[-2pt]{\tiny[73.6, 79.4]}} & \makecell{72.3\\[-2pt]{\tiny[69.8, 75.0]}} & \makecell{70.4\\[-2pt]{\tiny[69.2, 72.7]}} & \makecell{61.0\\[-2pt]{\tiny[58.3, 63.2]}} & \makecell{63.6\\[-2pt]{\tiny[62.0, 64.3]}} & \makecell{76.2\\[-2pt]{\tiny[74.7, 78.2]}} & \makecell{73.3\\[-2pt]{\tiny[70.5, 76.2]}} & \makecell{69.5\\[-2pt]{\tiny[67.9, 72.1]}}  \\
nnF-CNN~\cite{nnfoundation} & \makecell{57.1\\[-2pt]{\tiny[56.0, 59.6]}} & \makecell{60.7\\[-2pt]{\tiny[58.9, 61.1]}} & \makecell{57.7\\[-2pt]{\tiny[55.4, 59.1]}} & \makecell{47.8\\[-2pt]{\tiny[45.3, 49.9]}} & \makecell{52.3\\[-2pt]{\tiny[51.9, 54.0]}} & \makecell{62.6\\[-2pt]{\tiny[59.9, 63.6]}} & \makecell{58.5\\[-2pt]{\tiny[57.3, 61.2]}} & \makecell{53.0\\[-2pt]{\tiny[52.1, 54.9]}}  \\
\addlinespace[3pt]
\multicolumn{9}{@{}l}{\textit{Specialists}} \\
Merlin~\cite{merlin} & \makecell{78.9\\[-2pt]{\tiny[77.9, 81.8]}} & \makecell{73.1\\[-2pt]{\tiny[70.6, 74.7]}} & \makecell{76.3\\[-2pt]{\tiny[75.7, 77.5]}} & \makecell{72.0\\[-2pt]{\tiny[70.3, 74.8]}} & \makecell{69.1\\[-2pt]{\tiny[68.4, 70.9]}} & \makecell{\textbf{79.9}\\[-2pt]{\tiny[77.7, 81.7]}} & \makecell{76.3\\[-2pt]{\tiny[73.8, 78.1]}} & \makecell{77.6\\[-2pt]{\tiny[74.7, 79.6]}}  \\
CT-Instruct~\cite{ctinstruct} & \makecell{66.1\\[-2pt]{\tiny[64.2, 67.7]}} & \makecell{67.5\\[-2pt]{\tiny[65.5, 68.9]}} & \makecell{59.0\\[-2pt]{\tiny[57.1, 60.2]}} & \makecell{56.5\\[-2pt]{\tiny[55.6, 57.4]}} & \makecell{56.7\\[-2pt]{\tiny[54.7, 58.8]}} & \makecell{61.2\\[-2pt]{\tiny[59.6, 61.8]}} & \makecell{61.1\\[-2pt]{\tiny[58.7, 63.8]}} & \makecell{61.5\\[-2pt]{\tiny[58.7, 64.1]}}  \\
SPECTRE~\cite{spectre} & \makecell{81.0\\[-2pt]{\tiny[80.0, 82.0]}} & \makecell{78.8\\[-2pt]{\tiny[77.9, 79.7]}} & \makecell{75.1\\[-2pt]{\tiny[72.9, 76.2]}} & \makecell{70.2\\[-2pt]{\tiny[67.7, 72.8]}} & \makecell{\textbf{74.8}\\[-2pt]{\tiny[72.1, 76.7]}} & \makecell{76.6\\[-2pt]{\tiny[73.7, 78.5]}} & \makecell{78.5\\[-2pt]{\tiny[76.9, 80.6]}} & \makecell{77.5\\[-2pt]{\tiny[74.5, 80.3]}}  \\
VoxelFM~\cite{voxelfm} & \makecell{\textbf{83.9}\\[-2pt]{\tiny[83.1, 84.6]}} & \makecell{\textbf{78.9}\\[-2pt]{\tiny[77.9, 79.9]}} & \makecell{\textbf{77.5}$^\dagger$\\[-2pt]{\tiny[75.5, 80.1]}} & \makecell{65.8$^\dagger$\\[-2pt]{\tiny[64.8, 68.1]}} & \makecell{71.4$^\dagger$\\[-2pt]{\tiny[70.7, 72.4]}} & \makecell{79.4$^\dagger$\\[-2pt]{\tiny[76.9, 80.7]}} & \makecell{\textbf{78.9}$^\dagger$\\[-2pt]{\tiny[76.4, 79.6]}} & \makecell{78.1$^\dagger$\\[-2pt]{\tiny[77.3, 80.3]}}  \\
\bottomrule
\end{tabular}
\par\vspace{1em}
\begin{tabular}{@{}l cc cccccc c cc@{}}
\multicolumn{12}{@{}l}{(b) Multi-layer attention probe} \\
\toprule
 & \multicolumn{2}{c}{STOIC} & \multicolumn{6}{c}{RSNA 2023} && \multicolumn{2}{c}{over all 16 cells} \\
\cmidrule(lr){2-3}\cmidrule(lr){4-9}\cmidrule(l){11-12}
encoder & COVID & severe & any & bowel & extrav. & kidney & liver & spleen && \makecell{mean\\rank} & \makecell{inv.\ norm.\\rank} \\
\midrule
\multicolumn{12}{@{}l}{\textit{Unified models}} \\
\ours & \makecell{\textbf{85.1}\\[-2pt]{\tiny[84.1, 86.2]}} & \makecell{80.6\\[-2pt]{\tiny[79.7, 81.5]}} & \makecell{80.8\\[-2pt]{\tiny[78.6, 81.3]}} & \makecell{79.6\\[-2pt]{\tiny[77.5, 81.6]}} & \makecell{\textbf{78.3}\\[-2pt]{\tiny[76.4, 79.7]}} & \makecell{86.3\\[-2pt]{\tiny[84.9, 89.2]}} & \makecell{\textbf{83.9}\\[-2pt]{\tiny[83.4, 84.4]}} & \makecell{\textbf{84.1}\\[-2pt]{\tiny[81.2, 85.0]}} &  & \textbf{1.3} & \textbf{0.93}  \\
nnF-ViT~\cite{nnfoundation} & \makecell{69.1\\[-2pt]{\tiny[68.4, 70.0]}} & \makecell{67.9\\[-2pt]{\tiny[65.4, 70.7]}} & \makecell{61.9\\[-2pt]{\tiny[61.4, 63.4]}} & \makecell{63.6\\[-2pt]{\tiny[62.9, 64.7]}} & \makecell{55.4\\[-2pt]{\tiny[54.4, 57.5]}} & \makecell{64.0\\[-2pt]{\tiny[62.7, 64.9]}} & \makecell{65.2\\[-2pt]{\tiny[63.5, 65.7]}} & \makecell{59.5\\[-2pt]{\tiny[58.8, 62.5]}} &  & 4.1 & 0.38  \\
nnF-CNN~\cite{nnfoundation} & \makecell{60.2\\[-2pt]{\tiny[59.3, 61.5]}} & \makecell{59.4\\[-2pt]{\tiny[57.1, 62.0]}} & \makecell{56.0\\[-2pt]{\tiny[53.2, 56.8]}} & \makecell{49.9\\[-2pt]{\tiny[47.8, 52.8]}} & \makecell{56.2\\[-2pt]{\tiny[55.6, 58.4]}} & \makecell{57.2\\[-2pt]{\tiny[54.6, 58.5]}} & \makecell{51.8\\[-2pt]{\tiny[50.7, 53.8]}} & \makecell{53.7\\[-2pt]{\tiny[52.1, 54.6]}} &  & 5.8 & 0.05  \\
\addlinespace[3pt]
\multicolumn{12}{@{}l}{\textit{Specialists}} \\
Merlin~\cite{merlin} & \makecell{75.5\\[-2pt]{\tiny[73.9, 77.0]}} & \makecell{72.9\\[-2pt]{\tiny[71.0, 74.6]}} & \makecell{73.2\\[-2pt]{\tiny[72.0, 74.5]}} & \makecell{75.3\\[-2pt]{\tiny[72.7, 76.4]}} & \makecell{62.9\\[-2pt]{\tiny[61.0, 63.3]}} & \makecell{74.5\\[-2pt]{\tiny[72.2, 75.8]}} & \makecell{75.4\\[-2pt]{\tiny[74.9, 76.5]}} & \makecell{75.1\\[-2pt]{\tiny[74.1, 78.0]}} &  & 2.7 & 0.66  \\
CT-Instruct~\cite{ctinstruct} & \makecell{66.5\\[-2pt]{\tiny[65.2, 67.6]}} & \makecell{67.2\\[-2pt]{\tiny[65.9, 70.1]}} & \makecell{59.0\\[-2pt]{\tiny[57.0, 61.0]}} & \makecell{53.8\\[-2pt]{\tiny[51.5, 55.2]}} & \makecell{53.2\\[-2pt]{\tiny[51.7, 55.3]}} & \makecell{55.9\\[-2pt]{\tiny[55.5, 56.8]}} & \makecell{62.8\\[-2pt]{\tiny[61.5, 63.8]}} & \makecell{60.9\\[-2pt]{\tiny[58.8, 62.3]}} &  & 5.1 & 0.17  \\
SPECTRE~\cite{spectre} & \makecell{83.2\\[-2pt]{\tiny[82.2, 84.1]}} & \makecell{78.5\\[-2pt]{\tiny[77.7, 79.7]}} & \makecell{\textbf{81.2}\\[-2pt]{\tiny[79.0, 82.7]}} & \makecell{\textbf{80.0}\\[-2pt]{\tiny[77.9, 80.5]}} & \makecell{77.8\\[-2pt]{\tiny[76.2, 78.9]}} & \makecell{\textbf{87.4}\\[-2pt]{\tiny[86.6, 89.2]}} & \makecell{83.1\\[-2pt]{\tiny[81.4, 85.0]}} & \makecell{82.7\\[-2pt]{\tiny[80.3, 85.4]}} &  & 1.9 & 0.81  \\
VoxelFM~\cite{voxelfm} & \makecell{\textbf{85.1}\\[-2pt]{\tiny[84.1, 86.0]}} & \makecell{\textbf{80.7}\\[-2pt]{\tiny[79.8, 81.8]}} & \makecell{79.1$^\dagger$\\[-2pt]{\tiny[76.4, 81.6]}} & \makecell{73.2$^\dagger$\\[-2pt]{\tiny[72.3, 76.2]}} & \makecell{72.6$^\dagger$\\[-2pt]{\tiny[70.6, 73.2]}} & \makecell{78.4$^\dagger$\\[-2pt]{\tiny[76.1, 81.4]}} & \makecell{80.8$^\dagger$\\[-2pt]{\tiny[79.4, 83.0]}} & \makecell{80.1$^\dagger$\\[-2pt]{\tiny[78.9, 81.1]}} &  & N/A & N/A  \\
\bottomrule
\end{tabular}
\end{table}

Averaged over the eight CT targets, the attention probe adds 3.3 AUROC points to \ours and 5.2 to SPECTRE, and it costs every encoder that is not a self-distilled ViT.

\begin{table}[H]
\centering\small\setlength{\tabcolsep}{5pt}
\caption{\textbf{The attention probe helps the self-distilled ViTs on CT} (AUROC \%, mean over the 8 targets of
\cref{tab:ct-cls}, frozen encoders). $\Delta$: attention probe minus linear probe, from unrounded means.}
\label{tab:supp-ct-probe-gain}
\begin{tabular}{l ccc}
\toprule
encoder & linear & attention & $\Delta$ \\
\midrule
Merlin~\cite{merlin} & 75.4 & 73.1 & $-2.3$ \\
CT-Instruct~\cite{ctinstruct} & 61.2 & 59.9 & $-1.3$ \\
nnF-CNN~\cite{nnfoundation} & 56.2 & 55.6 & $-0.7$ \\
nnF-ViT~\cite{nnfoundation} & 70.3 & 63.3 & $-7.0$ \\
SPECTRE~\cite{spectre} & 76.6 & 81.7 & $+5.2$ \\
VoxelFM~\cite{voxelfm} & 76.7 & 78.8 & $+2.0$ \\
\ours & \textbf{79.1} & \textbf{82.3} & $+3.3$ \\
\bottomrule
\end{tabular}
\end{table}

\paragraph{Protocols.} Each dataset has one fixed patient-level train/test split (\cref{tab:supp-cls-splits}). On UCSF-PDGM and LLD-MMRI it is a seeded 80/20 split of all patients, stratified on IDH status and grade for UCSF-PDGM. The grade target is WHO grade 4 against grades 2--3. STOIC severe cases are a subset of the COVID-19 cases, and severe is scored over all test patients. On RSNA 2023, kidney, liver and spleen are graded healthy, low or high, and are scored as injured (low or high) against healthy. UCSF-PDGM is read from T1c and FLAIR, LLD-MMRI is scored by macro one-vs-rest AUROC, and CT classification runs at 2\,mm, with a 384\,mm field of view on RSNA 2023. The k-NN vote and the two probes follow \maincref{sec:protocol} and \cref{sec:supp-impl}.

\begin{table}[H]
\centering\footnotesize\setlength{\tabcolsep}{4pt}
\caption{\textbf{Classification splits.} Patients (RSNA 2023: cases) per split, and class counts as train / test; the RSNA 2023 targets other than any injury give train counts only. ICC: intrahepatic cholangiocarcinoma. FNH: focal nodular hyperplasia. HCC: hepatocellular carcinoma.}
\label{tab:supp-cls-splits}
\begin{tabular}{@{}l r r l >{\raggedright\arraybackslash}p{9.6cm}@{}}
\toprule
dataset & train & test & target & classes, train / test \\
\midrule
UCSF-PDGM & 395 & 100 & IDH & wildtype 313 / 79, mutant 82 / 21 \\
 & & & grade & grades 2--3: 79 / 20, grade 4: 316 / 80 \\
\addlinespace[2pt]
LLD-MMRI & 398 & 100 & lesion type & hemangioma 61 / 18, ICC 48 / 10, abscess 47 / 7, metastasis 41 / 10, cyst 44 / 9,
FNH 34 / 12, HCC 123 / 34 \\
\addlinespace[2pt]
STOIC & 1,600 & 400 & COVID-19 & positive 961 / 244 \\
 & & & severe & severe 243 / 58 \\
\addlinespace[2pt]
RSNA 2023 & 2,512 & 628 & any injury & injured 731 / 174 \\
 & & & bowel & injured 52 \\
 & & & extravasation & injured 174 \\
 & & & kidney & low 113, high 65 \\
 & & & liver & low 218, high 53 \\
 & & & spleen & low 168, high 133 \\
\bottomrule
\end{tabular}
\end{table}

\FloatBarrier
\section{Lesion and anatomy segmentation}
\label{sec:supp-seg}\label{sec:supp-mriseg}\label{sec:supp-ctseg}
Segmentation tests whether the frozen tokens resolve lesions and anatomy under a trained decoder. This section gives every segmentation cell behind \maincref{fig:r1,fig:r2}: MRI and CT with bootstrap intervals (\cref{tab:mri-seg,tab:ct-seg}), FDG PET (\cref{sec:supp-pet}) and the label-efficiency sweeps (\cref{sec:supp-le}).

On MRI, \ours has the best Dice on 8 of the 10 rows (ATLAS gives a liver and a tumor row) and an average rank of 1.2 (\cref{tab:mri-seg}). On the other two, MSD Hippocampus and ISLES'22, \nnfcnn is ahead by 0.2 and 1.1 points and level with it under the paired test.

\begin{table}[!htbp]
\centering\footnotesize\setlength{\tabcolsep}{3pt}
\caption{\textbf{MRI segmentation} (Dice \%, frozen encoders, mean over 3 folds, 95\% bootstrap CI beside each value). Bold: best per row. Hippocampus: MSD Hippocampus.}
\label{tab:mri-seg}
\begin{tabular}{l ccc @{\hspace{14pt}} cccc}
\toprule
& \multicolumn{3}{c}{Unified models} & \multicolumn{4}{c}{Specialists} \\
\cmidrule(lr){2-4}\cmidrule(lr){5-8}
dataset & \ours & \makecell{nnF-ViT\\\cite{nnfoundation}} & \makecell{nnF-CNN\\\cite{nnfoundation}} & \makecell{Triad\\CNN $\cdot$ MAE\\\cite{triad}} & \makecell{Triad\\CNN $\cdot$ SimMIM\\\cite{triad}} & \makecell{Triad\\Swin $\cdot$ MAE\\\cite{triad}} & \makecell{Triad\\Swin $\cdot$ SimMIM\\\cite{triad}} \\
\midrule
Prostate158 & \textbf{81.2}\,{\tiny[80.1, 82.2]} & 76.9\,{\tiny[75.6, 78.1]} & 78.2\,{\tiny[76.7, 79.5]} & 76.1\,{\tiny[74.6, 77.5]} & 76.4\,{\tiny[75.1, 77.7]} & 76.2\,{\tiny[74.7, 77.4]} & 77.2\,{\tiny[75.9, 78.3]} \\
ACDC & \textbf{91.0}\,{\tiny[90.5, 91.6]} & 88.7\,{\tiny[87.8, 89.5]} & 89.1\,{\tiny[88.2, 90.0]} & 88.2\,{\tiny[87.3, 89.1]} & 88.1\,{\tiny[87.1, 89.1]} & 86.1\,{\tiny[84.8, 87.2]} & 86.8\,{\tiny[85.7, 87.8]} \\
ATLAS liver & \textbf{91.4}\,{\tiny[89.7, 92.9]} & 89.5\,{\tiny[87.3, 91.3]} & 90.5\,{\tiny[88.0, 92.5]} & 89.5\,{\tiny[87.5, 91.3]} & 89.6\,{\tiny[87.4, 91.3]} & 88.6\,{\tiny[86.6, 90.4]} & 90.1\,{\tiny[88.2, 91.7]} \\
ATLAS tumor & \textbf{67.1}\,{\tiny[61.1, 72.1]} & 56.3\,{\tiny[49.3, 62.8]} & 57.6\,{\tiny[49.7, 64.8]} & 47.6\,{\tiny[40.5, 54.4]} & 50.7\,{\tiny[43.5, 58.1]} & 45.3\,{\tiny[38.7, 52.0]} & 52.2\,{\tiny[44.7, 59.1]} \\
AMOS-MRI & \textbf{85.7}\,{\tiny[84.5, 86.6]} & 75.8\,{\tiny[74.0, 77.6]} & 79.1\,{\tiny[77.3, 80.9]} & 81.3\,{\tiny[79.7, 83.0]} & 81.4\,{\tiny[79.8, 83.0]} & 78.6\,{\tiny[76.8, 80.3]} & 81.5\,{\tiny[80.0, 82.9]} \\
MM-WHS & \textbf{86.9}\,{\tiny[85.7, 88.5]} & 84.0\,{\tiny[81.8, 86.2]} & {86.1}\,{\tiny[84.0, 88.3]} & {86.3}\,{\tiny[84.6, 88.0]} & {86.3}\,{\tiny[84.6, 88.1]} & 84.6\,{\tiny[82.3, 86.8]} & {86.6}\,{\tiny[84.9, 88.4]} \\
Hippocampus & 89.3\,{\tiny[88.9, 89.7]} & 89.0\,{\tiny[88.6, 89.3]} & \textbf{89.5}\,{\tiny[89.1, 89.9]} & 89.1\,{\tiny[88.8, 89.5]} & 89.1\,{\tiny[88.7, 89.5]} & 88.9\,{\tiny[88.5, 89.2]} & 88.9\,{\tiny[88.5, 89.2]} \\
TotalSeg-MRI & \textbf{71.0}\,{\tiny[67.2, 74.7]} & 47.1\,{\tiny[42.4, 52.0]} & 47.6\,{\tiny[42.5, 53.0]} & 59.4\,{\tiny[55.3, 63.5]} & 52.7\,{\tiny[49.2, 56.2]} & 46.8\,{\tiny[43.1, 50.4]} & 57.8\,{\tiny[53.9, 61.3]} \\
BraTS-Africa & \textbf{79.9}\,{\tiny[76.4, 82.9]} & 76.3\,{\tiny[72.1, 80.1]} & 78.4\,{\tiny[74.4, 81.8]} & 77.0\,{\tiny[72.9, 80.6]} & 76.0\,{\tiny[72.0, 79.6]} & 75.0\,{\tiny[71.1, 78.8]} & 76.9\,{\tiny[72.9, 80.5]} \\
ISLES'22 & 77.1\,{\tiny[73.7, 80.2]} & 76.8\,{\tiny[73.7, 79.7]} & \textbf{78.2}\,{\tiny[74.9, 81.3]} & 75.0\,{\tiny[71.4, 78.1]} & 75.0\,{\tiny[71.4, 78.5]} & 71.3\,{\tiny[67.4, 75.0]} & 71.9\,{\tiny[68.0, 75.6]} \\
\midrule
mean rank & \textbf{1.2} & 4.8 & 2.7 & 4.3 & 4.3 & 6.7 & 4.0 \\
inv.\ norm.\ rank & \textbf{0.97} & 0.36 & 0.72 & 0.45 & 0.44 & 0.06 & 0.51 \\
\bottomrule
\end{tabular}
\end{table}

On CT, \ours has the best Dice on 4 of 8 rows: TotalSegmentator and TriALS (visible), where its lead is significant, TriALS (all) and ImageCAS (\cref{tab:ct-seg}). Merlin leads on MSD Pancreas and AISD, and SPECTRE on AMOS22 and MSD Colon, both in its pretraining. \ours has the best average rank, 1.6. \Cref{sec:supp-le} adds TotalSegmentator at 1\% and 10\% of the labels.

\begin{table}[!htbp]
\centering\footnotesize\setlength{\tabcolsep}{3pt}
\caption{\textbf{CT segmentation} (Dice \%, frozen encoders, mean over 3 folds, TotalSegmentator on its official test split; 95\% bootstrap CI beside each value). Bold: best per row. TriALS (visible): the 37 cases whose lesions are predominantly visible on non-contrast CT, selected from the ground-truth volumes alone, so the same cases for every encoder. TriALS (all): every case. $^\dagger$: images (not labels) of the test set are in the encoder's pretraining; CT-Instruct also trained on its segmentation labels (\cref{sec:supp-scope}). TriALS (visible) is the TriALS row of \maincref{fig:r1}.}
\label{tab:ct-seg}
\begin{tabular}{l ccc @{\hspace{14pt}} cccc}
\toprule
& \multicolumn{3}{c}{Unified models} & \multicolumn{4}{c}{Specialists} \\
\cmidrule(lr){2-4}\cmidrule(lr){5-8}
dataset & \ours & \makecell{nnF-ViT\\\cite{nnfoundation}} & \makecell{nnF-CNN\\\cite{nnfoundation}} & \makecell{Merlin\\\cite{merlin}} & \makecell{CT-\\Instruct\\\cite{ctinstruct}} & \makecell{SPECTRE\\\cite{spectre}} & \makecell{VoxelFM\\\cite{voxelfm}} \\
\midrule
AMOS22 & 86.9\,{\tiny[85.6, 88.2]} & 76.7\,{\tiny[74.7, 78.6]} & 75.5\,{\tiny[73.3, 77.4]} & 85.1\,{\tiny[83.6, 86.5]} & 71.4$^\dagger$\,{\tiny[69.1, 73.5]} & \textbf{87.9}$^\dagger$\,{\tiny[86.6, 89.0]} & 85.4\,{\tiny[83.9, 86.7]} \\
MSD Pancreas & 59.9\,{\tiny[58.4, 62.1]} & 52.9\,{\tiny[48.4, 57.8]} & 52.4\,{\tiny[48.5, 56.7]} & \textbf{61.2}\,{\tiny[57.2, 65.6]} & 51.8$^\dagger$\,{\tiny[47.6, 56.3]} & 60.6$^\dagger$\,{\tiny[56.7, 64.8]} & 57.8$^\dagger$\,{\tiny[53.7, 62.2]} \\
MSD Colon & {45.8}\,{\tiny[35.1, 58.6]} & {29.3}\,{\tiny[20.4, 38.6]} & 19.5\,{\tiny[12.8, 26.6]} & 40.6\,{\tiny[28.9, 53.2]} & 22.9$^\dagger$\,{\tiny[14.3, 32.0]} & \textbf{46.7}$^\dagger$\,{\tiny[35.5, 58.8]} & 34.1\,{\tiny[22.5, 46.5]} \\
TotalSegmentator & \textbf{91.5}\,{\tiny[88.0, 94.2]} & 89.5\,{\tiny[85.7, 92.6]} & 89.4\,{\tiny[85.6, 92.6]} & 88.8\,{\tiny[85.2, 91.9]} & 82.5$^\dagger$\,{\tiny[77.6, 86.8]} & 87.6$^\dagger$\,{\tiny[83.8, 92.3]} & 86.9$^\dagger$\,{\tiny[83.1, 91.9]} \\
AISD & 43.2\,{\tiny[40.1, 48.8]} & 16.9\,{\tiny[14.9, 18.9]} & 12.0\,{\tiny[10.5, 13.6]} & \textbf{45.1}\,{\tiny[41.3, 49.2]} & 34.3\,{\tiny[31.0, 37.9]} & 40.5\,{\tiny[36.8, 44.6]} & 40.8\,{\tiny[37.2, 44.9]} \\
TriALS (visible) & \textbf{47.2}\,{\tiny[36.5, 57.5]} & 13.1\,{\tiny[4.5, 23.4]} & 12.0\,{\tiny[3.5, 22.0]} & 34.2\,{\tiny[21.5, 46.9]} & 14.9\,{\tiny[7.1, 23.7]} & {45.1}\,{\tiny[30.9, 58.9]} & 39.0\,{\tiny[25.9, 52.0]} \\
TriALS (all) & \textbf{32.6}\,{\tiny[24.8, 41.2]} & 8.0\,{\tiny[3.2, 13.3]} & 6.9\,{\tiny[2.2, 12.1]} & {24.0}\,{\tiny[15.9, 33.0]} & {9.1}\,{\tiny[4.6, 14.2]} & 31.5\,{\tiny[22.0, 41.7]} & {26.6}\,{\tiny[18.2, 35.9]} \\
ImageCAS & \textbf{82.2}\,{\tiny[81.6, 82.7]} & 80.1\,{\tiny[79.4, 80.8]} & 81.3\,{\tiny[80.7, 82.0]} & 82.1\,{\tiny[81.5, 82.6]} & 80.8\,{\tiny[80.2, 81.4]} & 81.2\,{\tiny[80.6, 81.7]} & 81.5\,{\tiny[80.9, 82.0]} \\
\midrule
mean rank & \textbf{1.6} & 5.2 & 5.9 & 2.9 & 6.0 & 2.8 & 3.6 \\
inv.\ norm.\ rank & \textbf{0.90} & 0.29 & 0.19 & 0.69 & 0.17 & 0.71 & 0.56 \\
\bottomrule
\end{tabular}
\end{table}

\subsection{FDG PET}
\label{sec:supp-pet}
This subsection gives the fluorodeoxyglucose (FDG) PET values behind \maincref{fig:r1}, on a modality \ours never sees in pretraining. PET is scored for eight encoders: \ours, the two nnFoundation models, SPECTRE, VoxelFM, Merlin, Triad CNN $\cdot$ MAE and Curia-MAE~\cite{curiamae}, a residual-encoder CNN (ResEnc-L~\cite{nnunetrevisited}) pretrained on CT and MRI that we add for this task. A floor, \ours with random weights (its own initialization, seed 0), is read through the same teacher path, frozen, with the same UNETR head and protocol, and measures what pretraining adds. \ours has the best lesion Dice and F1 pooled over the five folds of autoPET~\cite{autopet} and in each fold, with the fewest false positives.

Pooled over the five folds, \ours has the highest lesion Dice, over the lesion-positive studies and over all studies, and the highest lesion F1 (\cref{tab:pet}). Each of its pooled margins over another encoder in lesion-positive Dice and in lesion F1 is significant on the paired intervals of \cref{tab:supp-pet-froc}. Of the two encoders pretrained on PET ($\checkmark$), \nnfcnn ranks second on lesion Dice and last on lesion F1, and \nnfvit last on lesion Dice. The random-weight floor trails \ours by 14.5 lesion Dice and 16.8 lesion F1 points pooled, with paired intervals that exclude zero in every fold, and on lesion F1 the point estimates of both nnFoundation encoders fall below it.

\begin{table}[!htbp]
  \centering
  \caption{\textbf{FDG PET lesion segmentation, a modality no stage of \ours trains on} (autoPET, five folds pooled, 1,014 studies; 95\% bootstrap CI beside each value). Dice lesion$+$: over the lesion-positive studies. All: over all studies (nnU-Net rule). $\checkmark$: pretrained on PET. $^\S$: UNet3D head matched to UNETR. Mean rank and inverse normalized rank: over the three cells, among the eight encoders; the random-weight floor is not ranked. Bold: best per column.}
  \label{tab:pet}
  \setlength{\tabcolsep}{5pt}\renewcommand{\arraystretch}{1.15}
  \footnotesize
  \begin{tabular}{@{}lcccccc@{}}
    \toprule
    & PET in & \multicolumn{2}{c}{Dice} & lesion & \multicolumn{2}{c}{over 3 cells} \\
    \cmidrule(lr){3-4}\cmidrule(l){6-7}
    encoder & pretraining & lesion$+$ & all & F1 & \makecell{mean\\rank} & \makecell{inv.\ norm.\\rank} \\
    \midrule
\multicolumn{7}{@{}l}{\textit{Unified models}} \\
    \ours & $\times$ & \textbf{69.7}\,{\tiny[67.6, 71.8]} & \textbf{38.3}\,{\tiny[35.7, 40.8]} & \textbf{59.1}\,{\tiny[55.8, 61.9]} & \textbf{1.0} & \textbf{1.00} \\
    \nnfvit~\cite{nnfoundation} & \checkmark & 55.9\,{\tiny[53.5, 58.2]} & 28.1\,{\tiny[26.0, 30.3]} & 40.0\,{\tiny[35.4, 44.3]} & 7.7 & 0.05 \\
    nnF-CNN$^\S$~\cite{nnfoundation} & \checkmark & 66.6\,{\tiny[64.3, 68.9]} & 35.2\,{\tiny[32.8, 37.7]} & 39.6\,{\tiny[34.1, 45.2]} & 4.0 & 0.57 \\
    Curia-MAE$^\S$~\cite{curiamae} & $\times$ & 64.0\,{\tiny[61.7, 66.3]} & 33.6\,{\tiny[31.2, 35.9]} & 50.5\,{\tiny[45.8, 54.8]} & 4.0 & 0.57 \\
\addlinespace[3pt]
\multicolumn{7}{@{}l}{\textit{Specialists}} \\
    SPECTRE~\cite{spectre} & $\times$ & 66.0\,{\tiny[63.8, 68.1]} & 34.0\,{\tiny[31.6, 36.4]} & 53.3\,{\tiny[49.3, 57.1]} & 2.7 & 0.76 \\
    VoxelFM~\cite{voxelfm} & $\times$ & 63.3\,{\tiny[60.9, 65.5]} & 32.8\,{\tiny[30.4, 35.1]} & 53.3\,{\tiny[49.3, 56.7]} & 4.3 & 0.52 \\
    Merlin$^\S$~\cite{merlin} & $\times$ & 61.0\,{\tiny[58.6, 63.3]} & 30.4\,{\tiny[28.2, 32.7]} & 47.7\,{\tiny[42.8, 52.6]} & 6.0 & 0.29 \\
    Triad CNN $\cdot$ MAE$^\S$~\cite{triad} & $\times$ & 59.4\,{\tiny[57.1, 61.8]} & 30.1\,{\tiny[27.9, 32.3]} & 50.4\,{\tiny[46.0, 54.5]} & 6.3 & 0.24 \\
\addlinespace[3pt]
\multicolumn{7}{@{}l}{\textit{Floor}} \\
    \ours, random weights & $\times$ & 55.2\,{\tiny[52.8, 57.5]} & 27.4\,{\tiny[25.4, 29.5]} & 42.3\,{\tiny[38.1, 46.4]} & -- & -- \\
\bottomrule
  \end{tabular}
\end{table}

The lead holds in each of the five folds of 198 to 208 validation studies: \ours has the highest lesion-positive Dice and lesion F1 in every fold, with an SD over folds of 2.2 Dice and 3.7 F1 points. Its per-fold paired 95\% CI against the other seven encoders excludes 0 in 34 of 35 Dice and all 35 F1 comparisons, the exception being \nnfcnn in fold 1 (Dice), and against the random-weight floor in all 10.

\ours also predicts the fewest false-positive components per study, about 5 against 7.5 to 16 for the others, and has the highest competition performance metric (CPM), the mean sensitivity of the free-response ROC (FROC) at 1/8 to 8 false positives per study (\cref{tab:supp-pet-froc}). SPECTRE and Triad CNN $\cdot$ MAE come within two points of its CPM and reach a slightly higher sensitivity at 0.5 false positives per study, none significantly.

\begin{table}[!htbp]
\centering\footnotesize\setlength{\tabcolsep}{4pt}
\caption{\textbf{FDG PET: paired differences, false positives and detection} (autoPET, five folds pooled, 1,014 studies, 8,620 lesions). $\Delta$: encoder minus \ours, so a negative $\Delta$ favors \ours, with the paired 95\% CI below (\cref{sec:supp-impl}). FP/study: false-positive components per study. CPM: competition performance metric. Sensitivity (\%) at 0.5, 1 and 2 false positives per study, with the paired $\Delta$ below. $^\S$: UNet3D head matched to UNETR.}
\label{tab:supp-pet-froc}
\begin{tabular}{@{}l cc c c ccc@{}}
\toprule
 & \multicolumn{2}{c}{$\Delta$ vs \ours} & & & \multicolumn{3}{c}{sensitivity at FP/study} \\
\cmidrule(lr){2-3}\cmidrule(l){6-8}
encoder & Dice lesion$+$ & lesion F1 & FP/study & CPM & 0.5 & 1 & 2 \\
\midrule
\multicolumn{8}{@{}l}{\textit{Unified models}} \\
\ours & ref & ref & 5.0 & 36.1 & 23.1 & 34.7 & 47.3 \\
\nnfvit~\cite{nnfoundation} & \makecell{$-$13.9\\[-2pt]{\tiny[$-$15.4, $-$12.3]}} & \makecell{$-$19.2\\[-2pt]{\tiny[$-$21.7, $-$16.6]}} & 15.9 & 27.5 & \makecell{15.2\\[-2pt]{\tiny$-$7.9 [$-$13.1, $-$0.3]}} & \makecell{25.7\\[-2pt]{\tiny$-$9.0 [$-$15.2, $-$3.8]}} & \makecell{35.4\\[-2pt]{\tiny$-$12.0 [$-$17.0, $-$7.8]}} \\
nnF-CNN$^\S$~\cite{nnfoundation} & \makecell{$-$3.1\\[-2pt]{\tiny[$-$4.1, $-$2.0]}} & \makecell{$-$19.6\\[-2pt]{\tiny[$-$23.1, $-$15.6]}} & 15.7 & 27.8 & \makecell{14.4\\[-2pt]{\tiny$-$8.6 [$-$13.4, $-$3.1]}} & \makecell{25.4\\[-2pt]{\tiny$-$9.3 [$-$15.0, $-$4.1]}} & \makecell{37.4\\[-2pt]{\tiny$-$9.9 [$-$14.8, $-$5.3]}} \\
Curia-MAE$^\S$~\cite{curiamae} & \makecell{$-$5.7\\[-2pt]{\tiny[$-$6.9, $-$4.5]}} & \makecell{$-$8.6\\[-2pt]{\tiny[$-$11.0, $-$6.3]}} & 8.7 & 32.1 & \makecell{18.4\\[-2pt]{\tiny$-$4.6 [$-$7.1, $-$1.0]}} & \makecell{31.1\\[-2pt]{\tiny$-$3.6 [$-$6.6, $-$0.6]}} & \makecell{42.2\\[-2pt]{\tiny$-$5.1 [$-$8.0, $-$2.0]}} \\
\addlinespace[3pt]
\multicolumn{8}{@{}l}{\textit{Specialists}} \\
SPECTRE~\cite{spectre} & \makecell{$-$3.7\\[-2pt]{\tiny[$-$4.7, $-$2.8]}} & \makecell{$-$5.8\\[-2pt]{\tiny[$-$7.3, $-$4.2]}} & 7.5 & 34.5 & \makecell{23.7\\[-2pt]{\tiny+0.6 [$-$2.0, 3.5]}} & \makecell{33.7\\[-2pt]{\tiny$-$1.0 [$-$4.2, 1.7]}} & \makecell{44.5\\[-2pt]{\tiny$-$2.9 [$-$6.0, $-$1.0]}} \\
VoxelFM~\cite{voxelfm} & \makecell{$-$6.5\\[-2pt]{\tiny[$-$7.8, $-$5.2]}} & \makecell{$-$5.9\\[-2pt]{\tiny[$-$7.4, $-$4.4]}} & 7.5 & 32.9 & \makecell{20.2\\[-2pt]{\tiny$-$2.9 [$-$4.5, 0.8]}} & \makecell{30.2\\[-2pt]{\tiny$-$4.5 [$-$7.2, $-$1.6]}} & \makecell{42.2\\[-2pt]{\tiny$-$5.1 [$-$8.6, $-$3.4]}} \\
Merlin$^\S$~\cite{merlin} & \makecell{$-$8.8\\[-2pt]{\tiny[$-$10.1, $-$7.5]}} & \makecell{$-$11.5\\[-2pt]{\tiny[$-$14.2, $-$8.5]}} & 11.3 & 29.0 & \makecell{17.0\\[-2pt]{\tiny$-$6.0 [$-$8.6, $-$2.7]}} & \makecell{25.2\\[-2pt]{\tiny$-$9.5 [$-$12.4, $-$6.6]}} & \makecell{37.1\\[-2pt]{\tiny$-$10.2 [$-$14.2, $-$7.6]}} \\
Triad CNN $\cdot$ MAE$^\S$~\cite{triad} & \makecell{$-$10.3\\[-2pt]{\tiny[$-$11.7, $-$8.9]}} & \makecell{$-$8.7\\[-2pt]{\tiny[$-$10.6, $-$6.7]}} & 8.5 & 34.4 & \makecell{24.9\\[-2pt]{\tiny+1.8 [$-$2.0, 6.2]}} & \makecell{33.2\\[-2pt]{\tiny$-$1.4 [$-$6.0, 2.6]}} & \makecell{42.3\\[-2pt]{\tiny$-$5.0 [$-$9.1, $-$1.8]}} \\
\addlinespace[3pt]
\multicolumn{8}{@{}l}{\textit{Floor}} \\
\ours, random weights & \makecell{$-$14.5\\[-2pt]{\tiny[$-$16.0, $-$13.0]}} & \makecell{$-$16.8\\[-2pt]{\tiny[$-$19.2, $-$14.4]}} & 14.6 & 28.6 & \makecell{18.1\\[-2pt]{\tiny$-$5.0 [$-$8.2, $-$1.8]}} & \makecell{25.9\\[-2pt]{\tiny$-$8.8 [$-$11.6, $-$4.9]}} & \makecell{37.0\\[-2pt]{\tiny$-$10.4 [$-$14.4, $-$8.1]}} \\
\bottomrule
\end{tabular}
\end{table}

\paragraph{Protocols.} The five folds are the patient-level splits of nnU-Net~\cite{nnunet}: each trains a head on 806 to 816 studies and scores the other 198 to 208, 8,620 lesions in all. Each encoder reads PET on its own input scale. The log-SUV volume, $\log(1+\min(\mathrm{SUV},100))/\log 101$ with SUV the standardized uptake value, goes through one fixed affine map per encoder and fold, fitted on 20 of the fold's training studies so that PET patches match the mean and spread of that encoder's input on the co-registered CT, and is left unclipped. Curia-MAE takes the per-image z-score of its pretraining, which is the input rule of \nnfcnn, so the two share their map. At each end of a volume, the run of slices with less than 1\% body and 3 more slices are set to SUV 0 and background in training, validation and scoring alike. The blanking keeps the hot voxels of these noisy end slices out of the image branch of the UNETR~\cite{unetr} decoder. In folds 0 and 4 the blanked slices hold no validation lesion voxel; in folds 1 to 3 they hold a few, and the 22 lesions that lie entirely inside them leave the reference for every encoder alike. The ViTs use the UNETR decoder of \maincref{sec:protocol}. The CNNs use a UNet3D~\cite{cicek2016unet3d} matched to it: 9.6M trainable parameters (against 13.5M and 15.2M for their standard decoders), a deepest trainable level at stride 8 (the stride-16 and stride-32 features are resampled into it) and the same patch size. All eight encoders use $64{\times}192{\times}160$ patches. Dice is the mean over the lesion-positive studies, and over all studies with the nnU-Net rule: an empty prediction on a lesion-free study is skipped, and any false positive there scores 0. Lesion F1 uses 26-connected components. A lesion is found if any predicted voxel overlaps it, and counts are pooled over studies. The pooled values cover all 1,014 studies, with 95\% intervals from a bootstrap that resamples studies within each fold and shares one resample index set across encoders, so differences are paired. A false positive is a predicted 26-connected component that touches no reference voxel, at the default threshold (margin 0).

\subsection{Label efficiency}
\label{sec:supp-le}
This subsection gives the label-efficiency sweeps of \maincref{sec:label-eff} (\cref{tab:supp-le,tab:supp-ts-le}) and the per-structure values behind the two maps of \maincref{fig:r2} (\cref{tab:supp-anat-maps}). Against \nnfvit, the lead of \ours grows as labels shrink on three of the four sweeps, and on TotalSegmentator CT a tenth of the labels matches SPECTRE trained on all of them.

On ACDC, MSD Hippocampus and the ISLES'24 circle of Willis, the gap between \ours and \nnfvit widens as labels shrink (\cref{tab:supp-le}). On MSD Hippocampus, near its ceiling for every encoder, it stays small. A quarter of the labels brings \ours within half a Dice point of \nnfvit trained on all of them on ACDC (88.1 against 88.2) and MSD Hippocampus (88.4 against 88.9). A tenth does the same on AISD (16.8 against 16.9) and passes it on ISLES'24 (41.9 against 39.8). AISD is the exception to the trend: the gap to \nnfvit narrows as labels shrink, SPECTRE leads at 5 and 10\% of the labels, and \ours and SPECTRE are level at 25\%.

\begin{table}[!htbp]
\centering\footnotesize\setlength{\tabcolsep}{6pt}
\caption{\textbf{Label efficiency with the UNETR decoder} (Dice \%, frozen encoders, mean $\pm$ SD over three folds of the sweep's own splits, so its 100\% values differ from \cref{tab:mri-seg,tab:ct-seg,tab:isles}). Columns: share of the training labels. Under each dataset: its training cases per fold at each share. CoW: circle of Willis. NCCT: non-contrast CT.}
\label{tab:supp-le}
\begin{tabular}{@{}ll ccccc@{}}
\toprule
dataset & encoder & 5\% & 10\% & 25\% & 50\% & 100\% \\
\midrule
\makecell[l]{ACDC~\cite{data_acdc}\\[-1pt]{\tiny MRI, cases 8/16/40/80/160}} & \ours & 74.7{\tiny$\pm$4.7} & 83.9{\tiny$\pm$0.8} & 88.1{\tiny$\pm$0.7} & 89.5{\tiny$\pm$1.0} & 90.9{\tiny$\pm$0.9} \\
 & \nnfvit~\cite{nnfoundation} & 58.9{\tiny$\pm$6.4} & 74.0{\tiny$\pm$1.4} & 82.1{\tiny$\pm$1.4} & 85.7{\tiny$\pm$1.2} & 88.2{\tiny$\pm$1.1} \\
\midrule
\makecell[l]{MSD Hippocampus~\cite{data_msd}\\[-1pt]{\tiny MRI, cases 10/21/52/104/208}} & \ours & 86.9{\tiny$\pm$0.4} & 87.9{\tiny$\pm$0.4} & 88.4{\tiny$\pm$0.2} & 88.8{\tiny$\pm$0.6} & 89.3{\tiny$\pm$0.5} \\
 & \nnfvit~\cite{nnfoundation} & 85.6{\tiny$\pm$0.4} & 87.0{\tiny$\pm$0.6} & 87.6{\tiny$\pm$0.4} & 88.2{\tiny$\pm$0.4} & 88.9{\tiny$\pm$0.4} \\
\midrule
\makecell[l]{ISLES'24 CoW NCCT~\cite{data_isles24}\\[-1pt]{\tiny CT, cases 6/12/30/59/118}} & \ours & 37.8{\tiny$\pm$4.5} & 41.9{\tiny$\pm$5.3} & 46.3{\tiny$\pm$3.3} & 50.5{\tiny$\pm$3.0} & 54.1{\tiny$\pm$3.3} \\
 & \nnfvit~\cite{nnfoundation} & 9.7{\tiny$\pm$3.3} & 18.0{\tiny$\pm$3.3} & 27.3{\tiny$\pm$1.8} & 33.7{\tiny$\pm$1.4} & 39.8{\tiny$\pm$1.4} \\
 & SPECTRE~\cite{spectre} & 23.7{\tiny$\pm$3.1} & 29.9{\tiny$\pm$4.7} & 36.9{\tiny$\pm$1.2} & 43.5{\tiny$\pm$2.7} & 46.5{\tiny$\pm$3.4} \\
\midrule
\makecell[l]{AISD~\cite{data_aisd}\\[-1pt]{\tiny CT, cases 14/27/68/136/272}} & \ours & 10.3{\tiny$\pm$0.8} & 16.8{\tiny$\pm$6.4} & 31.4{\tiny$\pm$2.2} & 39.6{\tiny$\pm$0.9} & 42.7{\tiny$\pm$2.2} \\
 & \nnfvit~\cite{nnfoundation} & 5.2{\tiny$\pm$0.7} & 9.0{\tiny$\pm$2.3} & 15.0{\tiny$\pm$3.6} & 16.4{\tiny$\pm$2.2} & 16.9{\tiny$\pm$1.0} \\
 & SPECTRE~\cite{spectre} & 14.5{\tiny$\pm$4.1} & 23.0{\tiny$\pm$6.0} & 31.7{\tiny$\pm$4.3} & 37.2{\tiny$\pm$3.0} & 40.5{\tiny$\pm$3.3} \\
\bottomrule
\end{tabular}
\end{table}

On TotalSegmentator CT, \ours leads at every share (\cref{tab:supp-ts-le}). With a tenth of the labels, it matches SPECTRE trained on all of them ($+0.9$ [$-0.4$, $+2.2$] Dice, paired over the 34 test cases) and passes VoxelFM ($+1.5$ [$+0.1$, $+3.2$]), two encoders that saw TotalSegmentator images in pretraining.

\begin{table}[!htbp]
\centering\small\setlength{\tabcolsep}{6pt}
\caption{\textbf{Label efficiency on TotalSegmentator CT} (Dice \%, frozen encoders, class-averaged over the 34 official test cases, 95\% bootstrap CI beside each value). Columns: share of the training labels, with the number of training cases; the 100\% column repeats \cref{tab:ct-seg}. Bold: best per column. $^\dagger$: images (not labels) of the test set are in the encoder's pretraining; CT-Instruct also trained on its segmentation labels.}
\label{tab:supp-ts-le}
\begin{tabular}{@{}l ccc@{}}
\toprule
 & 1\% (3 cases) & 10\% (29) & 100\% (293) \\
\midrule
\multicolumn{4}{@{}l}{\textit{Unified models}} \\
\ours & \textbf{76.8}\,{\tiny[71.2, 81.7]} & \textbf{88.4}\,{\tiny[84.8, 91.5]} & \textbf{91.5}\,{\tiny[88.0, 94.2]} \\
\nnfvit~\cite{nnfoundation} & 70.9\,{\tiny[63.8, 77.3]} & 84.5\,{\tiny[79.6, 88.5]} & 89.5\,{\tiny[85.7, 92.6]} \\
\nnfcnn~\cite{nnfoundation} & 72.0\,{\tiny[66.1, 77.7]} & 82.2\,{\tiny[76.3, 87.2]} & 89.4\,{\tiny[85.6, 92.6]} \\
\addlinespace[3pt]
\multicolumn{4}{@{}l}{\textit{Specialists}} \\
Merlin~\cite{merlin} & 70.4\,{\tiny[63.3, 76.6]} & 83.8\,{\tiny[79.4, 87.7]} & 88.8\,{\tiny[85.2, 91.9]} \\
CT-Instruct~\cite{ctinstruct} & 53.0$^\dagger$\,{\tiny[44.8, 60.4]} & 71.8$^\dagger$\,{\tiny[65.6, 77.6]} & 82.5$^\dagger$\,{\tiny[77.6, 86.8]} \\
SPECTRE~\cite{spectre} & 69.7$^\dagger$\,{\tiny[62.9, 75.5]} & 80.5$^\dagger$\,{\tiny[75.6, 86.3]} & 87.6$^\dagger$\,{\tiny[83.8, 92.3]} \\
VoxelFM~\cite{voxelfm} & 66.0$^\dagger$\,{\tiny[58.9, 72.3]} & 79.9$^\dagger$\,{\tiny[74.9, 86.1]} & 86.9$^\dagger$\,{\tiny[83.1, 91.9]} \\
\bottomrule
\end{tabular}
\end{table}

Per structure, \ours leads \nnfvit on all three ACDC structures at every label share, each margin significant (\cref{tab:supp-anat-maps}). On TotalSegmentator CT it leads SPECTRE on 16 of 20 structures with three training cases (6 significantly), on 19 with 29 and on 18 with 293, and it is never significantly behind.

\begin{table}[!htbp]
\centering\footnotesize\setlength{\tabcolsep}{4pt}
\caption{\textbf{Per-structure label efficiency} (Dice \%, frozen encoders), the values behind the two maps of \maincref{fig:r2}. (a) ACDC~\cite{data_acdc}, UNETR, 120 validation cases over three folds. (b) TotalSegmentator CT~\cite{totalseg}, official test cases. RV, LV: right and left ventricle. $\Delta$: paired mean difference, \ours minus the baseline. Bold: its paired 95\% CI excludes 0 after Holm correction within the panel (\cref{sec:supp-impl}). $^\dagger$: images (not labels) of the test set are in SPECTRE's pretraining.}
\label{tab:supp-anat-maps}
\begin{tabular}{@{}l ccc c ccc c ccc@{}}
\toprule
\multicolumn{12}{@{}l}{(a) ACDC, \ours vs.\ \nnfvit} \\
 & \multicolumn{3}{c}{5\% (8 cases)} && \multicolumn{3}{c}{25\% (40)} && \multicolumn{3}{c}{100\% (160)} \\
\cmidrule{2-4}\cmidrule{6-8}\cmidrule{10-12}
structure & \ours & \nnfvit & $\Delta$ && \ours & \nnfvit & $\Delta$ && \ours & \nnfvit & $\Delta$ \\
\midrule
RV & 67.3 & 52.5 & $\mathbf{+14.8}$ && 85.9 & 77.9 & $\mathbf{+8.0}$ && 89.8 & 85.1 & $\mathbf{+4.7}$ \\
Myocardium & 74.8 & 55.5 & $\mathbf{+19.3}$ && 85.3 & 79.0 & $\mathbf{+6.3}$ && 88.7 & 86.8 & $\mathbf{+1.9}$ \\
LV & 82.1 & 68.7 & $\mathbf{+13.4}$ && 92.9 & 89.5 & $\mathbf{+3.5}$ && 94.2 & 92.7 & $\mathbf{+1.5}$ \\
\midrule
\multicolumn{12}{@{}l}{(b) TotalSegmentator CT, \ours vs.\ SPECTRE$^\dagger$} \\
 & \multicolumn{3}{c}{1\% (3 cases)} && \multicolumn{3}{c}{10\% (29)} && \multicolumn{3}{c}{100\% (293)} \\
\cmidrule{2-4}\cmidrule{6-8}\cmidrule{10-12}
structure & \ours & SPECTRE & $\Delta$ && \ours & SPECTRE & $\Delta$ && \ours & SPECTRE & $\Delta$ \\
\midrule
Liver & 84.4 & 81.8 & $+2.6$ && 90.5 & 86.9 & $+3.5$ && 93.0 & 91.4 & $+1.6$ \\
Spleen & 80.0 & 80.6 & $-0.6$ && 96.0 & 93.1 & $+2.9$ && 96.4 & 97.4 & $-1.0$ \\
Kidney R & 71.7 & 78.2 & $-6.5$ && 84.2 & 83.7 & $+0.5$ && 86.7 & 84.9 & $+1.8$ \\
Kidney L & 71.0 & 77.2 & $-6.1$ && 87.9 & 84.6 & $+3.3$ && 89.6 & 88.2 & $+1.5$ \\
Pancreas & 75.6 & 52.5 & $\mathbf{+23.2}$ && 90.2 & 76.2 & $\mathbf{+14.0}$ && 92.3 & 86.3 & $\mathbf{+6.0}$ \\
Gallbladder & 51.7 & 27.2 & $\mathbf{+24.5}$ && 75.9 & 44.4 & $\mathbf{+31.5}$ && 79.3 & 69.1 & $+10.2$ \\
Stomach & 78.7 & 67.9 & $+10.8$ && 90.1 & 86.2 & $+3.9$ && 96.3 & 94.0 & $+2.3$ \\
Duodenum & 51.4 & 42.9 & $+8.5$ && 74.9 & 62.1 & $\mathbf{+12.8}$ && 78.7 & 72.2 & $+6.5$ \\
Small bowel & 74.5 & 64.1 & $\mathbf{+10.4}$ && 87.5 & 73.1 & $\mathbf{+14.4}$ && 91.2 & 82.3 & $\mathbf{+8.9}$ \\
Colon & 78.2 & 62.7 & $\mathbf{+15.5}$ && 87.4 & 77.0 & $\mathbf{+10.3}$ && 90.4 & 86.3 & $\mathbf{+4.0}$ \\
Bladder & 72.1 & 65.5 & $+6.6$ && 85.8 & 79.0 & $\mathbf{+6.8}$ && 89.0 & 85.3 & $\mathbf{+3.7}$ \\
Prostate & 42.6 & 44.8 & $-2.1$ && 62.2 & 64.6 & $-2.4$ && 70.2 & 75.9 & $-5.7$ \\
T12 & 79.6 & 71.4 & $+8.2$ && 91.5 & 83.9 & $+7.6$ && 97.6 & 91.2 & $\mathbf{+6.4}$ \\
L1 & 87.1 & 82.0 & $+5.1$ && 97.2 & 88.8 & $\mathbf{+8.4}$ && 98.8 & 92.3 & $\mathbf{+6.5}$ \\
L2 & 90.2 & 83.4 & $+6.8$ && 94.5 & 88.0 & $+6.5$ && 98.8 & 91.7 & $\mathbf{+7.1}$ \\
L3 & 83.9 & 78.5 & $+5.4$ && 93.9 & 83.0 & $\mathbf{+10.9}$ && 93.8 & 87.1 & $\mathbf{+6.7}$ \\
L4 & 92.7 & 80.8 & $\mathbf{+11.9}$ && 97.3 & 85.4 & $\mathbf{+11.9}$ && 96.7 & 92.9 & $\mathbf{+3.8}$ \\
L5 & 90.9 & 86.8 & $+4.1$ && 98.5 & 87.9 & $\mathbf{+10.7}$ && 98.6 & 95.8 & $\mathbf{+2.8}$ \\
S1 vertebra & 90.7 & 88.0 & $+2.7$ && 96.4 & 92.5 & $\mathbf{+4.0}$ && 96.6 & 94.8 & $\mathbf{+1.8}$ \\
Sacrum & 96.8 & 84.2 & $\mathbf{+12.6}$ && 97.4 & 94.3 & $\mathbf{+3.1}$ && 97.6 & 96.0 & $\mathbf{+1.6}$ \\
\bottomrule
\end{tabular}
\end{table}

\paragraph{Protocols.} The sweep of \cref{tab:supp-le} trains the UNETR head on one nested subset of the training cases per share and fold, the same for every encoder, and runs SPECTRE~\cite{spectre} on the CT datasets only. Each encoder reads its own input. The sweep of \cref{tab:supp-ts-le} trains the decoder heads of \cref{tab:ct-seg} for every encoder of that table on 1\%, 10\% and 100\% of the TotalSegmentator CT training cases and scores the 34 official test cases. Per structure (\cref{tab:supp-anat-maps}), a structure absent from a case and left empty by one model only scores 100 for that model, and cases where both are undefined are dropped.

\FloatBarrier
\section{Robustness}
\label{sec:supp-robust}\label{sec:supp-probes}
This section tests whether the frozen tokens hold under shifts of intensity, contrast and modality (\maincref{sec:res-robust}): synthetic intensity remaps (\cref{sec:supp-remap}), cross-contrast correspondence (\cref{sec:supp-corr}), vessels from non-contrast CT (\cref{sec:supp-lvo}) and MRI-to-CT registration (\cref{sec:supp-registration}). The last three give the values behind the three panels of \maincref{fig:r3}.

\subsection{Intensity remapping}
\label{sec:supp-remap}
\ours tokens change the least under intensity remapping on MRI, CT and PET, against all ten other encoders on the modalities each one reads, and every paired interval excludes 0 (\cref{tab:supp-robust}). The lead covers PET, which \ours never sees and nnFoundation pretrains on, and holds when the remaps act on each encoder's normalized input, past its per-scan percentile rescaling or z-scoring. Under Gaussian noise on CT, SPECTRE, which trains with noise augmentation, and \nnfcnn, whose tokens are 64\,mm cells, keep more.

\begin{table}[!htbp]
\centering\small\setlength{\tabcolsep}{2.5pt}
\caption{\textbf{Token robustness to intensity remapping} (240 patches per modality from TotalSeg-MRI, AMOS22 CT and autoPET FDG PET). Robustness: floor-corrected token similarity, as a fraction (1 = unchanged, 0 = unrelated), with the seven remaps on the raw intensities or on the normalized input. Floor $f$: the similarity of clean patches of two scans. CT noise: median raw token cosine under Gaussian noise on CT. 95\% cluster-bootstrap CI below. --: modality not read by the encoder. Bold: best per column.}
\label{tab:supp-robust}
\resizebox{\ifdim\width>\linewidth\linewidth\else\width\fi}{!}{%
\begin{tabular}{l ccc ccc ccc c}
\toprule
 & \multicolumn{3}{c}{remaps on raw intensities} & \multicolumn{3}{c}{remaps on the normalized input} & \multicolumn{3}{c}{floor $f$} & CT noise \\
\cmidrule(lr){2-4}\cmidrule(lr){5-7}\cmidrule(lr){8-10}
encoder & MRI & CT & PET & MRI & CT & PET & MRI & CT & PET & cosine \\
\midrule
\multicolumn{11}{@{}l}{\textit{Unified models}} \\
\ours & \makecell{\textbf{0.892}\\[-2pt]{\tiny[0.88, 0.90]}} & \makecell{\textbf{0.868}\\[-2pt]{\tiny[0.86, 0.88]}} & \makecell{\textbf{0.903}\\[-2pt]{\tiny[0.90, 0.91]}} & \makecell{\textbf{0.918}\\[-2pt]{\tiny[0.91, 0.93]}} & \makecell{\textbf{0.878}\\[-2pt]{\tiny[0.87, 0.88]}} & \makecell{\textbf{0.903}\\[-2pt]{\tiny[0.90, 0.91]}} & $-$0.01 & 0.00 & 0.01 & 0.810 \\
nnF-ViT~\cite{nnfoundation} & \makecell{0.680\\[-2pt]{\tiny[0.66, 0.70]}} & \makecell{0.667\\[-2pt]{\tiny[0.65, 0.68]}} & \makecell{0.545\\[-2pt]{\tiny[0.52, 0.57]}} & \makecell{0.644\\[-2pt]{\tiny[0.63, 0.66]}} & \makecell{0.629\\[-2pt]{\tiny[0.61, 0.65]}} & \makecell{0.545\\[-2pt]{\tiny[0.52, 0.57]}} & 0.00 & 0.01 & 0.06 & 0.606 \\
nnF-CNN~\cite{nnfoundation} & \makecell{0.816\\[-2pt]{\tiny[0.78, 0.85]}} & \makecell{0.832\\[-2pt]{\tiny[0.81, 0.85]}} & \makecell{0.712\\[-2pt]{\tiny[0.69, 0.73]}} & \makecell{0.803\\[-2pt]{\tiny[0.77, 0.84]}} & \makecell{0.831\\[-2pt]{\tiny[0.81, 0.85]}} & \makecell{0.712\\[-2pt]{\tiny[0.69, 0.73]}} & 0.19 & 0.25 & 0.33 & \textbf{0.975} \\
\addlinespace[3pt]
\multicolumn{11}{@{}l}{\textit{Specialists}} \\
SPECTRE~\cite{spectre} & -- & \makecell{0.672\\[-2pt]{\tiny[0.65, 0.69]}} & \makecell{0.452\\[-2pt]{\tiny[0.43, 0.47]}} & -- & \makecell{0.657\\[-2pt]{\tiny[0.64, 0.68]}} & \makecell{0.452\\[-2pt]{\tiny[0.43, 0.47]}} & -- & 0.01 & 0.03 & 0.910 \\
VoxelFM~\cite{voxelfm} & -- & \makecell{0.477\\[-2pt]{\tiny[0.45, 0.50]}} & \makecell{0.548\\[-2pt]{\tiny[0.53, 0.56]}} & -- & \makecell{0.499\\[-2pt]{\tiny[0.48, 0.52]}} & \makecell{0.548\\[-2pt]{\tiny[0.53, 0.56]}} & -- & 0.01 & 0.04 & 0.506 \\
Merlin~\cite{merlin} & -- & \makecell{0.495\\[-2pt]{\tiny[0.47, 0.52]}} & \makecell{0.510\\[-2pt]{\tiny[0.48, 0.54]}} & -- & \makecell{0.494\\[-2pt]{\tiny[0.47, 0.52]}} & \makecell{0.510\\[-2pt]{\tiny[0.48, 0.54]}} & -- & 0.06 & 0.23 & 0.579 \\
CT-Instruct~\cite{ctinstruct} & -- & \makecell{0.452\\[-2pt]{\tiny[0.39, 0.50]}} & \makecell{0.075\\[-2pt]{\tiny[$-$0.09, 0.21]}} & -- & \makecell{0.199\\[-2pt]{\tiny[0.12, 0.27]}} & \makecell{0.075\\[-2pt]{\tiny[$-$0.09, 0.21]}} & -- & 0.09 & 0.67 & 0.513 \\
Triad CNN $\cdot$ MAE~\cite{triad} & \makecell{0.628\\[-2pt]{\tiny[0.60, 0.65]}} & -- & -- & \makecell{0.619\\[-2pt]{\tiny[0.59, 0.65]}} & -- & -- & 0.58 & -- & -- & -- \\
Triad CNN $\cdot$ SimMIM~\cite{triad} & \makecell{0.676\\[-2pt]{\tiny[0.65, 0.70]}} & -- & -- & \makecell{0.632\\[-2pt]{\tiny[0.60, 0.66]}} & -- & -- & 0.77 & -- & -- & -- \\
Triad Swin $\cdot$ MAE~\cite{triad} & \makecell{0.621\\[-2pt]{\tiny[0.59, 0.64]}} & -- & -- & \makecell{0.646\\[-2pt]{\tiny[0.62, 0.67]}} & -- & -- & 0.40 & -- & -- & -- \\
Triad Swin $\cdot$ SimMIM~\cite{triad} & \makecell{0.680\\[-2pt]{\tiny[0.66, 0.70]}} & -- & -- & \makecell{0.689\\[-2pt]{\tiny[0.67, 0.71]}} & -- & -- & 0.09 & -- & -- & -- \\
\bottomrule
\end{tabular}}
\end{table}

\paragraph{Protocols.} Each modality gives 240 patches, six from each of 40 scans of TotalSeg-MRI~\cite{dantonoli2025totalsegmri}, AMOS22~\cite{amos} CT and autoPET~\cite{autopet} FDG PET: a crop at 2\,mm around a random foreground voxel, centered at the same voxel for every encoder, 224\,mm wide, or 256\,mm for the stride-32 encoders \nnfcnn, Merlin, CT-Instruct and Triad. The seven remaps are GIN~\cite{gin} at $\alpha$ 0.15 and 0.3, a histogram shift, gamma 0.7 and 1.5, inversion and Gaussian noise of $\sigma$ 0.1. Robustness is the mean over the remaps of $(c-f)/(1-f)$: $c$ is the median, over a patch's foreground last-layer tokens, of the centered cosine to the clean token, and $f$ is the same cosine between clean patches of two scans at the same grid position, so the score is 1 for unchanged tokens and 0 for unrelated ones. The floor $f$ is near 0 for every ViT and 0.06 to 0.77 for the stride-32 encoders; the 64\,mm cells of \nnfcnn and Triad tile a 256\,mm crop four times per axis. Remaps on raw intensities act before each encoder's own normalization, and remaps on the normalized input act on what the encoder reads, so no normalization undoes them. On PET every encoder reads log-SUV unchanged, without the per-encoder affine map of \cref{sec:supp-pet}, so the two settings coincide. The 95\% intervals come from a cluster bootstrap over scans, then patches within a scan. CT-only encoders are scored on CT and PET, MRI-only encoders on MRI.

\subsection{Cross-contrast correspondence}
\label{sec:supp-corr}
A token that keeps its similarity across contrasts must also stay specific to its location. Cross-contrast correspondence, the correspondence panel of \maincref{fig:r3}, tests this: every 8\,mm cell of one image ranks all cells of a co-registered second contrast by token cosine (\cref{tab:supp-correspondence}). \ours ranks the true cell highest of all encoders on both datasets, at 0.975 and 0.991 AUROC, and puts its best match within one cell for half of the MRI cells and two thirds of the CT cells. MIND-SSC~\cite{heinrich2013mindssc}, the classical cross-contrast descriptor, reaches 0.937 and 0.893. \nnfcnn and Triad, whose tokens are 64\,mm cells, cannot place a match within 8\,mm (top-1 at most 0.08 even for identical images) and reach 0.61 to 0.75 AUROC.

\begin{table}[!htbp]
\centering\small\setlength{\tabcolsep}{3pt}
\caption{\textbf{Cross-contrast correspondence} (frozen last-layer tokens, 60 patients per dataset). BraTS-Africa: the six pairs of its four MRI sequences. ISLES'24: NCCT against CTA. AUROC, as a fraction: rank of the true 8\,mm cell among all cells. Top-1: share of cells whose best match lies within 8\,mm. Error: median distance of the best match (mm). 95\% bootstrap CI below. --: modality not read by the encoder. Bold: best encoder per column.}
\label{tab:supp-correspondence}
\resizebox{\ifdim\width>\linewidth\linewidth\else\width\fi}{!}{%
\begin{tabular}{l ccc c ccc}
\toprule
 & \multicolumn{3}{c}{BraTS-Africa (MRI sequences)} && \multicolumn{3}{c}{ISLES'24 (NCCT vs.\ CTA)} \\
\cmidrule(lr){2-4}\cmidrule(lr){6-8}
descriptor & AUROC & top-1 & error (mm) && AUROC & top-1 & error (mm) \\
\midrule
\ours & \makecell{\textbf{0.975}\\[-2pt]{\tiny[0.973, 0.977]}} & \makecell{\textbf{0.504}\\[-2pt]{\tiny[0.491, 0.518]}} & \textbf{16} &  & \makecell{\textbf{0.991}\\[-2pt]{\tiny[0.988, 0.993]}} & \makecell{\textbf{0.664}\\[-2pt]{\tiny[0.610, 0.718]}} & \textbf{14} \\
nnF-ViT~\cite{nnfoundation} & \makecell{0.927\\[-2pt]{\tiny[0.921, 0.932]}} & \makecell{0.449\\[-2pt]{\tiny[0.432, 0.467]}} & 18 &  & \makecell{0.971\\[-2pt]{\tiny[0.964, 0.978]}} & \makecell{0.632\\[-2pt]{\tiny[0.572, 0.693]}} & \textbf{14} \\
nnF-CNN~\cite{nnfoundation} & \makecell{0.629\\[-2pt]{\tiny[0.622, 0.636]}} & \makecell{0.015\\[-2pt]{\tiny[0.014, 0.016]}} & 65 &  & \makecell{0.612\\[-2pt]{\tiny[0.598, 0.625]}} & \makecell{0.009\\[-2pt]{\tiny[0.008, 0.010]}} & 79 \\
SPECTRE~\cite{spectre} & -- & -- & -- &  & \makecell{0.920\\[-2pt]{\tiny[0.911, 0.929]}} & \makecell{0.168\\[-2pt]{\tiny[0.148, 0.190]}} & 29 \\
VoxelFM~\cite{voxelfm} & -- & -- & -- &  & \makecell{0.858\\[-2pt]{\tiny[0.848, 0.868]}} & \makecell{0.098\\[-2pt]{\tiny[0.087, 0.109]}} & 39 \\
Merlin~\cite{merlin} & -- & -- & -- &  & \makecell{0.922\\[-2pt]{\tiny[0.916, 0.927]}} & \makecell{0.063\\[-2pt]{\tiny[0.059, 0.067]}} & 42 \\
CT-Instruct~\cite{ctinstruct} & -- & -- & -- &  & \makecell{0.778\\[-2pt]{\tiny[0.773, 0.784]}} & \makecell{0.038\\[-2pt]{\tiny[0.035, 0.041]}} & 49 \\
Triad CNN $\cdot$ MAE~\cite{triad} & \makecell{0.718\\[-2pt]{\tiny[0.713, 0.723]}} & \makecell{0.029\\[-2pt]{\tiny[0.028, 0.029]}} & 47 &  & -- & -- & -- \\
Triad CNN $\cdot$ SimMIM~\cite{triad} & \makecell{0.749\\[-2pt]{\tiny[0.744, 0.754]}} & \makecell{0.033\\[-2pt]{\tiny[0.032, 0.034]}} & 44 &  & -- & -- & -- \\
Triad Swin $\cdot$ MAE~\cite{triad} & \makecell{0.685\\[-2pt]{\tiny[0.680, 0.689]}} & \makecell{0.026\\[-2pt]{\tiny[0.026, 0.027]}} & 49 &  & -- & -- & -- \\
Triad Swin $\cdot$ SimMIM~\cite{triad} & \makecell{0.662\\[-2pt]{\tiny[0.657, 0.666]}} & \makecell{0.020\\[-2pt]{\tiny[0.020, 0.021]}} & 53 &  & -- & -- & -- \\
\midrule
MIND-SSC~\cite{heinrich2013mindssc} & \makecell{0.937\\[-2pt]{\tiny[0.933, 0.942]}} & \makecell{0.289\\[-2pt]{\tiny[0.276, 0.303]}} & 51 &  & \makecell{0.893\\[-2pt]{\tiny[0.881, 0.904]}} & \makecell{0.371\\[-2pt]{\tiny[0.355, 0.388]}} & 39 \\
raw-intensity patch & \makecell{0.688\\[-2pt]{\tiny[0.678, 0.698]}} & \makecell{0.198\\[-2pt]{\tiny[0.189, 0.206]}} & 55 &  & \makecell{0.913\\[-2pt]{\tiny[0.899, 0.924]}} & \makecell{0.419\\[-2pt]{\tiny[0.400, 0.439]}} & 30 \\
\bottomrule
\end{tabular}}
\end{table}

Per MRI sequence pair, \ours ranks the true cell highest on all six pairs of BraTS-Africa, ahead of \nnfvit, MIND-SSC and the raw-intensity patch, and every paired 95\% interval against them excludes 0 (\cref{tab:supp-corr-pairs}).

\begin{table}[!htbp]
\centering\small\setlength{\tabcolsep}{5pt}
\caption{\textbf{Cross-sequence correspondence per MRI sequence pair} (BraTS-Africa, AUROC as a fraction, 60 patients; the right panel of the correspondence row of \maincref{fig:r3}). Protocol as in \cref{tab:supp-correspondence}. Bold: best per pair.}
\label{tab:supp-corr-pairs}
\begin{tabular}{l cccccc}
\toprule
descriptor & T1--T1c & T1--T2 & T1--FLAIR & T1c--T2 & T1c--FLAIR & T2--FLAIR \\
\midrule
\ours & \textbf{0.983} & \textbf{0.972} & \textbf{0.975} & \textbf{0.970} & \textbf{0.971} & \textbf{0.979} \\
\nnfvit~\cite{nnfoundation} & 0.958 & 0.895 & 0.957 & 0.895 & 0.942 & 0.914 \\
MIND-SSC & 0.948 & 0.950 & 0.939 & 0.916 & 0.917 & 0.953 \\
raw-intensity patch & 0.904 & 0.473 & 0.817 & 0.502 & 0.799 & 0.635 \\
\bottomrule
\end{tabular}
\end{table}

\paragraph{Protocols.} The pairs are the four co-registered sequences of one brain (BraTS-Africa~\cite{data_bratsafrica}: the six pairs of T1, T1c, T2 and FLAIR) and non-contrast CT (NCCT) against CT angiography (CTA) of one head (ISLES'24~\cite{data_isles24}), 60 patients each, at 2\,mm and under each encoder's own normalization. The crop of the second image is shifted by 16, 24 or 32\,mm per axis (random sign), so that crop position cannot solve the match. Every scored 8\,mm cell, inside both crops and in the foreground, takes the token that contains its center and ranks all scored cells of the other image by centered cosine, in both directions. AUROC is the rank of the true cell among all cells (ties count half). Top-1 is the share of cells whose best match lies within one cell (8\,mm), and the error is the median distance of the best match. With identical images, the token size caps top-1 at 1.00 for \ours and \nnfvit, 0.45 for SPECTRE, 0.39 for VoxelFM, 0.20 for Merlin and CT-Instruct and 0.08 for the 64\,mm cells of \nnfcnn and Triad. MIND-SSC uses 12 channels per cell, and the raw-intensity patch the cell's 64 voxels. The 95\% intervals come from a bootstrap over patients. Encoders are scored on the modality they read.

\subsection{Circle of Willis and large-vessel occlusion}
\label{sec:supp-lvo}
ISLES'24, the contrast panel of \maincref{fig:r3}, tests whether the tokens find vessels without the contrast agent that shows them. From NCCT, where the vessels are barely visible, \ours has the best circle-of-Willis Dice by 7.1 points and loses the least to the missing contrast (\cref{tab:isles}). On large-vessel occlusion (LVO) detection from CTA, it has the highest CPM, about 4 points above Merlin~\cite{merlin}, and Merlin, CT-Instruct and SPECTRE are level with it under the paired test. From NCCT it is second, about 2 points behind SPECTRE~\cite{spectre}.

\begin{table}[!htbp]
\centering\footnotesize\setlength{\tabcolsep}{2pt}
\caption{\textbf{Circle of Willis (CoW, Dice \%) and LVO detection (CPM \%) from NCCT and CTA} (ISLES'24, 3 folds, 95\% bootstrap CI below). NCCT: non-contrast CT. CTA: CT angiography. Bold: best per column. Mean rank and inverse normalized rank: over the two NCCT cells, among the seven encoders (the CT, no-contrast group of \maincref{fig:teaser}).}
\label{tab:isles}
\begin{tabular}{l cc c cc c cc}
\toprule
 & \multicolumn{2}{c}{NCCT} && \multicolumn{2}{c}{CTA} && \multicolumn{2}{c}{NCCT, 2 cells} \\
\cmidrule{2-3}\cmidrule{5-6}\cmidrule{8-9}
encoder & CoW & LVO && CoW & LVO && \makecell{mean\\rank} & \makecell{inv.\ norm.\\rank} \\
\midrule
\multicolumn{9}{@{}l}{\textit{Unified models}} \\
\ours      & \makecell{\textbf{54.0}\\[-2pt]{\tiny[51.2, 56.6]}} & \makecell{{19.7}\\[-2pt]{\tiny[14.2, 23.9]}} && \makecell{\textbf{84.8}\\[-2pt]{\tiny[82.4, 86.6]}} & \makecell{\textbf{43.0}\\[-2pt]{\tiny[35.6, 52.7]}} && \textbf{1.5} & \textbf{0.92} \\
nnF-ViT~\cite{nnfoundation}     & \makecell{39.8\\[-2pt]{\tiny[37.1, 42.4]}} & \makecell{15.2\\[-2pt]{\tiny[9.9, 20.5]}} && \makecell{82.0\\[-2pt]{\tiny[79.4, 84.1]}} & \makecell{17.5\\[-2pt]{\tiny[12.4, 23.6]}} && 4.0 & 0.50 \\
nnF-CNN~\cite{nnfoundation}     & \makecell{35.8\\[-2pt]{\tiny[33.1, 38.5]}} & \makecell{10.0\\[-2pt]{\tiny[5.5, 14.4]}} && \makecell{81.6\\[-2pt]{\tiny[78.6, 84.2]}} & \makecell{13.2\\[-2pt]{\tiny[8.6, 18.3]}} && 7.0 & 0.00 \\
\addlinespace[3pt]
\multicolumn{9}{@{}l}{\textit{Specialists}} \\
Merlin~\cite{merlin}      & \makecell{45.9\\[-2pt]{\tiny[43.2, 48.6]}} & \makecell{12.8\\[-2pt]{\tiny[7.1, 19.6]}} && \makecell{81.2\\[-2pt]{\tiny[80.9, 82.8]}} & \makecell{38.8\\[-2pt]{\tiny[28.9, 49.3]}} && 5.0 & 0.33 \\
CT-Instruct~\cite{ctinstruct} & \makecell{39.0\\[-2pt]{\tiny[36.5, 41.4]}} & \makecell{14.6\\[-2pt]{\tiny[9.1, 21.3]}} && \makecell{81.9\\[-2pt]{\tiny[79.6, 83.9]}} & \makecell{34.0\\[-2pt]{\tiny[25.1, 44.1]}} && 5.0 & 0.33 \\
SPECTRE~\cite{spectre}     & \makecell{46.5\\[-2pt]{\tiny[44.0, 48.7]}} & \makecell{\textbf{21.6}\\[-2pt]{\tiny[15.1, 24.2]}} && \makecell{79.7\\[-2pt]{\tiny[73.2, 83.8]}} & \makecell{36.3\\[-2pt]{\tiny[26.9, 48.2]}} && 2.0 & 0.83 \\
VoxelFM~\cite{voxelfm}     & \makecell{46.9\\[-2pt]{\tiny[44.3, 49.3]}} & \makecell{{14.0}\\[-2pt]{\tiny[9.7, 20.3]}}&& \makecell{80.3\\[-2pt]{\tiny[77.9, 82.3]}} & \makecell{{34.8}\\[-2pt]{\tiny[25.5, 47.7]}} && 3.5 & 0.58 \\
\bottomrule
\end{tabular}
\end{table}

\paragraph{Protocols.} ISLES'24 gives circle-of-Willis masks on CTA, a silver standard (\maincref{sec:res-robust}), and marks large-vessel occlusions on CTA. A decoder head on each frozen encoder segments the circle of Willis or predicts an occlusion map, from CTA or from the co-registered NCCT, so LVO detection runs as two tasks, from CTA and from NCCT; \cref{sec:supp-stage-ablation} scores the occlusion region of the same two tasks by Dice. We score LVO detection by CPM as in LUNA16~\cite{luna16}: the mean sensitivity at 1/8, 1/4, 1/2, 1, 2, 4 and 8 false positives per patient, with a hit counted when a predicted component lies within 10\,mm of an annotated occlusion. An occlusion is a short vessel segment, so this task tests whether the tokens localize one small target. The paired tests resample cases within folds for the circle of Willis and patients for LVO.

\subsection{MRI-to-CT registration}
\label{sec:supp-registration}
In MRI-to-CT registration, the modality panel of \maincref{fig:r3}, \ours tokens match anatomy across modalities from both starts (\cref{tab:supp-reg}). Alone, they have the highest Dice, significantly above anatomix~\cite{dey2025anatomix}, a network built for the task, from landmarks and above \nnfvit from the label-free start. Fused with MIND-SSC, they are level with anatomix fused the same way and significantly ahead of MIND-SSC alone, uniGradICON and ANTs SyN.

\begin{table}[!htbp]
\centering\footnotesize\setlength{\tabcolsep}{3.5pt}
\caption{\textbf{MRI-to-CT registration} (the 43 test pairs that show a scored organ; the modality panel of \maincref{fig:r3}). Dice as a fraction, with 95\% CI; MRI\,/\,CT: with that modality fixed. HD95 in mm. Start: affine (aff) or center of mass (com). $\Delta$: the matching \ours row minus the row, in Dice points; $\ast$: its paired 95\% CI excludes 0. $^\ddagger$: no setting meets the folding cap. Ranks: over both starts, among the six fused or own-pipeline rows (\maincref{fig:teaser}). Bold: best rank.}
\label{tab:supp-reg}
\begin{tabular}{@{}l ccccc@{}}
\multicolumn{6}{@{}l}{(a) Landmark start} \\
\toprule
method & Dice & MRI\,/\,CT & folding\,\% & HD95 & $\Delta$ \\
\midrule
\multicolumn{6}{@{}l}{\textit{Unified models}} \\
\ours features & 0.495\,{\tiny[0.440, 0.547]} & 0.493\,/\,0.496 & 0.90 & 35.4 & -- \\
\ours $+$ MIND-SSC & 0.547\,{\tiny[0.489, 0.601]} & 0.555\,/\,0.540 & 0.19 & 34.6 & -- \\
\nnfvit~\cite{nnfoundation} features & 0.491\,{\tiny[0.425, 0.552]} & 0.503\,/\,0.478 & 0.46 & 36.0 & +0.4\,{\tiny[$-$2.4, +3.2]} \\
\nnfvit~\cite{nnfoundation} $+$ MIND-SSC & 0.530\,{\tiny[0.469, 0.586]} & 0.543\,/\,0.517 & 0.25 & 36.3 & +1.7$^\ast$\,{\tiny[+0.4, +3.1]} \\
\addlinespace[3pt]
\multicolumn{6}{@{}l}{\textit{Specialists}} \\
anatomix~\cite{dey2025anatomix} features & 0.439\,{\tiny[0.385, 0.489]} & 0.396\,/\,0.481 & 0.20 & 37.6 & +5.6$^\ast$\,{\tiny[+1.4, +9.7]} \\
anatomix~\cite{dey2025anatomix} $+$ MIND-SSC & 0.556\,{\tiny[0.498, 0.611]} & 0.556\,/\,0.556 & 0.19 & 33.0 & $-$0.9\,{\tiny[$-$2.8, +1.1]} \\
MIND-SSC $+$ ConvexAdam & 0.488\,{\tiny[0.435, 0.536]} & 0.500\,/\,0.477 & 0.90 & 38.3 & +5.9$^\ast$\,{\tiny[+2.3, +9.2]} \\
uniGradICON~\cite{tian2024unigradicon}$^\ddagger$ & 0.409\,{\tiny[0.362, 0.455]} & 0.424\,/\,0.395 & 1.34 & 40.6 & +13.8$^\ast$\,{\tiny[+10.2, +17.3]} \\
ANTs SyN-MI~\cite{avants2008syn} & 0.433\,{\tiny[0.387, 0.476]} & 0.443\,/\,0.423 & 0.02 & 40.0 & +11.4$^\ast$\,{\tiny[+8.6, +14.3]} \\
\addlinespace[3pt]
\multicolumn{6}{@{}l}{\textit{Reference}} \\
pre-alignment only & 0.376\,{\tiny[0.338, 0.412]} & 0.376\,/\,0.377 & 0.00 & 40.7 & -- \\
\bottomrule
\end{tabular}
\par\vspace{1em}
\begin{tabular}{@{}l cccccc c cc@{}}
\multicolumn{10}{@{}l}{(b) Label-free start} \\
\toprule
 & & & & & & && \multicolumn{2}{c}{both starts} \\
\cmidrule(l){9-10}
method & Dice & MRI\,/\,CT & folding\,\% & HD95 & start & $\Delta$ && \makecell{mean\\rank} & \makecell{inv.\ norm.\\rank} \\
\midrule
\multicolumn{10}{@{}l}{\textit{Unified models}} \\
\ours features & 0.246\,{\tiny[0.176, 0.318]} & 0.211\,/\,0.280 & 0.91 & 64.7 & aff & -- &  & -- & -- \\
\ours $+$ MIND-SSC & 0.333\,{\tiny[0.254, 0.412]} & 0.362\,/\,0.304 & 0.34 & 60.0 & aff & -- &  & \textbf{1.5} & \textbf{0.90} \\
\nnfvit~\cite{nnfoundation} features & 0.193\,{\tiny[0.125, 0.265]} & 0.139\,/\,0.246 & 0.66 & 81.6 & aff & +5.3$^\ast$\,{\tiny[+0.1, +10.6]} &  & -- & -- \\
\nnfvit~\cite{nnfoundation} $+$ MIND-SSC & 0.287\,{\tiny[0.208, 0.370]} & 0.290\,/\,0.284 & 0.45 & 68.1 & aff & +4.6$^\ast$\,{\tiny[+0.8, +8.9]} &  & 3.0 & 0.60 \\
\addlinespace[3pt]
\multicolumn{10}{@{}l}{\textit{Specialists}} \\
anatomix~\cite{dey2025anatomix} features & 0.196\,{\tiny[0.141, 0.253]} & 0.123\,/\,0.269 & 0.32 & 80.1 & aff & +5.0\,{\tiny[$-$1.0, +10.9]} &  & -- & -- \\
anatomix~\cite{dey2025anatomix} $+$ MIND-SSC & 0.322\,{\tiny[0.243, 0.402]} & 0.312\,/\,0.333 & 0.25 & 61.5 & aff & +1.1\,{\tiny[$-$4.4, +6.5]} &  & \textbf{1.5} & \textbf{0.90} \\
MIND-SSC $+$ ConvexAdam & 0.217\,{\tiny[0.150, 0.285]} & 0.247\,/\,0.187 & 0.29 & 76.9 & aff & +11.6$^\ast$\,{\tiny[+4.1, +18.6]} &  & 4.0 & 0.40 \\
uniGradICON~\cite{tian2024unigradicon}$^\ddagger$ & 0.172\,{\tiny[0.120, 0.228]} & 0.203\,/\,0.142 & 0.89 & 79.8 & aff & +16.0$^\ast$\,{\tiny[+9.6, +22.4]} &  & 5.5 & 0.10 \\
ANTs SyN-MI~\cite{avants2008syn} & 0.119\,{\tiny[0.080, 0.162]} & 0.105\,/\,0.133 & 1.27 & 82.9 & com & +21.4$^\ast$\,{\tiny[+14.0, +28.5]} &  & 5.5 & 0.10 \\
\addlinespace[3pt]
\multicolumn{10}{@{}l}{\textit{Reference}} \\
pre-alignment only & 0.075\,{\tiny[0.040, 0.115]} & 0.081\,/\,0.069 & 0.00 & 104.0 & com & -- &  & -- & -- \\
\bottomrule
\end{tabular}
\end{table}

\paragraph{Pairs and scoring.}
We register TotalSeg-MRI~\cite{dantonoli2025totalsegmri} to portal-venous TriALS~\cite{data_trials} CT of other patients, both ways, under our own protocol, as neither segmentation dataset has one; TriALS organ labels are TotalSegmentator~\cite{totalseg} predictions. Both scans are resampled to 2\,mm, cropped to the body and placed on one $320{\times}192{\times}192$ grid. Each of the 80 TriALS scans, in sorted order, takes the first unused TotalSeg-MRI scan sharing at least four labeled structures of at least 1\,ml; the first 28 pairs tune, the other 52 test. We score the liver, spleen and both kidneys (median volume 150\,ml to 1.4\,l), a set fixed in advance from the feature-free rows alone: every structure that appears in at least 15 of 29 development pairs of TotalSeg-MRI to TotalSegmentator CT and reaches 0.40 Dice with MIND-SSC and ConvexAdam. Vertebrae are excluded, as the two label sets define them differently; small and hollow structures such as the pancreas, stomach and bowel stay below 0.25 Dice there. Dice is per organ inside the common field of view (0 if moved off the grid), averaged over the organs both scans show, then both directions; 27 tune and 43 test pairs show at least one. HD95 is the 95th-percentile Hausdorff distance. $\Delta$ compares token features alone with \ours features and fused and own-pipeline rows with \ours $+$ MIND-SSC, in Dice points ($100\times$ the difference of fractions), with a paired 95\% bootstrap interval over pairs.

\paragraph{Starts.}
The landmark start isolates the deformable step: an isotropic scale and a translation align the centroids of labeled structures fully inside both scans, removing most of the global offset. The label-free start also tests whether a descriptor finds the global pose (pre-alignment rows): it aligns the body masks' centers of mass, then fits an affine on the row's own descriptor, which each row keeps (aff) or drops (com) on tune.

\paragraph{Descriptors and baselines.}
MIND-SSC~\cite{heinrich2013mindssc}, the classical multimodal descriptor, codes local structure, which MRI and CT share, as patch similarity around each voxel normalized by a local variance estimate, and so is invariant to a global intensity scale. Token features follow DINO-Reg~\cite{song2024dinoreg}: last-layer tokens on an 8\,mm grid (for \ours, teacher tokens after the final LayerNorm), centered and normalized per scan, reduced by one PCA over both scans to $k$ components and interpolated to 2\,mm; \nnfvit tokens and anatomix~\cite{dey2025anatomix} U-Net features take the same steps. Fusion appends 24 token channels, weighted by $w$, to MIND-SSC. MIND-SSC with ConvexAdam is the classical pipeline, and two published pipelines register the images directly: the registration foundation model uniGradICON~\cite{tian2024unigradicon}, zero-shot or with instance optimization, and ANTs SyN~\cite{avants2008syn} under mutual information (ANTs SyN-MI).

\paragraph{Optimization and selection.}
On tune, each method takes its best setting by mean Dice under a 1\% folding cap, folding being the share of fixed body voxels where the deformation folds. ConvexAdam~\cite{siebert2025convexadam} has no learned parameters: a discrete search over coarse-cell shifts (capture range $\pm$64 or $\pm$128\,mm) gives a smooth, inverse-consistent field that 40 Adam steps refine under a diffusion penalty weighted by $\lambda$; the rest follows the authors' abdominal MR-to-CT script. Rescaling every feature image to the pair's total MIND-SSC variance gives $\lambda$ one meaning across descriptors. Each descriptor gets 24 settings: $\lambda \in \{0.1, 0.3, 0.6, 1.5\}$ and both capture ranges, crossed with three MIND-SSC radius and dilation pairs, PCA sizes $k$ or fusion weights $w$. Both \ours rows select $\lambda = 0.3$ and $\pm$64\,mm, with $k = 12$ alone and $w = 0.5$ fused. No uniGradICON setting meets the cap, so the table shows its lowest-folding one ($^\ddagger$), still ranked in the MRI-to-CT registration group of \maincref{fig:teaser}. The folding column comes from the test pairs, where ANTs SyN exceeds the cap from the label-free start.
\FloatBarrier

\section{Labels, lesions and anatomy with no trained head}
\label{sec:supp-analysis}
The main paper reads the frozen tokens of \ours with no trained head (\maincref{fig:analysis}): a k-NN vote recovers
patient labels, the distance to lesion-free tissue finds lesions, and k-means groups anatomy. This section gives the
values behind these three panels and the protocols of the probes. The MRI k-NN values are in \cref{tab:mri-cls}, and
\cref{sec:supp-stage-ablation} scores the stage panel.

\paragraph{k-NN vote and class token.} With no trained head, the class token carries the glioma labels
(\cref{tab:supp-clstok}a). Its k-NN vote reaches 92.3 and 92.8 AUROC on IDH status and WHO grade, against 86.1 and 85.6
for the patch-mean vote of \cref{tab:mri-cls}. On LLD-MMRI, whose crops center on the lesion, the two readouts are
within a point. No other MRI encoder in the comparison has a class token. On STOIC, the CT set of the k-NN panel, the
patch-mean vote ranks \ours first on both targets, ahead of VoxelFM and \nnfvit (\cref{tab:supp-clstok}b).

\begin{table}[!htbp]
\centering\small\setlength{\tabcolsep}{6pt}
\caption{\textbf{k-NN vote with no trained head} (AUROC \%, the training-free vote of \cref{tab:mri-cls}). (a) The
class token of \ours against its patch mean on the MRI targets. (b) The patch-mean vote on STOIC, the CT set of
\maincref{fig:analysis}. Bold: best per column.}
\label{tab:supp-clstok}
\begin{tabular}[t]{l ccc}
\toprule
\multicolumn{4}{@{}l}{(a) MRI, \ours} \\
readout & IDH & grade & LLD \\
\midrule
patch mean & 86.1 & 85.6 & \textbf{79.7} \\
class token & \textbf{92.3} & \textbf{92.8} & 78.8 \\
\bottomrule
\end{tabular}\hspace{3em}%
\begin{tabular}[t]{l cc}
\toprule
\multicolumn{3}{@{}l}{(b) STOIC, patch mean} \\
encoder & COVID-19 & severe \\
\midrule
\ours & \textbf{73.8} & \textbf{68.8} \\
VoxelFM~\cite{voxelfm} & 71.2 & 68.0 \\
nnF-ViT~\cite{nnfoundation} & 70.3 & 66.8 \\
\bottomrule
\end{tabular}
\end{table}

\paragraph{Lesion anomaly.} With no head, \ours finds stroke and glioma lesions better than voxel intensity does,
and it has the highest anomaly AUPRC, the area under the precision-recall curve, on both MRI sets and, with the bank of
every lesion-free training liver, on TriALS (\cref{tab:supp-lesion}). Its lead over \nnfvit holds case by case:
\ours has the higher AUPRC on 92 to 97\% of the cases of each set (TriALS with the all-train bank), the win shares of
\maincref{fig:analysis}. It is the only encoder that beats the voxel-intensity floor on both MRI sets: \ours $-$ floor
is $+$0.118 [$+$0.023, $+$0.230] AUPRC on ISLES'22 and
$+$0.399 [$+$0.274, $+$0.553] on BraTS-Africa (paired bootstrap, \cref{sec:supp-impl}). With the three-reference bank,
SPECTRE leads on TriALS, and its lead is significant. Large lesions carry the TriALS means (\maincref{sec:backbone}):
the median per-case AUPRC of \ours is 0.027, with an interquartile range of 0.005 to 0.279.

\begin{table}[!htbp]
\centering\small\setlength{\tabcolsep}{3pt}
\caption{\textbf{Lesion anomaly probe} (voxel AUPRC as a fraction, mean with 95\% bootstrap CI below,
\maincref{fig:analysis}). TriALS banks: every lesion-free training liver (all-train) or three reference cases (3-ref).
Floor: voxel-intensity z-score (MRI) or hypodensity score (CT). Wins: share of cases on which the AUPRC of \ours exceeds
that of \nnfvit. $\ast$: the paired 95\% CI of (\ours $-$ model) excludes 0. --: not scored. Bold: best per column.}
\label{tab:supp-lesion}
\resizebox{\ifdim\width>\linewidth\linewidth\else\width\fi}{!}{%
\begin{tabular}{l cccc}
\toprule
 & ISLES'22 & BraTS-Africa & TriALS & TriALS \\
encoder & (MRI) & (MRI) & (all-train bank) & (3-ref bank) \\
\midrule
\ours & \makecell{\textbf{0.325}\\[-2pt]{\tiny[0.227, 0.432]}} & \makecell{\textbf{0.722}\\[-2pt]{\tiny[0.602, 0.828]}} & \makecell{\textbf{0.163}\\[-2pt]{\tiny[0.109, 0.222]}} & \makecell{0.179\\[-2pt]{\tiny[0.125, 0.240]}} \\
nnF-ViT~\cite{nnfoundation} & \makecell{0.088$^\ast$\\[-2pt]{\tiny[0.050, 0.130]}} & \makecell{0.174$^\ast$\\[-2pt]{\tiny[0.130, 0.221]}} & \makecell{0.067$^\ast$\\[-2pt]{\tiny[0.041, 0.097]}} & \makecell{0.064$^\ast$\\[-2pt]{\tiny[0.040, 0.092]}} \\
SPECTRE~\cite{spectre} & -- & -- & \makecell{0.092$^\ast$\\[-2pt]{\tiny[0.056, 0.134]}} & \makecell{\textbf{0.255}$^\ast$\\[-2pt]{\tiny[0.180, 0.330]}} \\
Triad CNN $\cdot$ MAE~\cite{triad} & \makecell{0.017$^\ast$\\[-2pt]{\tiny[0.011, 0.026]}} & \makecell{0.139$^\ast$\\[-2pt]{\tiny[0.090, 0.195]}} & -- & -- \\
\addlinespace[2pt]
\ours\ wins vs.\ \nnfvit & 92\% {\tiny(45/49)} & 97\% {\tiny(28/29)} & 96\% {\tiny(70/73)} & -- \\
\midrule
floor & \makecell{0.207\\[-2pt]{\tiny[0.128, 0.312]}} & \makecell{0.323\\[-2pt]{\tiny[0.209, 0.449]}} & \makecell{0.081\\[-2pt]{\tiny[0.049, 0.118]}} & -- \\
chance & 0.013 & 0.091 & 0.071 & -- \\
\bottomrule
\end{tabular}}
\end{table}

\paragraph{Anatomy with no head.} k-means on the tokens of one case groups its anatomy with no labels. On the four
MRI sets of \cref{tab:supp-kmeans} and AMOS22 CT, \ours has the highest Dice in 14 of the 15 cells of dataset and
cluster count $K$. At $K{=}32$ on ATLAS liver, its liver splits over several clusters and \nnfvit leads. At $K{=}16$,
the setting of \maincref{fig:analysis}, \ours has a higher Dice than \nnfvit on 67 to 100\% of the cases, and it has the
highest normalized mutual information (NMI) on four of the five sets, all but Prostate158, where Triad CNN $\cdot$ MAE
leads.

\begin{table}[!htbp]
\centering\small\setlength{\tabcolsep}{5pt}
\caption{\textbf{Anatomy with no head: k-means} (Dice \%, validation cases of one fold, \maincref{fig:analysis}). Each
structure is matched to one cluster, and Dice is the mean over structures. $K$: clusters per case. NMI: normalized
mutual information at $K{=}16$. Win: share of cases (\%) on which \ours has a higher Dice than \nnfvit at $K{=}16$.
Bold: best per dataset and column.}
\label{tab:supp-kmeans}
\begin{tabular}{@{}ll ccc c c@{}}
\toprule
& & \multicolumn{3}{c}{Dice (\%)} & & \\
\cmidrule(lr){3-5}
dataset & encoder & $K{=}8$ & $K{=}16$ & $K{=}32$ & NMI & win \\
\midrule
Prostate158    & \ours & \textbf{30.9} & \textbf{39.9} & \textbf{51.9} & 0.17 & 100 \\
               & \nnfvit & 10.8 & 12.3 & 14.4 & 0.16 & -- \\
               & Triad CNN $\cdot$ MAE & 16.9 & 31.0 & 41.5 & \textbf{0.23} & -- \\
\addlinespace[2pt]
ACDC           & \ours & \textbf{13.7} & \textbf{24.7} & \textbf{38.1} & \textbf{0.20} & 100 \\
               & \nnfvit & 3.7 & 4.5 & 6.0 & 0.11 & -- \\
               & Triad CNN $\cdot$ MAE & 3.9 & 7.0 & 10.8 & 0.08 & -- \\
\addlinespace[2pt]
ATLAS liver    & \ours & \textbf{46.9} & \textbf{42.8} & 31.2 & \textbf{0.16} & 67 \\
               & \nnfvit & 31.2 & 32.2 & \textbf{33.2} & 0.03 & -- \\
               & Triad CNN $\cdot$ MAE & 35.2 & 32.1 & 28.0 & 0.07 & -- \\
\addlinespace[2pt]
AMOS-MRI       & \ours & \textbf{11.8} & \textbf{15.2} & \textbf{24.5} & \textbf{0.38} & 95 \\
               & \nnfvit & 6.0 & 7.0 & 8.0 & 0.13 & -- \\
               & Triad CNN $\cdot$ MAE & 6.8 & 9.4 & 12.9 & 0.26 & -- \\
\addlinespace[2pt]
AMOS22 (CT)    & \ours & \textbf{17.5} & \textbf{25.2} & \textbf{32.6} & \textbf{0.46} & 100 \\
               & \nnfvit & 3.9 & 6.0 & 9.0 & 0.17 & -- \\
               & VoxelFM & 15.5 & 22.9 & 29.6 & 0.42 & -- \\
\bottomrule
\end{tabular}
\end{table}

\paragraph{Protocols.} No probe trains anything on the frozen tokens. \emph{k-NN vote:} the vote of
\cref{tab:mri-cls} ($k{=}20$), with each dataset's split and targets (\cref{tab:supp-cls-splits}). Both readouts of
\cref{tab:supp-clstok}a take the final teacher's last-block tokens after the final LayerNorm, with the class token kept
in the forward pass. \emph{Lesion anomaly:} the 8\,mm grid, the anomaly score, the MRI banks, the five TriALS folds
and the bootstrap follow \cref{sec:supp-impl}. \Cref{tab:supp-lesion} reads frozen last-layer tokens and scores
ISLES'22~\cite{data_isles22} (49 cases),
BraTS-Africa~\cite{data_bratsafrica} (29 cases) and TriALS~\cite{data_trials}, whose five folds score 73 portal-venous
CT cases inside the liver. Both TriALS banks are fixed before any TriALS scoring: every lesion-free training
liver of the fold (the PatchCore rule) and, as an ablation, the first three training cases of the fold with a lesion
(the rule of the MRI banks). Encoders pretrained on one modality are scored only on it. \emph{k-means:} k-means (cosine,
5 restarts) clusters the five student taps that the segmentation heads read (before the final LayerNorm, each channel
standardized) of each validation case of one fold at a fixed $K$ of 8, 16 or
32. Each structure is matched to one cluster (Hungarian assignment) and scored by Dice (\%), the mean over structures.
NMI, between clusters and labels on foreground tokens at $K{=}16$, needs no matching. On CT, the third encoder is VoxelFM, as Triad CNN $\cdot$ MAE reads only MRI.

\FloatBarrier
\section{Pretraining data attribution}
\label{sec:supp-attribution}
{\scriptsize
The CC~BY licenses in the pretraining corpus require attribution, and three lists credit their sources: the 21 TCIA
collections of the MRI corpus, the 49 TCIA collections of the CT corpus and the 13 OpenNeuro datasets under CC~BY
(\cref{tab:supp-corpus-mri,tab:supp-corpus-ct}). The other 882 OpenNeuro datasets (848 under CC0~1.0, 31 under PDDL~1.0
and 3 in the public domain) are public-domain dedications and need no attribution. TCIA entries open with the
collection name used in \cref{tab:supp-corpus-mri,tab:supp-corpus-ct}, then list the creators, year, title, version,
publisher, DOI and license as registered with DataCite for the DOI; the licenses are CC~BY~4.0
(\url{https://creativecommons.org/licenses/by/4.0/}) and CC~BY~3.0 (\url{https://creativecommons.org/licenses/by/3.0/}). OpenNeuro entries open with the dataset accession
number and list the same fields without a version, and the 2 datasets without a registered DOI give their OpenNeuro page
and no year. The same changes apply to all attributed data: MRI and CT volumes are converted to NIfTI where distributed
as DICOM, and are resampled, intensity-normalized and cropped for model input. We use no other files and do not
redistribute the data.\par}
\newcommand{\attrentry}[1]{\unskip\enspace\textbf{[#1]}~\ignorespaces}

\par\smallskip\noindent\textbf{MRI: 21 TCIA collections}\par\nopagebreak\smallskip
{\tiny\raggedright\urlstyle{same}\setlength{\parindent}{0pt}\setlength{\parskip}{0.3pt}%
\attrentry{1}\textbf{ACRIN-6698}. Newitt, David C.; Partridge, Savannah C.; Zhang, Zheng; Gibbs, Jessica; Chenevert,
Thomas; Rosen, Mark; Bolan, Patrick; Marques, Helga; Romanoff, Justin; Cimino, Lisa; Joe, Bonnie N.; Umphrey, Heidi;
Ojeda-Fournier, Haydee; Dogan, Basak; Oh, Karen Y.; Abe, Hiroyuki; Drukteinis, Jennifer; Esserman, Laura J.; Hylton,
Nola M. (2021). \emph{ACRIN 6698/I-SPY2 Breast DWI} (Version 1). The Cancer Imaging Archive.
\url{https://doi.org/10.7937/tcia.kk02-6d95}. CC~BY~4.0.
\attrentry{2}\textbf{ACRIN-Contralateral-Breast-MR}. Kinahan, Paul; Muzi, Mark; Bialecki, Brian; Herman, Benjamin;
Coombs, Laura (2021). \emph{ACRIN-Contralateral-Breast-MR (ACRIN 6667)} (Version 1). The Cancer Imaging
Archive. \url{https://doi.org/10.7937/Q1EE-J082}. CC~BY~4.0.
\attrentry{3}\textbf{Advanced-MRI-Breast-Lesions}. Daniels, Dianne; Last, David; Cohen, Kfir; Mardor, Yael;
Sklair-Levy, Miri (2024). \emph{Standard and Delayed Contrast-Enhanced MRI of Malignant and Benign Breast Lesions with
Histological and Clinical Supporting Data (Advanced-MRI-Breast-Lesions)} (Version 2). The Cancer Imaging
Archive. \url{https://doi.org/10.7937/C7X1-YN57}. CC~BY~4.0.
\attrentry{4}\textbf{BREAST-DIAGNOSIS}. Bloch, B. Nicolas; Jain, Ashali; jaffe, conrade carl (2015). \emph{Data From
BREAST-DIAGNOSIS} (Version 1). The Cancer Imaging Archive.
\url{https://doi.org/10.7937/K9/TCIA.2015.SDNRQXXR}. CC~BY~3.0.
\attrentry{5}\textbf{Breast-MRI-NACT-Pilot}. Newitt, David; Hylton, Nola (2016). \emph{Single site breast DCE-MRI data
and segmentations from patients undergoing neoadjuvant chemotherapy} (Version 2). The Cancer Imaging
Archive. \url{https://doi.org/10.7937/K9/TCIA.2016.QHsyhJKy}. CC~BY~3.0.
\attrentry{6}\textbf{CPTAC-UCEC}. National Cancer Institute Clinical Proteomic Tumor Analysis Consortium (CPTAC)
(2019). \emph{The Clinical Proteomic Tumor Analysis Consortium Uterine Corpus Endometrial Carcinoma Collection
(CPTAC-UCEC)} (Version 13). The Cancer Imaging Archive.
\url{https://doi.org/10.7937/K9/TCIA.2018.3R3JUISW}. CC~BY~4.0.
\attrentry{7}\textbf{EA1141}. Comstock, Christopher E.; Gatsonis, Constantine; Newstead, Gillian M.; Snyder, Bradley
S.; Gareen, Ilana F.; Bergin, Jennifer T.; Rahbar, Habib; Sung, Janice S.; Jacobs, Christina; Harvey, Jennifer A.;
Nicholson, Mary H.; Ward, Robert C.; Holt, Jacqueline; Prather, Andrew; Miller, Kathy D.; Schnall, Mitchell D.; Kuhl,
Christiane K. (2023). \emph{Abbreviated Breast MRI and Digital Tomosynthesis Mammography in Screening Women With Dense
Breasts (EA1141)} (Version 2). The Cancer Imaging Archive. \url{https://doi.org/10.7937/2BAS-HR33}.
CC~BY~4.0.
\attrentry{8}\textbf{ISPY1}. Newitt, David; Hylton, Nola (2016). \emph{Multi-center breast DCE-MRI data and
segmentations from patients in the I-SPY 1/ACRIN 6657 trials} (Version 2). The Cancer Imaging Archive.
\url{https://doi.org/10.7937/K9/TCIA.2016.HdHpgJLK}. CC~BY~3.0.
\attrentry{9}\textbf{ISPY2}. Li, Wen; Newitt, David C.; Gibbs, Jessica; Wilmes, Lisa J.; Jones, Ella F.; Arasu,
Vignesh A.; Strand, Fredrik; Onishi, Natsuko; Nguyen, Alex Anh-Tu; Kornak, John; Joe, Bonnie N.; Price, Elissa R.;
Ojeda-Fournier, Haydee; Eghtedari, Mohammad; Zamora, Kathryn W.; Woodard, Stefanie A.; Umphrey, Heidi; Bernreuter,
Wanda; Nelson, Michael; Church, An Ly; Bolan, Patrick; Kuritza, Theresa; Ward, Kathleen; Morley, Kevin; Wolverton,
Dulcy; Fountain, Kelly; Lopez-Paniagua, Dan; Hardesty, Lara; Brandt, Kathy; McDonald, Elizabeth S.; Rosen, Mark;
Kontos, Despina; Abe, Hiroyuki; Sheth, Deepa; Crane, Erin P.; Dillis, Charlotte; Sheth, Pulin; Hovanessian-Larsen,
Linda; Bang, Dae Hee; Porter, Bruce; Oh, Karen Y.; Jafarian, Neda; Tudorica, Alina; Niell, Bethany L.; Drukteinis,
Jennifer; Newell, Mary S.; Cohen, Michael A.; Giurescu, Marina; Berman, Elise; Lehman, Constance; Partridge, Savannah
C.; Fitzpatrick, Kimberly A.; Borders, Marisa H.; Yang, Wei T.; Dogan, Basak; Goudreau, Sally; Chenevert, Thomas; Yau,
Christina; DeMichele, Angela; Berry, Don; Esserman, Laura J.; Hylton, Nola M. (2022). \emph{I-SPY 2 Breast Dynamic
Contrast Enhanced MRI (I-SPY2 TRIAL)} (Version 1). The Cancer Imaging Archive.
\url{https://doi.org/10.7937/TCIA.D8Z0-9T85}. CC~BY~4.0.
\attrentry{10}\textbf{Prostate Fused-MRI-Pathology}. Madabhushi, Anant; Feldman, Michael (2016). \emph{Fused
Radiology-Pathology Prostate Dataset} (Version 1). The Cancer Imaging Archive.
\url{https://doi.org/10.7937/k9/TCIA.2016.tlpmr1am}. CC~BY~3.0.
\attrentry{11}\textbf{Prostate-3T}. Litjens, Geert; Futterer, Jurgen; Huisman, Henkjan (2015). \emph{Data From
Prostate-3T} (Version 1). The Cancer Imaging Archive. \url{https://doi.org/10.7937/K9/TCIA.2015.QJTV5IL5}.
CC~BY~3.0.
\attrentry{12}\textbf{PROSTATE-DIAGNOSIS}. Bloch, B Nicolas; Jain, Ashali; Jaffe, C. Carl (2015). \emph{Data From
PROSTATE-DIAGNOSIS} (Version 2). The Cancer Imaging Archive.
\url{https://doi.org/10.7937/K9/TCIA.2015.FOQEUJVT}. CC~BY~3.0.
\attrentry{13}\textbf{PROSTATE-MRI}. Choyke, Peter; Turkbey, Baris; Pinto, Peter; Merino, Maria; Wood, Brad (2016).
\emph{Data From PROSTATE-MRI} (Version 1). The Cancer Imaging Archive.
\url{https://doi.org/10.7937/K9/TCIA.2016.6046GUDv}. CC~BY~3.0.
\attrentry{14}\textbf{Prostate-MRI-US-Biopsy}. Natarajan, Shyam; Priester, Alan; Margolis, Daniel; Huang, Jiaoti;
Marks, Leonard (2020). \emph{Prostate MRI and Ultrasound With Pathology and Coordinates of Tracked Biopsy
(Prostate-MRI-US-Biopsy)} (Version 2). The Cancer Imaging Archive.
\url{https://doi.org/10.7937/TCIA.2020.A61IOC1A}. CC~BY~4.0.
\attrentry{15}\textbf{PROSTATEx}. Litjens, Geert; Debats, Oscar; Barentsz, Jelle; Karssemeijer, Nico; Huisman, Henkjan
(2017). \emph{SPIE-AAPM PROSTATEx Challenge Data} (Version 2). The Cancer Imaging Archive.
\url{https://doi.org/10.7937/K9TCIA.2017.MURS5CL}. CC~BY~3.0.
\attrentry{16}\textbf{QIN Breast DCE-MRI}. Huang, Wei; Tudorica, Alina; Chui, Stephen; Kemmer, Kathleen; Naik, Arpana;
Troxell, Megan; Oh, Karen; Roy, Nicole; Afzal, Aneela; Holtorf, Megan (2014). \emph{Variations of dynamic
contrast-enhanced magnetic resonance imaging in evaluation of breast cancer therapy response: a multicenter data
analysis challenge (QIN Breast DCE-MRI)} (Version 2). The Cancer Imaging Archive.
\url{https://doi.org/10.7937/k9/tcia.2014.a2n1ixox}. CC~BY~3.0.
\attrentry{17}\textbf{ReMIND}. Juvekar, Parikshit; Dorent, Reuben; K\"{o}gl, Fryderyk; Torio, Erickson; Barr, Colton;
Rigolo, Laura; Galvin, Colin; Jowkar, Nick; Kazi, Anees; Haouchine, Nazim; Cheema, Harneet; Navab, Nassir; Pieper,
Steve; Wells, William M.; Bi, Wenya Linda; Golby, Alexandra; Frisken, Sarah; Kapur, Tina (2023). \emph{The Brain
Resection Multimodal Imaging Database (ReMIND)} (Version 1). The Cancer Imaging Archive.
\url{https://doi.org/10.7937/3RAG-D070}. CC~BY~4.0.
\attrentry{18}\textbf{TCGA-BRCA}. Lingle, Wilma; Erickson, Bradley J.; Zuley, Margarita L.; Jarosz, Rose; Bonaccio,
Ermelinda; Filippini, Joe; Net, Jose M.; Levi, Len; Morris, Elizabeth A.; Figler, Gloria G.; Elnajjar, Pierre; Kirk,
Shanah; Lee, Yueh; Giger, Maryellen; Gruszauskas, Nicholas (2016). \emph{The Cancer Genome Atlas Breast Invasive
Carcinoma Collection (TCGA-BRCA)} (Version 3). The Cancer Imaging Archive.
\url{https://doi.org/10.7937/K9/TCIA.2016.AB2NAZRP}. CC~BY~3.0.
\attrentry{19}\textbf{TCGA-PRAD}. Zuley, Margarita L.; Jarosz, Rose; Drake, Bettina F.; Rancilio, Danielle; Klim,
Aleksandra; Rieger-Christ, Kimberly; Lemmerman, John (2016). \emph{The Cancer Genome Atlas Prostate Adenocarcinoma
Collection (TCGA-PRAD)} (Version 4). The Cancer Imaging Archive.
\url{https://doi.org/10.7937/K9/TCIA.2016.YXOGLM4Y}. CC~BY~3.0.
\attrentry{20}\textbf{UPENN-GBM}. Bakas, Spyridon; Sako, Chiharu; Akbari, Hamed; Bilello, Michel; Sotiras, Aristeidis;
Shukla, Gaurav; Rudie, Jeffrey D.; Flores Santamaria, Natali; Fathi Kazerooni, Anahita; Pati, Sarthak; Rathore, Saima;
Mamourian, Elizabeth; Ha, Sung Min; Parker, William; Doshi, Jimit; Baid, Ujjwal; Bergman, Mark; Binder, Zev A.; Verma,
Ragini; Lustig, Robert; Desai, Arati; Bagley, Stephen; Mourelatos, Zissimos; Morrissette, Jennifer; Watt, Christopher;
Brem, Steven; Wolf, Ronald; Nasrallah, MacLean P.; Mohan, Suyash; O'Rourke, Donald M.; Davatzikos, Christos (2021).
\emph{Multi-parametric magnetic resonance imaging (mpMRI) scans for de novo Glioblastoma (GBM) patients from the
University of Pennsylvania Health System (UPENN-GBM)} (Version 2). The Cancer Imaging Archive.
\url{https://doi.org/10.7937/TCIA.709X-DN49}. CC~BY~4.0.
\attrentry{21}\textbf{Vestibular-Schwannoma-SEG}. Shapey, Jonathan; Kujawa, Aaron; Dorent, Reuben; Wang, Guotai;
Bisdas, Sotirios; Dimitriadis, Alexis; Grishchuck, Diana; Paddick, Ian; Kitchen, Neil; Bradford, Robert; Saeed,
Shakeel; Ourselin, Sebastien; Vercauteren, Tom (2021). \emph{Segmentation of Vestibular Schwannoma from Magnetic
Resonance Imaging: An Open Annotated Dataset and Baseline Algorithm (Vestibular-Schwannoma-SEG)} (Version 2). The Cancer Imaging Archive. \url{https://doi.org/10.7937/TCIA.9YTJ-5Q73}. CC~BY~4.0.
\par}

\par\smallskip\noindent\textbf{CT: 49 TCIA collections}\par\nopagebreak\smallskip
{\tiny\raggedright\urlstyle{same}\setlength{\parindent}{0pt}\setlength{\parskip}{0.3pt}%
\attrentry{1}\textbf{ADRENAL-ACC-KI67-SEG}. Moawad, Ahmed W.; Ahmed, Ayahallah A.; ElMohr, Mohab; Eltaher, Mohamed;
Habra, Mouhammed Amir; Fisher, Sarah; Perrier, Nancy; Zhang, Miao; Fuentes, David; Elsayes, Khaled (2023).
\emph{Voxel-level segmentation of pathologically-proven Adrenocortical carcinoma with Ki-67 expression
(Adrenal-ACC-Ki67-Seg)} (Version 1). The Cancer Imaging Archive. \url{https://doi.org/10.7937/1fpg-vm46}.
CC~BY~4.0.
\attrentry{2}\textbf{ANTI-PD-1\_LUNG}. Patnana, Madhavi; Patel, Sapna; Tsao, Anne S (2019). \emph{Data from Anti-PD-1
Immunotherapy Lung} (Version 1). The Cancer Imaging Archive.
\url{https://doi.org/10.7937/tcia.2019.zjjwb9ip}. CC~BY~3.0.
\attrentry{3}\textbf{BREAST-DIAGNOSIS}. Bloch, B. Nicolas; Jain, Ashali; jaffe, conrade carl (2015). \emph{Data From
BREAST-DIAGNOSIS} (Version 1). The Cancer Imaging Archive.
\url{https://doi.org/10.7937/k9/tcia.2015.sdnrqxxr}. CC~BY~3.0.
\attrentry{4}\textbf{CC-TUMOR-HETEROGENEITY}. Mayr, Nina; Yuh, William T.C.; Bowen, Stephen; Harkenrider, Matthew;
Knopp, Michael V.; Lee, Elaine Yuen-Phin; Leung, Eric; Lo, Simon S.; Small Jr., William; Wolfson, Aaron H. (2023).
\emph{Cervical Cancer - Tumor Heterogeneity: Serial Functional and Molecular Imaging Across the Radiation Therapy
Course in Advanced Cervical Cancer (CC-Tumor-Heterogeneity)} (Version 2). The Cancer Imaging Archive.
\url{https://doi.org/10.7937/erz5-qz59}. CC~BY~4.0.
\attrentry{5}\textbf{CMB-OV}. Cancer Moonshot Biobank (2024). \emph{Cancer Moonshot Biobank - Ovarian Carcinoma Cancer
Collection (CMB-OV)} (Version 4). The Cancer Imaging Archive. \url{https://doi.org/10.7937/4nx6-e061}.
CC~BY~4.0.
\attrentry{6}\textbf{COLORECTAL-LIVER-METASTASES}. Simpson, Amber L.; Peoples, Jacob; Creasy, John M.; Fichtinger,
Gabor; Gangai, Natalie; Lasso, Andras; Keshava Murthy, Krishna Nand; Shia, Jinru; D'Angelica, Michael I.; Do, Richard
K. G. (2023). \emph{Preoperative CT and Survival Data for Patients Undergoing Resection of Colorectal Liver Metastases
(Colorectal-Liver-Metastases)} (Version 2). The Cancer Imaging Archive.
\url{https://doi.org/10.7937/qxk2-qg03}. CC~BY~4.0.
\attrentry{7}\textbf{COVID-19-AR}. Desai, Shivang; Baghal, Ahmad; Wongsurawat, Thidathip; Al-Shukri, Shaymaa; Gates,
Kim; Farmer, Phillip; Rutherford, Michael; Blake, Geri D.; Nolan, Tracy; Powell, Thomas; Sexton, Kevin; Bennett,
William; Prior, Fred (2020). \emph{Chest Imaging with Clinical and Genomic Correlates Representing a Rural COVID-19
Positive Population} (Version 1). The Cancer Imaging Archive.
\url{https://doi.org/10.7937/tcia.2020.py71-5978}. CC~BY~4.0.
\attrentry{8}\textbf{COVID-19-NY-SBU}. Saltz, Joel; Saltz, Mary; Prasanna, Prateek; Moffitt, Richard; Hajagos, Janos;
Bremer, Erich; Balsamo, Joseph; Kurc, Tahsin (2021). \emph{Stony Brook University COVID-19 Positive Cases} (Version 1). The Cancer Imaging Archive. \url{https://doi.org/10.7937/tcia.bbag-2923}. CC~BY~4.0.
\attrentry{9}\textbf{CPTAC-AML}. National Cancer Institute Clinical Proteomic Tumor Analysis Consortium (CPTAC)
(2019). \emph{The Clinical Proteomic Tumor Analysis Consortium Acute Myeloid Leukemia Collection (CPTAC-AML)} (Version
5). The Cancer Imaging Archive. \url{https://doi.org/10.7937/tcia.2019.b6foe619}. CC~BY~4.0.
\attrentry{10}\textbf{CPTAC-CCRCC}. National Cancer Institute Clinical Proteomic Tumor Analysis Consortium (CPTAC)
(2018). \emph{The Clinical Proteomic Tumor Analysis Consortium Clear Cell Renal Cell Carcinoma Collection
(CPTAC-CCRCC)} (Version 14). The Cancer Imaging Archive.
\url{https://doi.org/10.7937/k9/tcia.2018.oblamn27}. CC~BY~4.0.
\attrentry{11}\textbf{CPTAC-CM}. National Cancer Institute Clinical Proteomic Tumor Analysis Consortium (CPTAC)
(2018). \emph{The Clinical Proteomic Tumor Analysis Consortium Cutaneous Melanoma Collection (CPTAC-CM)} (Version 11). The Cancer Imaging Archive. \url{https://doi.org/10.7937/k9/tcia.2018.odu24gze}. CC~BY~4.0.
\attrentry{12}\textbf{CPTAC-LUAD}. National Cancer Institute Clinical Proteomic Tumor Analysis Consortium (CPTAC)
(2018). \emph{The Clinical Proteomic Tumor Analysis Consortium Lung Adenocarcinoma Collection (CPTAC-LUAD)} (Version
13). The Cancer Imaging Archive. \url{https://doi.org/10.7937/k9/tcia.2018.pat12tbs}. CC~BY~4.0.
\attrentry{13}\textbf{CPTAC-PDA}. National Cancer Institute Clinical Proteomic Tumor Analysis Consortium (CPTAC)
(2018). \emph{The Clinical Proteomic Tumor Analysis Consortium Pancreatic Ductal Adenocarcinoma Collection
(CPTAC-PDA)} (Version 15). The Cancer Imaging Archive. \url{https://doi.org/10.7937/k9/tcia.2018.sc20fo18}.
CC~BY~4.0.
\attrentry{14}\textbf{CPTAC-SAR}. National Cancer Institute Clinical Proteomic Tumor Analysis Consortium (CPTAC)
(2019). \emph{The Clinical Proteomic Tumor Analysis Consortium Sarcomas Collection (CPTAC-SAR)} (Version 10). The Cancer Imaging Archive. \url{https://doi.org/10.7937/tcia.2019.9bt23r95}. CC~BY~4.0.
\attrentry{15}\textbf{CPTAC-STAD}. National Cancer Institute Clinical Proteomic Tumor Analysis Consortium (CPTAC)
(2026). \emph{The Clinical Proteomic Tumor Analysis Consortium Stomach Adenocarcinoma Collection (CPTAC-STAD)}
(Version 1). The Cancer Imaging Archive. \url{https://doi.org/10.7937/jw9a-8k71}. CC~BY~4.0.
\attrentry{16}\textbf{CPTAC-UCEC}. National Cancer Institute Clinical Proteomic Tumor Analysis Consortium (CPTAC)
(2019). \emph{The Clinical Proteomic Tumor Analysis Consortium Uterine Corpus Endometrial Carcinoma Collection
(CPTAC-UCEC)} (Version 13). The Cancer Imaging Archive.
\url{https://doi.org/10.7937/k9/tcia.2018.3r3juisw}. CC~BY~4.0.
\attrentry{17}\textbf{CT-LYMPH-NODES}. Roth, Holger; Lu, Le; Seff, Ari; Cherry, Kevin M; Hoffman, Joanne; Wang,
Shijun; Liu, Jiamin; Turkbey, Evrim; Summers, Ronald M. (2015). \emph{A new 2.5 D representation for lymph node
detection in CT (CT Lymph Nodes)} (Version 5). The Cancer Imaging Archive.
\url{https://doi.org/10.7937/k9/tcia.2015.aqiidcnm}. CC~BY~3.0.
\attrentry{18}\textbf{CT-VS-PET-VENTILATION-IMAGING}. Eslick, Enid M.; Kipritidis, John; Gradinscak, Denis; Stevens,
Mark J.; Bailey, Dale L.; Harris, Benjamin; Booth, Jeremy T.; Keall, Paul J. (2022). \emph{CT Ventilation as a
functional imaging modality for lung cancer radiotherapy (CT-vs-PET-Ventilation-Imaging)} (Version 1). The
Cancer Imaging Archive. \url{https://doi.org/10.7937/3ppx-7s22}. CC~BY~4.0.
\attrentry{19}\textbf{CTPRED-SUNITINIB-PANNET}. Chen, Luohai; Wang, Wei; Jin, Kaizhou; Yuan, Bing; Tan, Huangying;
Sun, Jian; Guo, Yu; Luo, Yanji; Feng, Shi-ting; Yu, Xianjun; Chen, Min-hu; Chen, Jie (2022). \emph{Prediction of
Sunitinib Efficacy using Computed Tomography in Patients with Pancreatic Neuroendocrine Tumors
(CTpred-Sunitinib-panNET)} (Version 1). The Cancer Imaging Archive.
\url{https://doi.org/10.7937/spgk-0p94}. CC~BY~4.0.
\attrentry{20}\textbf{EA1141}. Comstock, Christopher E.; Gatsonis, Constantine; Newstead, Gillian M.; Snyder, Bradley
S.; Gareen, Ilana F.; Bergin, Jennifer T.; Rahbar, Habib; Sung, Janice S.; Jacobs, Christina; Harvey, Jennifer A.;
Nicholson, Mary H.; Ward, Robert C.; Holt, Jacqueline; Prather, Andrew; Miller, Kathy D.; Schnall, Mitchell D.; Kuhl,
Christiane K. (2023). \emph{Abbreviated Breast MRI and Digital Tomosynthesis Mammography in Screening Women With Dense
Breasts (EA1141)} (Version 2). The Cancer Imaging Archive. \url{https://doi.org/10.7937/2bas-hr33}.
CC~BY~4.0.
\attrentry{21}\textbf{HCC-TACE-SEG}. Moawad, Ahmed W.; Fuentes, David; Morshid, Ali; Khalaf, Ahmed M.; Elmohr, Mohab
M.; Abusaif, Abdelrahman; Hazle, John D.; Kaseb, Ahmed O.; Hassan, Manal; Mahvash, Armeen; Szklaruk, Janio; Qayyom,
Aliyya; Elsayes, Khaled (2021). \emph{Multimodality annotated HCC cases with and without advanced imaging
segmentation} (Version 1). The Cancer Imaging Archive. \url{https://doi.org/10.7937/tcia.5fna-0924}.
CC~BY~4.0.
\attrentry{22}\textbf{LUNGCT-DIAGNOSIS}. Grove, Olya; Berglund, Anders E.; Schabath, Matthew B.; Aerts, Hugo J.W.L.;
Dekker, Andre; Wang, Hua; Velazquez, Emmanuel Rios; Lambin, Philippe; Gu, Yuhua; Balagurunathan, Yoganand; Eikman,
Edward; Gatenby, Robert A.; Eschrich, Steven; Gillies, Robert J. (2015). \emph{Data from: Quantitative computed
tomographic descriptors associate tumor shape complexity and intratumor heterogeneity with prognosis in lung
adenocarcinoma} (Version 1). The Cancer Imaging Archive.
\url{https://doi.org/10.7937/k9/tcia.2015.a6v7jiwx}. CC~BY~3.0.
\attrentry{23}\textbf{MEDIASTINAL-LYMPH-NODE-SEG}. Idris, Tagwa; Somarouthu, Supriya; Jacene, Heather; LaCasce, A;
Ziegler, Erik; Pieper, Steve; Khajavi, Roya; Dorent, Reuben; Pujol, Sonia; Kikinis, Ron; Harris, Gordon (2024).
\emph{Mediastinal Lymph Node Quantification (LNQ): Segmentation of Heterogeneous CT Data} (Version 1). The
Cancer Imaging Archive. \url{https://doi.org/10.7937/qvaz-ja09}. CC~BY~4.0.
\attrentry{24}\textbf{NaF PROSTATE}. Kurdziel, Karen A; Apolo, Andrea B.; Lindenberg, Liza; Mena, Esther; McKinney,
Yolanda Y.; Adler, Stephen S.; Turkbey, Baris; Dahut, William; Gulley, James L.; Madan, Ravi A.; Landgren, Ola;
Choyke, Peter L. (2015). \emph{Data From NaF PROSTATE} (Version 1). The Cancer Imaging Archive.
\url{https://doi.org/10.7937/k9/tcia.2015.isoqthko}. CC~BY~3.0.
\attrentry{25}\textbf{NLST}. National Lung Screening Trial Research Team (2013). \emph{Data from the National Lung
Screening Trial (NLST)} (Version 3). The Cancer Imaging Archive.
\url{https://doi.org/10.7937/tcia.hmq8-j677}. CC~BY~4.0.
\attrentry{26}\textbf{Pancreas-CT}. Roth, Holger; Farag, Amal; Turkbey, Evrim B.; Lu, Le; Liu, Jiamin; Summers, Ronald
M. (2016). \emph{Data From Pancreas-CT} (Version 2). The Cancer Imaging Archive.
\url{https://doi.org/10.7937/k9/tcia.2016.tnb1kqbu}. CC~BY~3.0.
\attrentry{27}\textbf{PANCREATIC-CT-CBCT-SEG}. Hong, Jun; Reyngold, Marsha; Crane, Christopher; Cuaron, John; Hajj,
Carla; Mann, Justin; Zinovoy, Melissa; Yorke, Ellen; LoCastro, Eve; Apte, Aditya P.; Mageras, Gig (2021).
\emph{Breath-hold CT and cone-beam CT images with expert manual organ-at-risk segmentations from radiation treatments
of locally advanced pancreatic cancer (Pancreatic-CT-CBCT-SEG)} (Version 2). The Cancer Imaging Archive.
\url{https://doi.org/10.7937/tcia.eshq-4d90}. CC~BY~4.0.
\attrentry{28}\textbf{Pelvic-Reference-Data}. Yorke, Afua A.; McDonald, Gary C.; Solis, David; Guerrero, Thomas
(2019). \emph{Pelvic Reference Data} (Version 1). The Cancer Imaging Archive.
\url{https://doi.org/10.7937/tcia.2019.woskq5oo}. CC~BY~3.0.
\attrentry{29}\textbf{PROSTATE-ANATOMICAL-EDGE-CASES}. Thompson, Reid F.; Kanwar, Aasheesh; Merz, Brandon; Cohen,
Ethan; Fisher, Helen; Rana, Shushan; Claunch, Cheryl; Hung, Arthur (2023). \emph{Stress-Testing Pelvic
Autosegmentation Algorithms Using Anatomical Edge Cases (Prostate-Anatomical-Edge-Cases)} (Version 1). The
Cancer Imaging Archive. \url{https://doi.org/10.7937/qstf-st65}. CC~BY~4.0.
\attrentry{30}\textbf{QIN LUNG CT}. Goldgof, Dmitry; Hall, Lawrence; Hawkins, Samuel; Schabath, Matthew; Stringfield,
Olya; Garcia, Alberto; Balagurunathan, Yoganand; Kim, Jongphil; Eschrich, Steven; Berglund, Anders; Gatenby, Robert;
Gillies, Robert (2015). \emph{Data From QIN LUNG CT} (Version 2). The Cancer Imaging Archive.
\url{https://doi.org/10.7937/k9/tcia.2015.npgzyzbz}. CC~BY~3.0.
\attrentry{31}\textbf{QIN-BREAST}. Li, Xia; Abramson, Richard G.; Arlinghaus, Lori R.; Chakravarthy, Anuradha B.;
Abramson, Vandana G.; Sanders, Melinda; Yankeelov, Thomas E. (2016). \emph{Data From QIN-Breast} (Version 2). The Cancer Imaging Archive. \url{https://doi.org/10.7937/k9/tcia.2016.21juebh0}. CC~BY~3.0.
\attrentry{32}\textbf{RIDER Pilot}. Lung Image Database Consortium (LIDC) (2023). \emph{RIDER Pilot} (Version 1). The Cancer Imaging Archive. \url{https://doi.org/10.7937/m87f-mz83}. CC~BY~4.0.
\attrentry{33}\textbf{RIDER-LUNG-CT}. Zhao, Binsheng; Schwartz, Lawrence H; Kris, Mark G; Riely, Gregory J (2015).
\emph{Coffee-break lung CT collection with scan images reconstructed at multiple imaging parameters} (Version 3). The Cancer Imaging Archive. \url{https://doi.org/10.7937/k9/tcia.2015.u1x8a5nr}. CC~BY~3.0.
\attrentry{34}\textbf{SPIE-AAPM-LUNG-CT-CHALLENGE}. Armato III, Samuel G.; Hadjiiski, Lubomir; Tourassi, Georgia D;
Drukker, Karen; Giger, Maryellen L.; Li, Feng; Redmond, George; Farahani, Keyvan; Kirby, Justin S.; Clarke, Laurence
P. (2015). \emph{SPIE-AAPM-NCI Lung Nodule Classification Challenge Dataset} (Version 2). The Cancer
Imaging Archive. \url{https://doi.org/10.7937/k9/tcia.2015.uzlsu3fl}. CC~BY~3.0.
\attrentry{35}\textbf{SPINAL-MULTIPLE-MYELOMA-SEG}. Nohel, Michal; Valek, Vlastimil; Rohan, Tomas; Stork, Martin;
Jakubicek, Roman; Chmelik, Jiri; Dost\'{a}l, Marek (2026). \emph{Segmentation of spinal multiple myeloma lesions in
dual-energy CT (Spinal-Multiple-Myeloma-SEG)} (Version 1). The Cancer Imaging Archive.
\url{https://doi.org/10.7937/k4qv-hh78}. CC~BY~4.0.
\attrentry{36}\textbf{SPINE-METS-CT-SEG}. Pieper, Steve; Haouchine, Nazim; Hackney, David; Wells, William; Sanhinova,
Malika; Balboni, Tracy; Spektor, Alexander; Huynh, M.; Tanguturi, Shyam; Kim, Ellen; Guenette, Jeffrey; Kozono, David;
Czajkowski, Bryan; Caplan, Sarah; Doyle, Patrick; Kang, Heejoo; Alkalay, Ron (2024). \emph{Spine metastatic bone
cancer: pre and post radiotherapy CT (Spine-Mets-CT-SEG)} (Version 1). The Cancer Imaging Archive.
\url{https://doi.org/10.7937/kh36-ds04}. CC~BY~4.0.
\attrentry{37}\textbf{STAGEII-COLORECTAL-CT}. Tong, Tong; Li, Menglei (2022). \emph{Abdominal or pelvic enhanced CT
images within 10 days before surgery of 230 patients with stage II colorectal cancer} (Version 2). The
Cancer Imaging Archive. \url{https://doi.org/10.7937/p5k5-tg43}. CC~BY~4.0.
\attrentry{38}\textbf{TCGA-COAD}. Kirk, Shanah; Lee, Yueh; Sadow, Cheryl A.; Levine, Seth; Roche, Charles; Bonaccio,
Ermelinda; Filiippini, Joe (2016). \emph{The Cancer Genome Atlas Colon Adenocarcinoma Collection (TCGA-COAD)} (Version
3). The Cancer Imaging Archive. \url{https://doi.org/10.7937/k9/tcia.2016.hjjhboxz}. CC~BY~3.0.
\attrentry{39}\textbf{TCGA-ESCA}. Lucchesi, Fabiano R.; Aredes, Nat\'{a}lia D. (2016). \emph{The Cancer Genome Atlas
Esophageal Carcinoma Collection (TCGA-ESCA)} (Version 3). The Cancer Imaging Archive.
\url{https://doi.org/10.7937/k9/tcia.2016.vptnrgfy}. CC~BY~3.0.
\attrentry{40}\textbf{TCGA-KIRC}. Akin, Oguz; Elnajjar, Pierre; Heller, Matthew; Jarosz, Rose; Erickson, Bradley J.;
Kirk, Shanah; Lee, Yueh; Linehan, Marston W.; Gautam, Rabindra; Vikram, Raghu; Garcia, Kimberly M.; Roche, Charles;
Bonaccio, Ermelinda; Filippini, Joe (2016). \emph{The Cancer Genome Atlas Kidney Renal Clear Cell Carcinoma Collection
(TCGA-KIRC)} (Version 3). The Cancer Imaging Archive. \url{https://doi.org/10.7937/k9/tcia.2016.v6pbvtdr}.
CC~BY~3.0.
\attrentry{41}\textbf{TCGA-KIRP}. Linehan, Marston; Gautam, Rabindra; Kirk, Shanah; Lee, Yueh; Roche, Charles;
Bonaccio, Ermelinda; Filippini, Joe; Rieger-Christ, Kimberly; Lemmerman, John; Jarosz, Rose (2016). \emph{The Cancer
Genome Atlas Cervical Kidney Renal Papillary Cell Carcinoma Collection (TCGA-KIRP)} (Version 4). The Cancer
Imaging Archive. \url{https://doi.org/10.7937/k9/tcia.2016.acwogbef}. CC~BY~3.0.
\attrentry{42}\textbf{TCGA-LUAD}. Albertina, Brad; Watson, Mark; Holback, Chandra; Jarosz, Rose; Kirk, Shanah; Lee,
Yueh; Rieger-Christ, Kimberly; Lemmerman, John (2016). \emph{The Cancer Genome Atlas Lung Adenocarcinoma Collection
(TCGA-LUAD)} (Version 4). The Cancer Imaging Archive. \url{https://doi.org/10.7937/k9/tcia.2016.jgnihep5}.
CC~BY~3.0.
\attrentry{43}\textbf{TCGA-LUSC}. Kirk, Shanahv; Lee, Yueh; Kumar, Prasanna; Filippini, Joe; Albertina, Brad; Watson,
Mark; Rieger-Christ, Kimberly; Lemmerman, John (2016). \emph{The Cancer Genome Atlas Lung Squamous Cell Carcinoma
Collection (TCGA-LUSC)} (Version 4). The Cancer Imaging Archive.
\url{https://doi.org/10.7937/k9/tcia.2016.tygkkfmq}. CC~BY~3.0.
\attrentry{44}\textbf{TCGA-OV}. Holback, Chandra; Jarosz, Rose; Prior, Fred; Mutch, David G.; Bhosale, Priya; Garcia,
Kimberly; Lee, Yueh; Kirk, Shanah; Sadow, Cheryl A.; Levine, Seth; Sala, Evis; Elnajjar, Pierre; Morgan, Tara;
Erickson, Bradley J. (2016). \emph{The Cancer Genome Atlas Ovarian Cancer Collection (TCGA-OV)} (Version 4). The Cancer Imaging Archive. \url{https://doi.org/10.7937/k9/tcia.2016.ndo1mdfq}. CC~BY~3.0.
\attrentry{45}\textbf{TCGA-PRAD}. Zuley, Margarita L.; Jarosz, Rose; Drake, Bettina F.; Rancilio, Danielle; Klim,
Aleksandra; Rieger-Christ, Kimberly; Lemmerman, John (2016). \emph{The Cancer Genome Atlas Prostate Adenocarcinoma
Collection (TCGA-PRAD)} (Version 4). The Cancer Imaging Archive.
\url{https://doi.org/10.7937/k9/tcia.2016.yxoglm4y}. CC~BY~3.0.
\attrentry{46}\textbf{TCGA-READ}. Kirk, Shanah; Lee, Yueh; Sadow, Cheryl A.; Levine, Seth (2016). \emph{The Cancer
Genome Atlas Rectum Adenocarcinoma Collection (TCGA-READ)} (Version 3). The Cancer Imaging Archive.
\url{https://doi.org/10.7937/k9/tcia.2016.f7ppnpnu}. CC~BY~3.0.
\attrentry{47}\textbf{TCGA-STAD}. Lucchesi, Fabiano R.; Aredes, Nat\'{a}lia D. (2016). \emph{The Cancer Genome Atlas
Stomach Adenocarcinoma Collection (TCGA-STAD)} (Version 3). The Cancer Imaging Archive.
\url{https://doi.org/10.7937/k9/tcia.2016.gdhl9kim}. CC~BY~3.0.
\attrentry{48}\textbf{TCGA-UCEC}. Erickson, Bradley J.; Mutch, David; Lippmann, Lynne; Jarosz, Rose (2016). \emph{The
Cancer Genome Atlas Uterine Corpus Endometrial Carcinoma Collection (TCGA-UCEC)} (Version 4). The Cancer
Imaging Archive. \url{https://doi.org/10.7937/k9/tcia.2016.gkj0zwac}. CC~BY~3.0.
\attrentry{49}\textbf{VAREPOP-APOLLO}. The Research for Precision Oncology Program and the Applied Proteogenomics
Organizational Learning and Outcomes (APOLLO) Research Network (2024). \emph{VA Research Precision Oncology Program -
APOLLO (VAREPOP-APOLLO)} (Version 3). The Cancer Imaging Archive. \url{https://doi.org/10.7937/ghkn-md15}.
CC~BY~4.0.
\par}

\par\smallskip\noindent\textbf{MRI: 13 OpenNeuro datasets under CC~BY}\par\nopagebreak\smallskip
{\tiny\raggedright\urlstyle{same}\setlength{\parindent}{0pt}\setlength{\parskip}{0.3pt}%
\attrentry{1}\textbf{ds001378}. Mario Mascalchi. \emph{SCA2 Diffusion Tensor Imaging}. OpenNeuro.
\url{https://openneuro.org/datasets/ds001378}. CC~BY~4.0.
\attrentry{2}\textbf{ds001553}. Javier Gonzalez-Castillo; Ziad S. Saad; Daniel A. Handwerker; Souheil J. Inati; Noah
Brenowitz; Peter A. Bandettini (2023). \emph{100 runs at 3T}. OpenNeuro.
\url{https://doi.org/10.18112/openneuro.ds001553.v1.0.1}. CC~BY, no version stated.
\attrentry{3}\textbf{ds001555}. Javier Gonzalez-Castilloa; Colin W. Hoy; Daniel A. Handwerker; Vinai Roopchansingh;
Souheil J. Inati; Ziad S. Saad; Robert W. Cox; Peter A. Bandettini (2023). \emph{Large Single-Subject Functional MRI
Datasets at 7T}. OpenNeuro. \url{https://doi.org/10.18112/openneuro.ds001555.v1.0.1}. CC~BY, no version stated.
\attrentry{4}\textbf{ds001634}. Adam Hampshire; Eyal Soreq (2019). \emph{Visual working memory: Study one Task fMRI
and Behavioural response}. OpenNeuro. \url{https://doi.org/10.18112/openneuro.ds001634.v1.0.1}. CC~BY~4.0.
\attrentry{5}\textbf{ds001635}. Adam Hampshire; Eyal Soreq (2019). \emph{Visual working memory: Study two Task fMRI
and Behavioural response}. OpenNeuro. \url{https://doi.org/10.18112/openneuro.ds001635.v1.0.1}. CC~BY~4.0.
\attrentry{6}\textbf{ds001652}. Elizabeth Valles (2019). \emph{Ironia VEV}. OpenNeuro.
\url{https://doi.org/10.18112/openneuro.ds001652.v2.0.0}. CC~BY~4.0.
\attrentry{7}\textbf{ds004829}. Ana F. Palenciano; Carlos Gonz\'{a}lez-Garc\'{\i}a; Jan de Houwer; Baptist Lieefoghe;
Marcel Brass. \emph{Concurrent response and action effect representations across the somatomotor cortices during novel
task preparation.}. OpenNeuro. \url{https://openneuro.org/datasets/ds004829}. CC~BY~4.0.
\attrentry{8}\textbf{ds005386}. Chema G. Penalver; David Lopez-Garcia; Marta Becerra Losada (2024).
\emph{AttExp\_fMRI}. OpenNeuro. \url{https://doi.org/10.18112/openneuro.ds005386.v1.0.0}. CC~BY~4.0.
\attrentry{9}\textbf{ds005454}. Emilie C. Sleight; Oscar Esteban; Fr\'{e}d\'{e}ric Grouiller (2024). \emph{fMRIPrep
boot camp -{}- 7T training dataset}. OpenNeuro. \url{https://doi.org/10.18112/openneuro.ds005454.v1.0.0}.
CC~BY~4.0.
\attrentry{10}\textbf{ds006642}. Egor Levchenko; Hugo Chow-Wing-Bom; Fred Dick; Adam Tierney; Jeremy I Skipper (2026).
\emph{Naturalistic Neuroimaging Database 3T+}. OpenNeuro.
\url{https://doi.org/10.18112/openneuro.ds006642.v1.1.1}. CC~BY~4.0.
\attrentry{11}\textbf{ds006644}. Klemens Egger; Milan Scheidegger (2025). \emph{DMT-HAR-MED: Effects of DMT and
harmine during meditation: Functional MRI dataset}. OpenNeuro.
\url{https://doi.org/10.18112/openneuro.ds006644.v1.0.1}. CC~BY~4.0.
\attrentry{12}\textbf{ds006798}. Mitja Nikolaus; Milad Mozafari; Isabelle Berry; Nicholas Asher; Leila Reddy; Rufin
VanRullen (2026). \emph{SemReps-8K}. OpenNeuro. \url{https://doi.org/10.18112/openneuro.ds006798.v1.0.1}.
CC~BY~4.0.
\attrentry{13}\textbf{ds007272}. Mitja Nikolaus; Milad Mozafari; Isabelle Berry; Nicholas Asher; Leila Reddy; Rufin
VanRullen (2026). \emph{SemReps-8K}. OpenNeuro. \url{https://doi.org/10.18112/openneuro.ds007272.v1.0.0}.
CC~BY~4.0.
\par}

\end{document}